\documentclass[10pt,twocolumn,letterpaper]{article}

\usepackage[pagenumbers]{cvpr}            %

\usepackage[dvipsnames]{xcolor}

\newcommand{\ours}[0]{RefineRoITr}
\newcommand{\parsection}[1]{\vspace{0mm}\noindent\textbf{#1}~}
\usepackage{url}

\definecolor{cvprblue}{rgb}{0.21,0.49,0.74}
\usepackage[pagebackref,breaklinks,colorlinks,citecolor=cvprblue]{hyperref}

\usepackage{algpseudocode,algorithm,algorithmicx}

\DeclareMathOperator*{\argmin}{arg\,min}

\newcommand*\Let[2]{\State #1 $\gets$ #2}
\algrenewcommand\algorithmicrequire{\textbf{Precondition:}}
\algrenewcommand\algorithmicensure{\textbf{Postcondition:}}
\newenvironment{myquote}%
  {\list{}{\leftmargin=0.3in\rightmargin=0.1in}\item[]}%
  {\endlist}

\usepackage[accsupp]{axessibility}  %

\title{ColabSfM: Collaborative Structure-from-Motion by Point Cloud Registration}

\author{
\begin{tabular}{ccc}
Johan Edstedt$^{*}$$^{1}$ & Andr\'e Mateus$^2$ & Alberto Jaenal$^2$ \\[0.3em]
\end{tabular}
\\[0.3em]
{\normalsize $^1$Linköping University \quad $^2$Ericsson Research, Sweden}
}

\begin{document}

\twocolumn[{%
\centering
\renewcommand\twocolumn[1][]{#1}%
\maketitle
    \includegraphics[width=\linewidth]{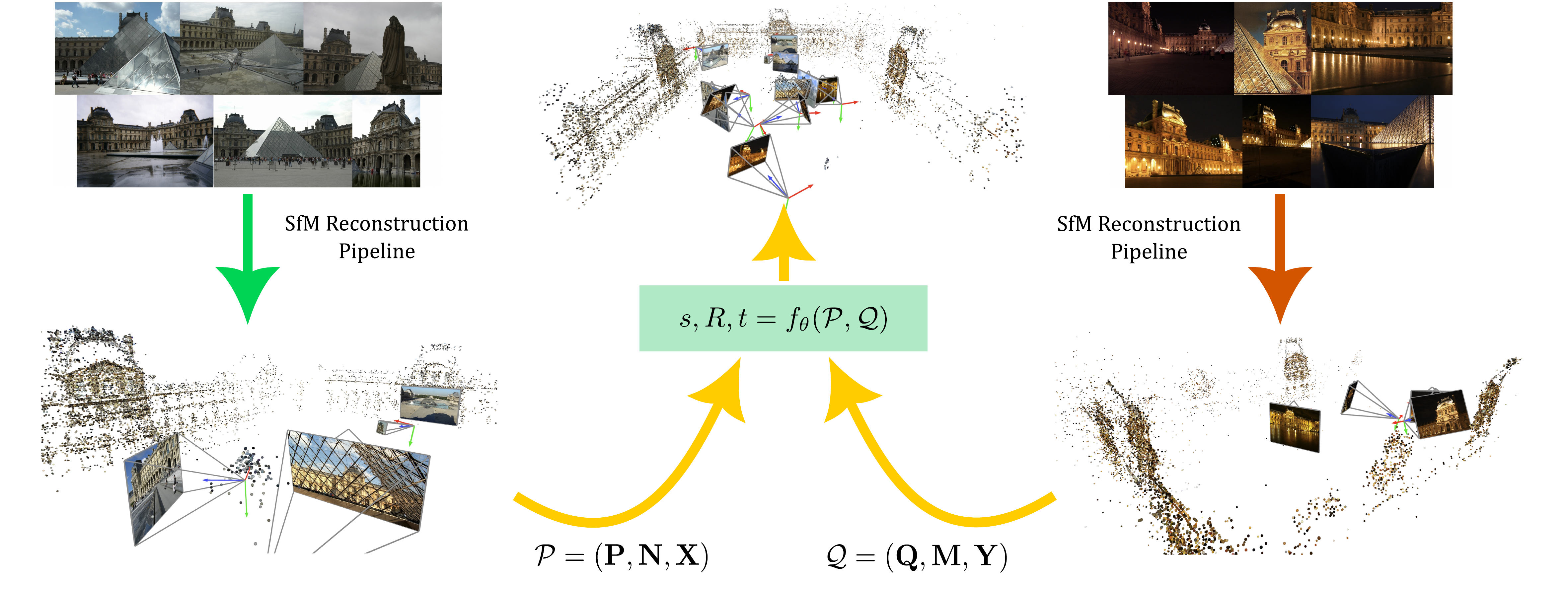}
    \captionof{figure}{\textbf{Our proposed registration paradigm for collaborative SfM reconstructions (ColabSfM).} Given two input SfM reconstructions $\mathcal{P}, \mathcal{Q}$ of the same scene, the task is to estimate the relative similarity transform $(s, R, t)$ between them. 
    Our first contribution is to address this as a point cloud registration problem, using \textbf{only} 3D SfM tracks.
    For this, we do not rely on the visual descriptors, but on the 3D coordinates of the points $\mathbf{P}, \mathbf{Q}$, their normals $\mathbf{N}, \mathbf{M}$ and, optionally, but not necessarily, features $\mathbf{X}, \mathbf{Y}$.
    To make point cloud registration methods perform well on this task, we propose as our second contribution a scalable pipeline to construct synthetic training datasets for SfM registration.
    Finally, we propose an improved version of RoITr~\cite{yu2023rotation} as registration method $f_\theta$.
    \vspace{2mm}
    }
    \label{fig:paradigm}}]
\renewcommand{\thefootnote}{\fnsymbol{footnote}}
\footnotetext[1]{Work done during an internship at Ericsson Research.}
\renewcommand{\thefootnote}{\arabic{footnote}}

\begin{abstract}

\vspace{-2mm}
Structure-from-Motion (SfM) is the task of estimating 3D structure and camera poses from images. We define Collaborative SfM (ColabSfM) as sharing distributed SfM reconstructions. Sharing maps requires estimating a joint reference frame, which is typically referred to as registration.
However, there is a lack of scalable methods and training datasets for registering SfM reconstructions.
In this paper, we tackle this challenge by proposing the scalable task of point cloud registration for SfM reconstructions. 
We find that current registration methods cannot register SfM point clouds when trained on existing datasets.
To this end, we propose a SfM registration dataset generation pipeline, leveraging partial reconstructions from synthetically generated camera trajectories for each scene. 
Finally, we propose a simple but impactful neural refiner on top of the SotA registration method RoITr that yields significant improvements, which we call \ours. 
Our extensive experimental evaluation shows that our proposed pipeline and model enables ColabSfM. Code is available at \url{https://github.com/EricssonResearch/ColabSfM}

\end{abstract}

\vspace{-2mm}
\section{Introduction}
\label{sec:intro}

\begin{figure*}
    \centering
    \begin{tabular}{cc}
     { OverlapPredator~\cite{huang2021predator}} & { \ours~(Ours)} \\ 
    \includegraphics[trim=0 0 2cm 0, clip,height=.18\textheight]{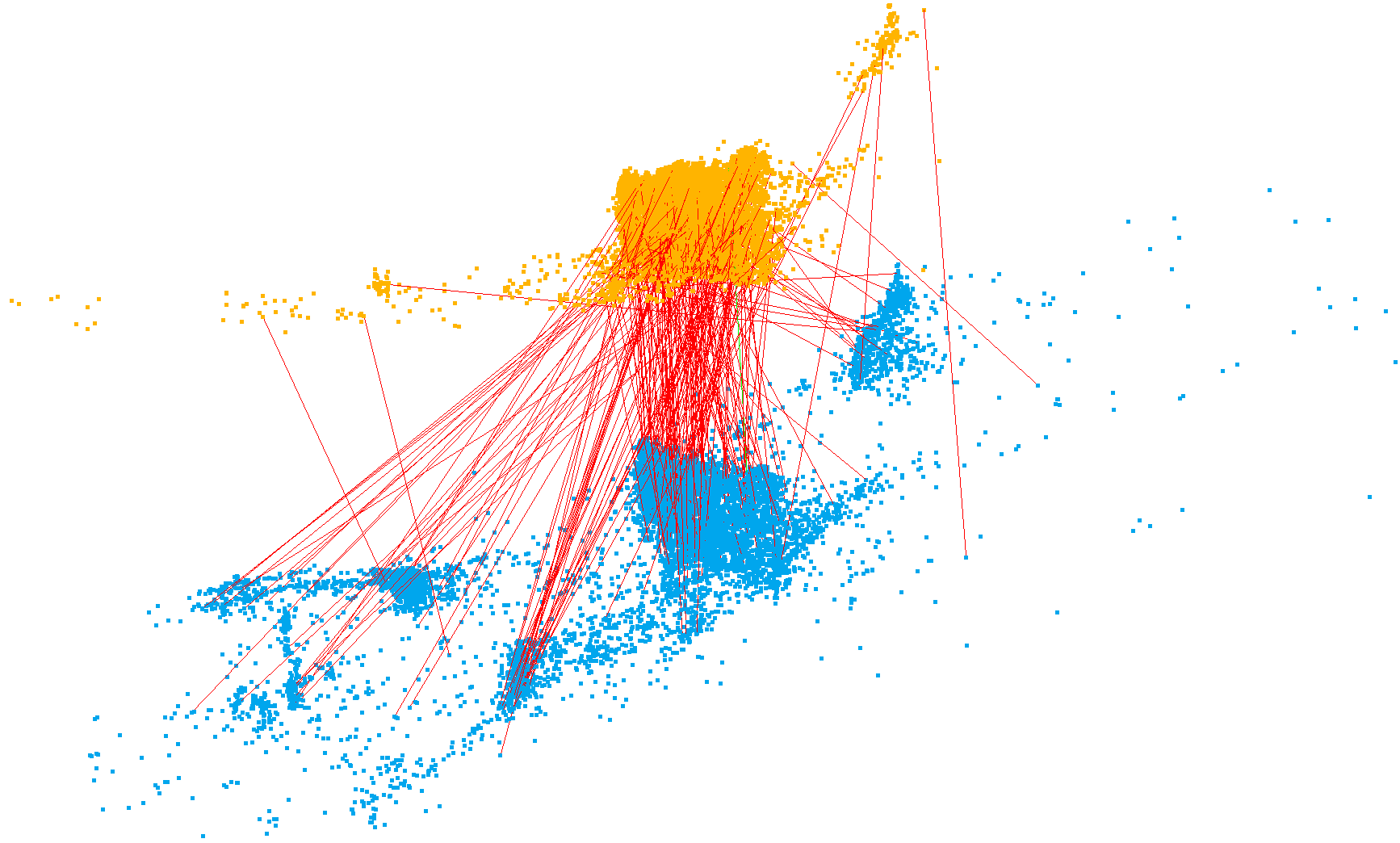} &
    \includegraphics[trim=0 0 4cm 0, clip,height=.18\textheight]{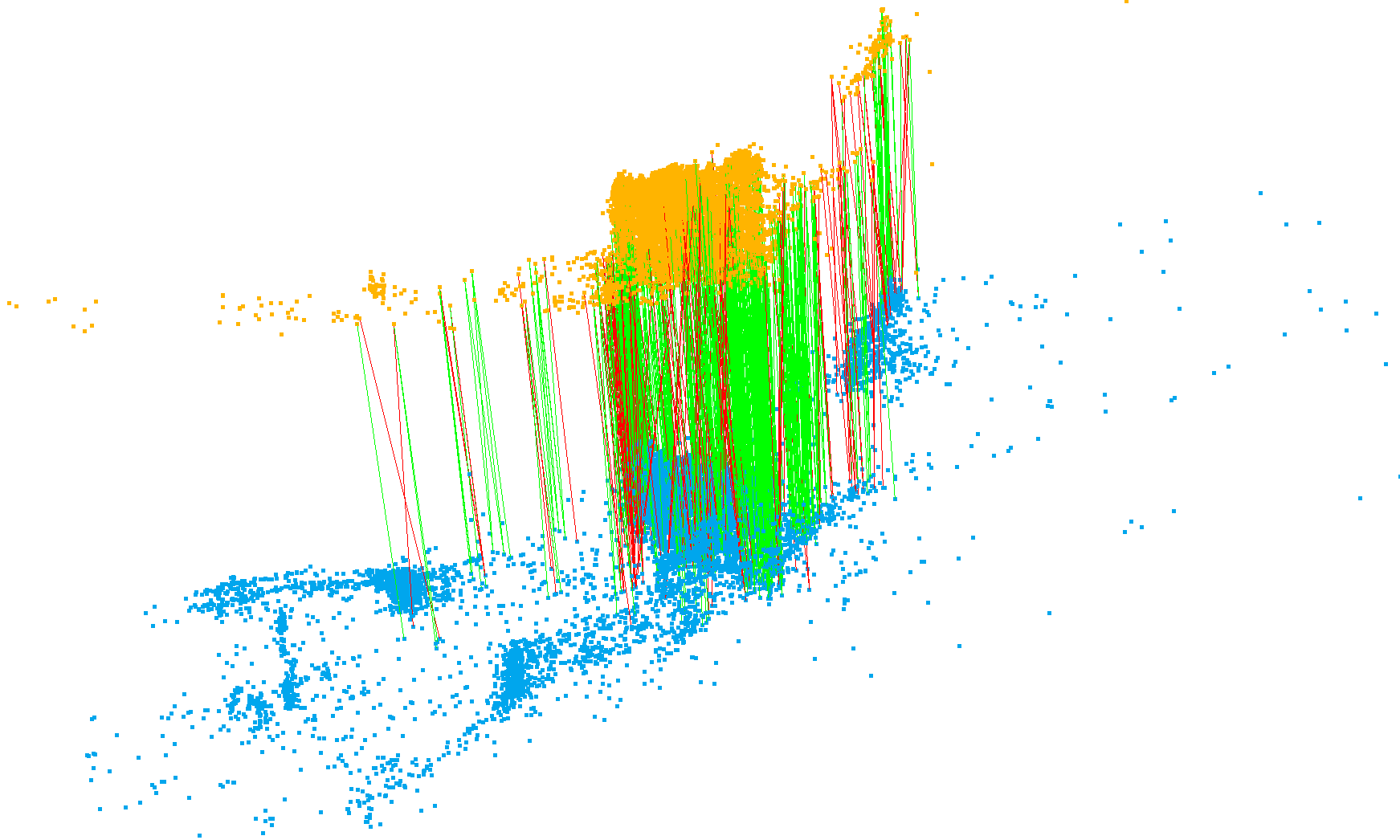} \\
    \midrule
    \includegraphics[trim=8cm 0 0cm 0, clip,height=.18\textheight]{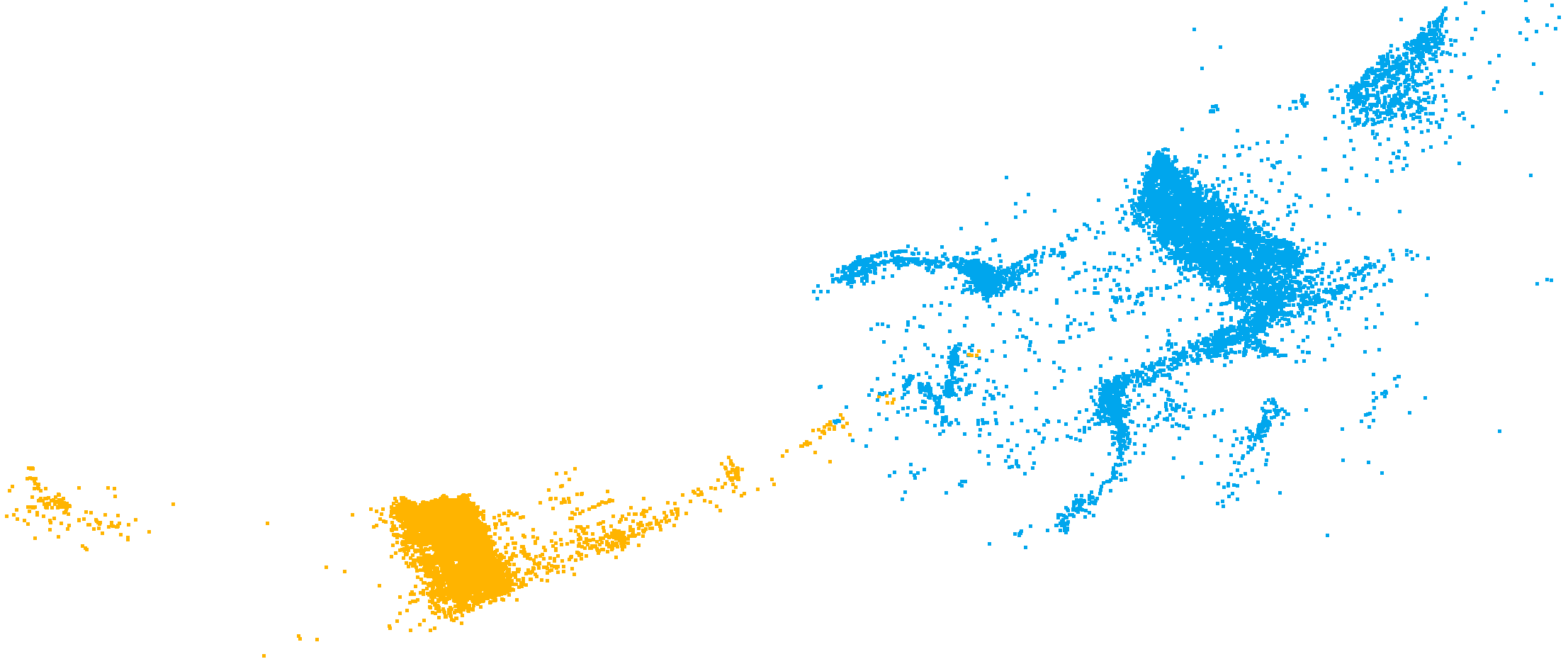} &
    \includegraphics[trim=0 0 4cm 0, clip,height=.18\textheight]{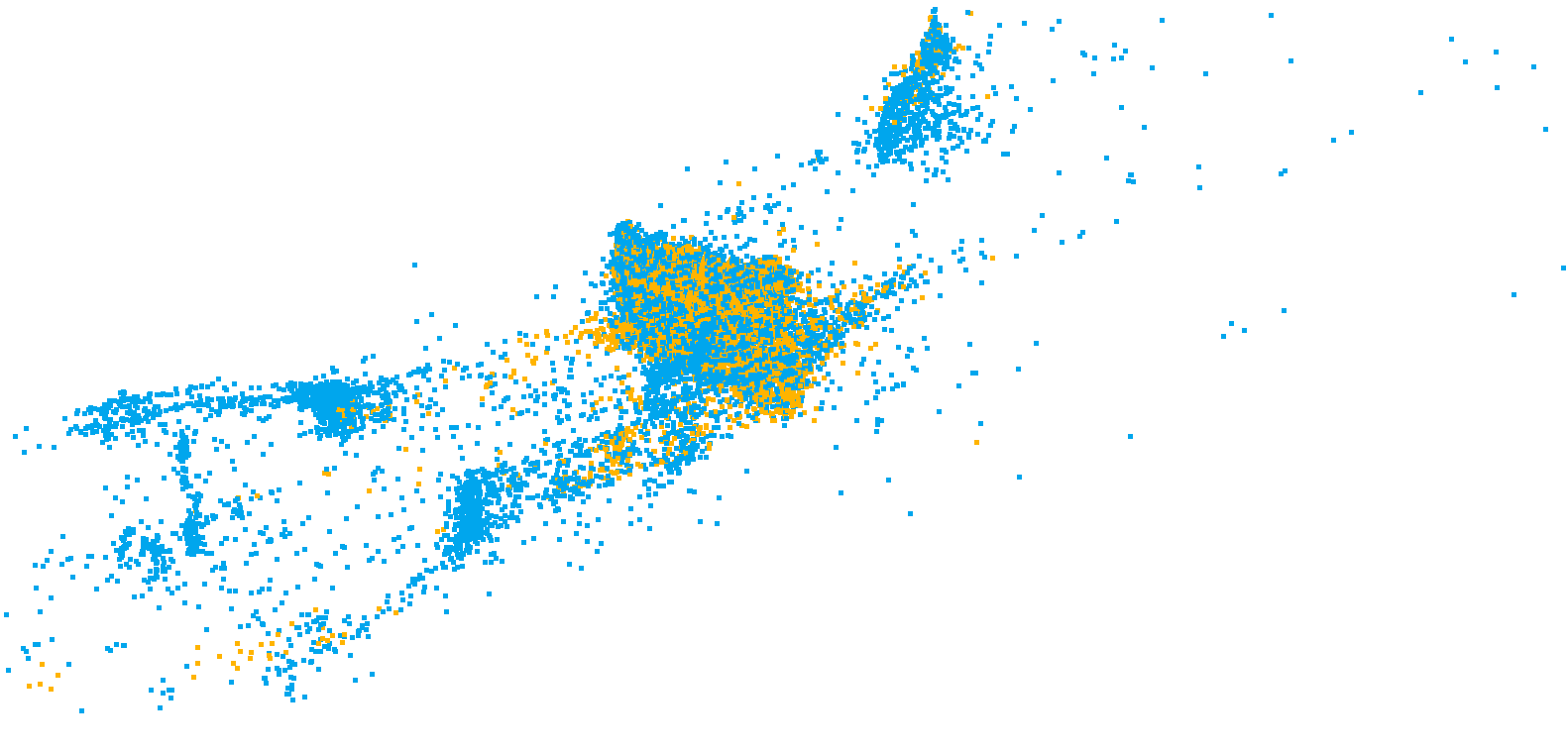} \\

    \end{tabular}
    \caption{\textbf{Qualitative comparison.} We compare our approach to the previous point cloud registration method OverlapPredator~\citep{huang2021predator} on the \emph{St.~Peter's Basilica} test scene. Without training on our proposed SfM registration dataset (column 1), previous methods are unable to produce sufficiently good matches (top row) and accurate relative pose estimation results (bottom row). In contrast, our proposed model \ours, trained on the proposed dataset, is able to find better matches and hence register the scenes well. The source and target point clouds are depicted in yellow and blue, respectively.}
    \label{fig:registration_results}
\end{figure*}

For robots and extended reality (XR) devices to become ubiquitous in our lives, they must be able to accurately build and localize in maps of the environment. 
Most available solutions rely on visual data and onboard computing, which can limit the accuracy in the context of small devices. 
Thus, in the last years there has been a push from industry to move computation to the cloud (e.g.  Google’s Visual Positioning System~\cite{googlevps}, Microsoft’s
Azure Spatial Anchors~\cite{azuresp}, and Niantic’s Lightship~\cite{nianticvps}).
However, the heterogeneous nature of these systems prevents devices of different vendors from interoperating, \ie, sharing their maps and localizing in others' maps.
If we are to make visual localization interoperable, we need ways to merge such maps.
Structure-from-Motion (SfM)~\cite{schoenberger2016sfm} is the most used method to process images and produce 3D maps (scene reconstructions consisting of a set of 3D points and the poses of the images expressed in a common coordinate system). 
To enable localization, a visual descriptor is stored for each map point, and for localization, 2D-3D matches are found by matching the descriptors of the query image with those of the map. 
Typically, SfM maps are merged using these visual descriptors, by finding 3D-3D matches~\cite{locher2018progressive, havlena2010efficient}, or inserting new sub-maps based on 2D-3D correspondences~\cite{sweeney2016large}.

However, these methods have limitations. %
First, different SfM pipelines tend to use different feature extractors, which yield incompatible descriptors. 
This is partially addressed by the cross-descriptor approach proposed in~\cite{dusmanu2021cross}, but the assumption that the descriptors are computed from the same keypoints is unrealistic.
Additionally, exposing visual descriptors may allow feature inversion attacks~\cite{pittaluga2019revealing,dosovitskiy2016inverting,weinzaepfel2011reconstructing,dangwal2021mitigating}, which can recover private elements from the images.
The second limitation concerns scalability, since storing visual descriptors leads to a map size increase between two and three orders of magnitude~\cite{zhou2022geometry}; and matching using a traditional pipeline~\cite{lowe2004distinctive} may suffer in performance; which can be enhanced by using learned methods~\cite{lindenberger2023lightglue,sarlin20superglue} at the cost of excessive runtimes.
Visual localization without descriptors was tackled in \cite{zhou2022geometry,wang2023dgc,zhang2025a2}, demonstrating competitive performance, but since they rely solely on 2D-3D geometry, which suffers from perspective distortions, their accuracy is not as high as their descriptor-based counterparts. Furthermore, they still rely on image retrieval, which again raises privacy, scalability, and interoperability concerns.
Nevertheless, their results show that geometry alone could be sufficient for matching.
Based on that observation, we pose the following question 
\begin{myquote}
\emph{Can we merge SfM maps using only 3D geometry?}
\end{myquote}
To this end, we propose to merge SfM 3D maps through 3D point cloud registration methods~\cite{zeng20173dmatch,aoki2019pointnetlk}, which rely solely on geometric information, see \cref{fig:paradigm}. We call our proposed paradigm Collaborative SfM (ColabSfM). In contrast to visual localization, which aims to localize a single image to a map, our proposed task involves registering a map to a map, and is image retrieval-free.

State of the art methods for 3D point cloud registration rely on learned features~\cite{yu2023rotation,huang2021predator,qin2022geometric}, therefore it is necessary to provide suitable datasets to train such networks.
There is a wide range of datasets available \cite{lopes2022survey} for the point cloud registration task.
However, those mostly consist of 3D sensor data, \eg, RGB-D~\cite{zeng20173dmatch} and/or LiDAR scans~\cite{geiger2012}, which differ from SfM point clouds. %
To the best of our knowledge, there are no large datasets for SfM point cloud registration. 
SfM datasets~\cite{li2018megadepth} provide a 3D point cloud for each scene, but since each scene is disconnected, no matching is possible between different scenes.
To tackle this issue, we propose a pipeline to generate synthetic datasets for SfM point cloud registration, which generates multiple synthetic trajectories for each scene.
The synthetic trajectories provide a trade-off between random internet images and sequential images (\eg, Cambridge Landmarks Dataset~\cite{kendall2015posenet}). We apply our pipeline and generate training and benchmarks datasets from MegaDepth~\cite{li2018megadepth} and a benchmark dataset from Quad6k~\cite{crandall2011discrete}. Our pipeline is presented in detail in~\cref{sec:dataset}.

SfM reference frames may be arbitrarily aligned relative to each other. As such, using a registration model invariant to rotation and translation SE(3), and robust to scale Sim(3), is important. To this end, we build on top of the recently proposed SotA rotation invariant model RoITr~\cite{yu2023rotation}. While this method already produces good results when trained on our dataset, we find that we can further improve results on ColabSfM by introducing a neural refinement stage to the model. We introduce our baseline model and our improved version, which we call \ours, in~\cref{sec:model}.

To summarize, our contributions are:
\begin{enumerate}
    \item A simple but powerful pipeline to synthetically generate SfM registration datasets, detailed in~\cref{sec:dataset}.
    \item An improved model for SfM Registration, detailed in \cref{sec:model}.
    \item The task of SfM point cloud registration, with extensive experiments in~\cref{sec:exp} demonstrating its potential.
\end{enumerate}

\section{Related Work}
\label{sec:related-work}

\parsection{Point cloud registration} is typically defined as estimating the rigid transform that aligns two 3D point clouds to a joint reference frame.
The classic approach is the Iterative Closest Point (ICP) algorithm~\citep{Chen1991,besl1992},
which estimates the pose by alternating between computing point correspondences using the current pose and estimating the pose using those correspondences.
Several improvements have been proposed over the years to handle outliers and to attempt to find a global minimum~\cite{Yang2013,Li2007}.
More recently, point cloud registration has been addressed with neural networks, such as DCP~\citep{wang2019dcp} and its iterative extension~\citep{wang2019prnet}, with similar methods presented in \cite{fu2021robust,yew2020rpm}.
Another class of iterative methods~\cite{aoki2019pointnetlk,xu2021omnet,huang2020feature}, extract global features from each point cloud, and exploits them to regress the pose. 

An alternative approach, which will be our main focus, consists of finding and extracting correspondences between the two point clouds, being those points \cite{Schonemann1966,miraldo19}, planes and lines~\cite{mateus2023fast}, and finding the pose using robust estimators such as RANSAC~\cite{fischler1981}.
Several point correspondence methods have been proposed, both handcrafted~\cite{Rusu2009,drost2010model} and learned~\cite{choy2019fully,gojcic2019perfect,deng2018ppfnet}.
In \cite{huang2021predator} an overlap-attention block is proposed to improve matching of point clouds with low overlap.
Overlap-aware methods are also presented in \cite{qin2022geometric,yew2022regtr, jin2024}, which focus on improving the matching accuracy and thus avoid the use of RANSAC to filter outliers.
The previous methods are not robust to large rotations, which are common in SfM.
This issue has been tackled by proposing methods that are rotation invariant~\cite{deng2018ppffold,yu2023rotation,bokman2022case}.
In this paper we will mainly consider point pair features (PPFs)~\cite{drost2010model} to ensure invariance to global rigid transformations, which have recently shown SotA performance~\cite{yu2023rotation}. %

\parsection{Localization without descriptors} has recently received renewed attention as a way to reduce computation, and preserve privacy.
BPnPNet~\cite{campbell2020solving} 
jointly estimates the pose and the correspondences between 2D keypoints on a query image and 3D points from a scene point cloud.
However, it is susceptible to outlier correspondences.
This issue was tackled by Zhou \etal~\cite{zhou2022geometry}, who use only 2D keypoints and the 3D map point cloud to localize a query image to a 3D map. 
They do this by projecting the map point cloud into retrieved map images and estimating 2D-2D correspondences with a graph neural network.
A similar method DGC-GNN~\cite{wang2023dgc} proposes to exploit geometry and color information to guide point matching by computing a geometric embedding of Euclidean and angular relations.
These methods have the advantage of not requiring storage of descriptors. 
However, they still require image retrieval %
to get sufficiently close map images to the query. This is problematic if the map is sparse in the region of the query, and introduces additional complexity. Furthermore, the projection from 3D into 2D loses important geometrical information, which we retain by directly registering the point clouds.

\parsection{Collaborative mapping} is a general term, typically referring to fusion of partial 3D reconstructions.
Cohen \etal~\cite{cohen2015merging} use the Manhattan world assumption~\cite{coughlan2000manhattan} to derive geometric constraints on partial reconstructions to enable model fusion. %
\citet{cohen2016indoor} fuse indoor and outdoor reconstructions by first leveraging images to obtain semantic masks with window positions. 
They then match the windows between the two models to solve for the pose with RANSAC~\cite{fischler1981}.
This approach requires images to be stored alongside the point clouds which brings additional space constraints~\cite{zhou2022geometry}.
Strecha \etal~\cite{strecha2010dynamic} and Untzelmann \etal~\cite{untzelmann2013scalable} merge multiple building reconstructions to a joint map by combining GPS coordinates and known building outlines to construct an alignment optimization objective. 
This has the downside of restricting the reconstructions to buildings that have previously been mapped and requires geo-tags.
Dusmanu \etal~\cite{dusmanu2021cross} investigated collaborative mapping and localization by translating descriptors to a joint embedding space. 
They consider a closed set of descriptors and learn a common embedding space for them. Collaborative mapping is achieved through embedding the descriptors into the joint space and running feature matching on the joint space. This requires access to all features for all images, which is expensive, and the required compute scales as standard SfM with the number of images.
In contrast to these methods, we make far fewer assumptions on the data available. Namely, we do not store any descriptors (neither local nor global), do not make assumptions on the structure of the scenes, and do not use any topological maps or data. Thus, we constrain ourselves to using only the geometry. %
Closest to our work~\citet{liu2021pluckernet} align \emph{line} reconstructions and simulate partial reconstructions by adding noise and dropout of lines to a single reconstruction.
In contrast we consider point-based reconstructions and propose a more realistic way to generate partial reconstructions through synthetic camera trajectories and retriangulation.

\section{Pipeline for SfM Registration Dataset Generation}
\label{sec:dataset}

\begin{figure*}[t]
    \centering
    \includegraphics[width=.8\linewidth]{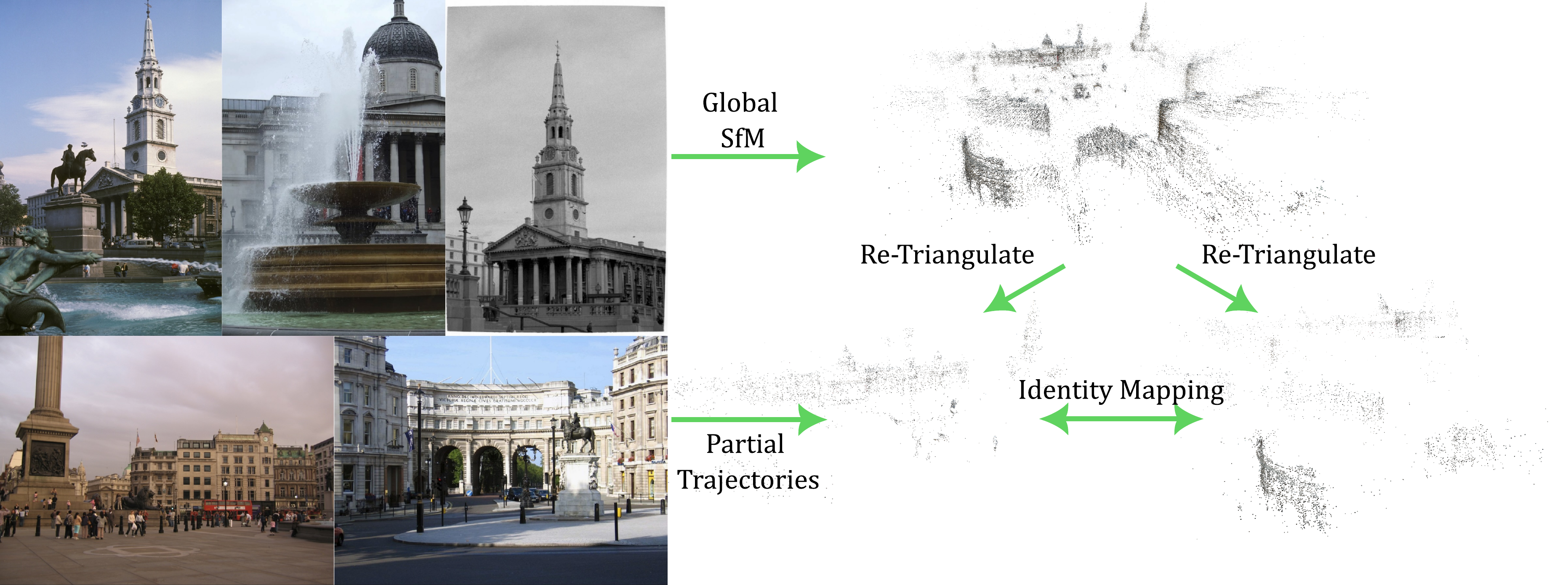}
    \caption{\textbf{Overview of our pipeline.} %
    For each scene, from a random image SfM dataset, \eg, MegaDepth~\cite{li2018megadepth}, we retriangulate partial reconstructions using partial trajectories from the scene. Since these trajectories are in the global SfM reference frame, the relative transformation is simply the identity mapping. %
    }
    \label{fig:dataset}
\end{figure*}

We find that current SotA point cloud registration models trained on 3DMatch~\cite{zeng20173dmatch}, the most commonly used training set for 3D registration, are not able to register SfM reconstruction out of the box (\cf \cref{tab:reg-sfm-bench,tab:cambridge,tab:quad6k}). Intuitively this is due to the large distribution shift between registration of RGB-D and/or LiDAR scans, and the much larger scenes covered by SfM reconstructions.
However, it is not obvious how to generate a diverse dataset with multiple aligned reconstructions. Manually collecting such datasets is prohibitively expensive, and publicly available datasets typically contain a single reconstruction per scene.

To tackle this issue, we propose a pipeline to generate a large synthetic dataset for registration between SfM models from random image collection datasets, \eg, MegaDepth~\cite{li2018megadepth}. 
Our key insight is that we can leverage these models and for each scene to construct partial reconstructions by sampling subsets of images/cameras.
An overview of our approach is presented in \cref{fig:dataset}.
In the next sections we describe our approach in more detail.

\subsection{Creating Partial Reconstructions}
\label{sec:create_partial_recon}
We use hloc~\cite{sarlin2019coarse} with COLMAP~\cite{schoenberger2016sfm} to create partial reconstructions from scenes. We use SIFT~\cite{lowe2004distinctive} and SOSNet~\cite{tian2019sosnet} retriangulations for training and evaluation. %
We consider two approaches to create partial reconstructions. 

\parsection{Random point sampling:}
In the simplest approach, we sample a set of 3D points from the scene for each partial reconstruction, and add the images in which they are visible until reaching approximately 200 images. %
For each scene we create 10 such partial reconstructions.
We found that while this approach works well when registering reconstructions from collections of random images, it does not generalize as well to reconstructions from a single camera trajectory, which are common in real-life scenarios, \eg, video. 
We believe that this is due to the point clouds exhibiting more significant variance when the scene is captured from different viewpoints. 
For example, keypoints are likely to be denser for nearby cameras, and natural occlusions will depend on viewpoints.
As random image based datasets are not naturally divisible into videos, it is not obvious how to bridge this gap. We next describe our approach addressing this.

\begin{figure}
    \centering
    \includegraphics[width=1\linewidth]{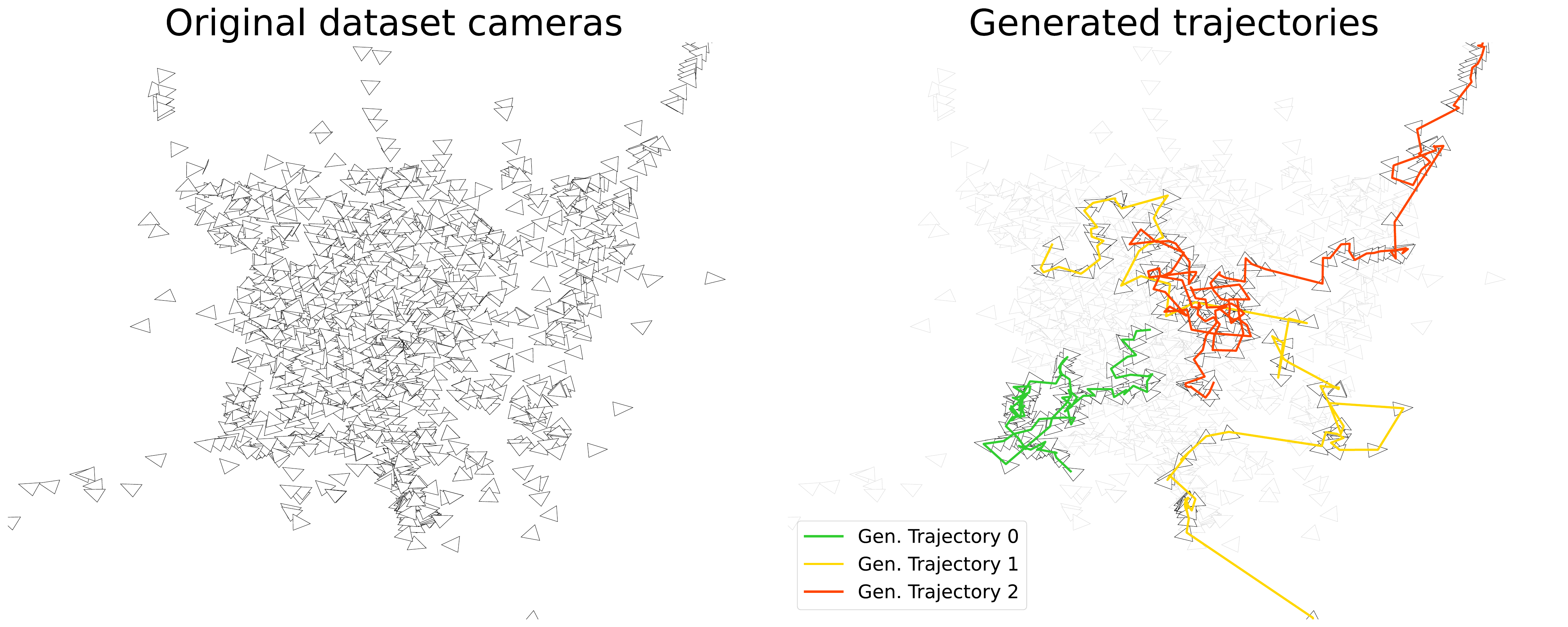}
    \vspace{-1em}
    \caption{\textbf{Example of synthetic trajectories in our proposed dataset}. We start with a large scale scene, consisting of hundreds of cameras (left). From this set of cameras we run \cref{alg:synthetic-trajectories} until the remaining set of cameras is small. Plotting the trajectories shows that plausible camera motion is achieved by this procedure (right). Sampling in this way bridges the gap between random image collections and video-based trajectories.}
    \label{fig:synthetic-trajectories}
\end{figure}

\parsection{Partial Trajectories:}
Our approach to bridge the random image $\to$ video gap is to generate synthetic trajectories from reconstructions of random images. 
Concretely, we randomly pick a starting image, and then sequentially (without replacement) pick the nearest neighbour of the image using a distance metric (which we refer to as $\text{dist}$). 
In practice we found that, for each trajectory, using a randomly weighted combination of geodesic rotation distance and Euclidean distance on the translation produced satisfactory results.%
Each trajectory has a size between $n_{\textup{low}}=75$ and $n_{\textup{high}}=300$.
The algorithm is presented in \cref{alg:synthetic-trajectories}. We show a qualitative example of three generated synthetic trajectories in \cref{fig:synthetic-trajectories}. 
Running this on all scenes gave us between 5 to 20 synthetic trajectories per scene.
Training is conducted on an equally weighted combination of the random point sampling approach and the synthetic trajectories.
\begin{algorithm}
  \caption{Generation of synthetic trajectory from a set of posed images
    \label{alg:synthetic-trajectories}}
  \begin{algorithmic}[1]
    \Require{$\mathbb{T}$ set of posed images, $\mathbb{I}$ remaining index set, distance function $\text{dist}$}
    \Statex
    \Function{GenerateTrajectory}{$\mathbb{T}, \mathbb{I}$}
      \Let{$\tau$}{$\emptyset$}
      \Let{i}{Random sample from $\mathbb{I}$}
      \Let{n}{Random sample from $\{n_{\text{low}},\hdots, n_{\text{high}}\}$}
      
      \Let{$\mathbb{I}$}{$\mathbb{I}\setminus i$}
      \Let{$\tau$}{$\tau \cup i$}
      \For{$k \gets 1 \textrm{ to } n$}
        \Let{$i'$}{$\argmin_{i'\in \mathbb{I}} \text{dist}(\mathbb{T}_i, \mathbb{T}_{i'})$}
        \Let{$\tau$}{$\tau \cup i'$}
        \Let{$\mathbb{I}$}{$\mathbb{I}\setminus i' $}
        \Let{$i$}{$i'$}
      \EndFor
      \State \Return{$\tau,\mathbb{I}$}
    \EndFunction
  \end{algorithmic}
\end{algorithm}

\subsection{Pipeline and Dataset}
\parsection{Reconstruction Pipeline:}
We are interested in having good ground truth correspondences between the partial reconstructions. As the SfM pipeline accuracy usually degrades with fewer available cameras, we use fixed camera poses from the joint reconstruction using all images and simply run  triangulation from feature matching. This also has the benefit of being significantly less computationally demanding than running the full reconstructions.

\parsection{Dataset Construction:}
We run our pipeline on all 196 MegaDepth scenes where 10 of them are reserved for the test benchmark. We use either SIFT~\cite{lowe2004distinctive} or SOSNet~\cite{tian2019sosnet} features for the retriangulation, however any local features could be used.
We compute the overlap between all pairs of partial reconstructions. To train models we enforce the overlap to be larger than 30\%. In total, after enforcing sufficient overlap, we are left with about 22000 pairs ($\sim$20000 for training and $\sim$2000 for testing), which constitutes our dataset. Details on how point cloud overlap is computed are presented in the supplementary material.

\begin{figure*}[t]
    \centering
    \includegraphics[width=.9\linewidth]{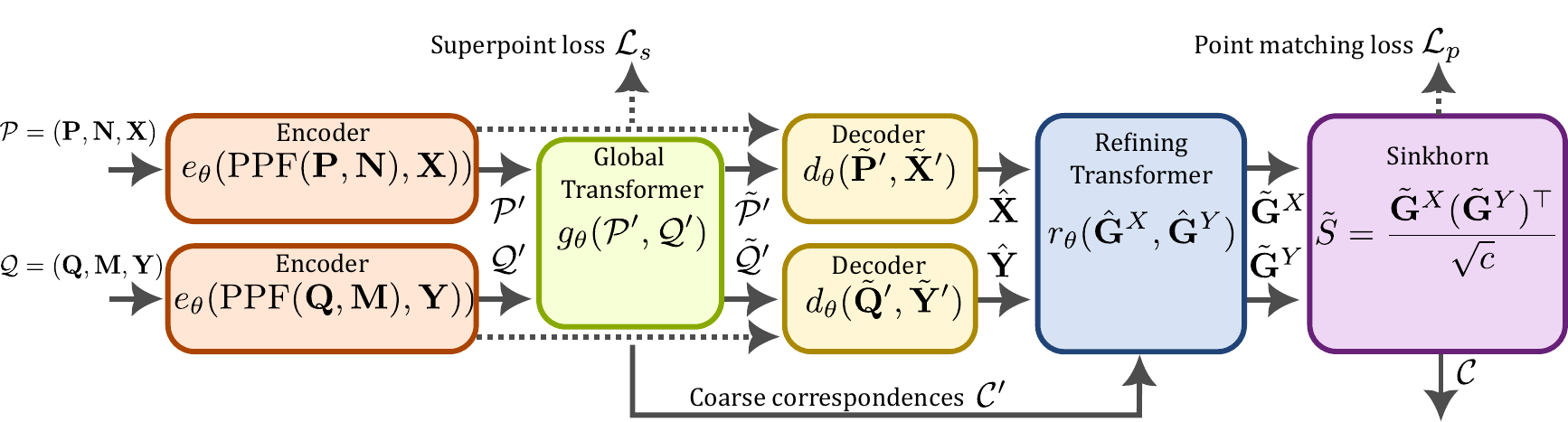}
    \caption{\textbf{Overview of our proposed model \ours.}  As input we take two point clouds $\mathcal{P} =(\mathbf{P},\mathbf{N}, \mathbf{X})$,
    $\mathcal{Q} =(\mathbf{Q},\mathbf{M}, \mathbf{Y})$, consisting of 3D points $\mathbf{P}/\mathbf{Q}\in\mathbb{R}^{n/m\times 3}$, and normals $\mathbf{N}/\mathbf{M}\in\mathbb{R}^{n/m\times 3}$. 
    In this work the features $\mathbf{X}/\mathbf{Y}$ are always assumed to be constant.
    We ensure rotation invariance by Point Pair Feature (PPF) encoding. The PPFs are fed together with the features into an encoder $e_{\theta}$, producing $N'$ coarse superpoints $\mathcal{P}'/\mathcal{Q}'$. These are passed through a global Transformer $g_{\theta}$, from which the top-$k$ coarse correspondences $\mathcal{C}'\in \mathbb{R}^{k\times 2}$ are extracted. A decoder $d_{\theta}$ takes the coarse features from $g_{\theta}$ and the finer features from $e_{\theta}$ and produces fine features $\hat{\mathbf{X}},\hat{\mathbf{Y}}$. Using the coarse correspondences we extract neighbourhoods $\hat{\mathbf{G}}^X,\hat{\mathbf{G}}^Y \in \mathbb{R}^{k \times 64 \times c}$, which we feed into our proposed refinement Transformer. The refined features from the Transformer are used to construct a cost matrix, where the Sinkhorn algorithm~\cite{untzelmann2013scalable, sarlin20superglue, yu2023rotation} is used to solve the optimal transport (OT) problem, producing our final matches $\mathcal{C}$. The network is trained using $\mathcal{L}_s + \mathcal{L}_p$.
}
    \label{fig:model}
\end{figure*}

\section{RoITr and \ours~for ColabSfM}
\label{sec:model}

In this section, we discuss our baseline and our proposed improvements for ColabSfM. An overview of the proposed registration model $f_{\theta}$ is presented in \Cref{fig:model}. For consistency we follow the notation of Yu \etal~\cite{yu2023rotation}.

\subsection{Preliminaries}
We are interested in registration methods that are robust to the challenges posed by SfM reconstructions. In particular, one of the main challenges is that SfM reconstructions are defined up-to-scale, and the rotation and origin of the reference frame is arbitrary\footnote{Although in practice, approximate scale can be obtained by, \eg, standardization, objects with known size, or additional sensors.}. Hence, we would like the registration model to be approximately invariant to the action of the similarity group, \ie, $\text{Sim(3)} = \text{SE(3)} + \text{scale}$. In practice we will achieve this by using an $\text{SE(3)}$ invariant model, and learn approximate scale invariance. We next describe our baseline, and then our proposed improvements.

\subsection{Baseline Model}
As a strong baseline, we base our approach on the SotA SE(3) invariant registration method RoITr~\cite{yu2023rotation}. 
We next give an overview of our baseline. 
An more thorough description of the architecture is given in the supplementary and by~\citet{yu2023rotation}.

\parsection{Rotation Invariance through PPFs:}
 RoITr~\cite{yu2023rotation} is based on point pair features (PPFs)~\cite{drost2010model}, which computes a 4D feature for point pairs by the angles between their relative position and their normals. Mathematically, the PPF is defined as
$\text{PPF}(p_1, p_2, n_1, n_2) = [\lVert d \rVert, \angle(n_1, d), \angle(n_2, d), \angle(n_1, n_2)]$,
where $d = p_2 - p_1$. 
It can easily be shown that this construction is invariant to rotations,
\begin{multline}
    \text{PPF}(Rp_1, Rp_2, Rn_1, Rn_2) = [\lVert Rd \rVert, \angle(Rn_1, Rd), \\ \quad \angle(Rn_2, Rd), \angle(Rn_1, Rn_2)]
    = [\lVert d \rVert, \angle(n_1, d), \angle(n_2, d), \\ \quad \angle(n_1, n_2)] = \text{PPF}(p_1, p_2, n_1, n_2),
\end{multline}
since norms and angles are invariant to rotations.

\parsection{Normalization Procedure:}
\label{sec:normalization}
SfM reconstructions do not in general come with metric scale. This poses several challenges for registration methods, that typically assume both point clouds to share the same scale. We tackle this by normalization. Specifically, for Sim(3) training we normalize the point clouds $\mathbf{P}, \mathbf{Q}$ with $\sigma_{\mathbf{P}}/\sqrt{2}, \sigma_{\mathbf{Q}}/\sqrt{2}$, where $\sigma$ is the largest singular value of the centered point cloud. For SE(3) training, we simply use $\sigma_{\mathbf{P}}$ to normalize both $\mathbf{P}$ and $\mathbf{Q}$. We find that this simple approach leads to models robust to scale variation, as we demonstrate in \cref{tab:reg-sfm-bench}.

\parsection{Normal Estimation:}
We follow RoITr~\cite{yu2023rotation} and use Open3D~\cite{Zhou2018open3d} with a neighborhood size of 33 for estimating normals, we found that this works well in practice.
In contrast to the RGB-D scans of 3DMatch, our dataset consists of point clouds from multiple sensors. This means that the simple approach of aligning the normal orientation towards the origin of the coordinate system, as done in RoITr, will not be consistent in general. To remedy this, we use the fact that all cameras observing a 3D track will yield consistent orientation estimates. In practice, we align the orientation towards the camera center of a random image observing the track, which we found yields consistent orientation. 

\parsection{Overview of Architecture:} 
RoITr consists of an encoder $e_{\theta}$, a global Transformer $g_{\theta}$, and a decoder $d_{\theta}$. $e_{\theta}$ and $d_{\theta}$ take as input sets of PPF encoded local neighbourhoods (typically around $8/16$ points) around each point/superpoint. In contrast, $g_{\theta}$ performs global attention on the coarse point cloud $\hat{\mathcal{P}}$. $e_{\theta}, d_{\theta}, g_{\theta}$ all use the Transformer architectures as described by Yu \etal~\cite{yu2023rotation}. Note that RoITr does not include the refining Transformer $r_{\theta}$, which will be detailed in our \ours.

\parsection{Losses:}
RoITr uses a superpoint matching loss $\mathcal{L}_s$ and a point matching $\mathcal{L}_p$ loss for training. We follow RoITr and use their overlap-aware~\cite{qin2022geometric} circle loss~\cite{sun2020circle} as $\mathcal{L}_s$. The point matching loss $\mathcal{L}_p$ is the Negative Log Likelihood (NLL) of the GT correspondences after running the Sinkhorn algorithm~\cite{sarlin20superglue} on the fine feature similarities $\tilde{S}$.

\subsection{\ours~(Ours)}
\label{sec:improvements}
\parsection{Refinement Transformer $r_{\theta}$:}
The refinement in RoITr is performed by optimizing the Sinkhorn~\cite{untzelmann2013scalable} distance between shallow features in a local neighbourhood around the coarse matches, denoted $\hat{\mathbf{G}}^X,\hat{\mathbf{G}}^Y$.
While this provides satisfactory results on 3DMatch, we found that we could improve results by incorporating a joint local Transformer $r_{\theta}$. This Transformer takes $\hat{\mathbf{G}}^X,\hat{\mathbf{G}}^Y$ as inputs and outputs enriched neighbourhoods $\tilde{\mathbf{G}}^X,\tilde{\mathbf{G}}^Y$. We found that these cross-point cloud enriched fine features from $r_{\theta}$ yielded improved results.

\parsection{Architectural Details of $r_{\theta}$:}
$r_{\theta}$ takes in local neighbourhoods of shape 
$m \times 64 \times c$, where $m$ is the number of neighbourhoods, $64$ is the number of neighbours (identical to RoITr), and $c$ is the feature dimensionality. We use layers of interleaved self-attention and cross-attention as in LightGlue~\citep{lindenberger2023lightglue}. Note however, that in contrast to LightGlue that performs global attention, our refinement Transformer performs attention locally, which is significantly cheaper. We use an input/output dimensionality of $c_{\rm in} = c_{\rm out} = c = 64$, and 4 self+cross attention layers. Each self/cross attention layer consists of single head attention, using dim $c$, which is stacked with the input into a 1 hidden layer MLP (input dim $2c$, hidden dim $2c$, output dim $c$) using the GELU~\cite{hendrycks2016gaussian} activation function, with layer normalization~\cite{ba2016layer} applied before GELU\@. Each self/cross block is residual, \ie, the output of the block is the output of the MLP plus the input.

\vfill

\begin{table}
\centering
    \caption{\textbf{Results of SfM registration on MegaDepth}. We evaluate 2 different scenarios. Unknown relative orientation SO(3) + unknown relative position = SE(3), SE(3) + unknown relative scale = Sim(3).
    The top portion contains methods only trained on the 3DMatch (3DM) dataset, the middle portion methods trained only on our proposed dataset (Mega), while the lower portion contains methods trained on the former and fine-tuned on the latter.}
    \resizebox{1\linewidth}{!}{\begin{tabular}{l ccc  ccc}
        \toprule
        & \multicolumn{3}{c}{SE(3)} & \multicolumn{3}{c}{Sim(3)} \\
        \cmidrule(lr){2-4} \cmidrule(lr){5-7} 
        Method & IR & FMR & RR & IR & FMR & RR \\
        \midrule
         OverlapPredator~\cite{huang2021predator} (3DM) & 6.1 & 35.5 & 10.0 & 3.6 & 21.3 & 2.1\\
         GeoTransformer~\cite{qin2022geometric} (3DM) & 2.1 & 14.6 & 3.3 & 1.3 & 8.6 & 0.4 \\
         PareNet~\cite{yao2025pare} (3DM) & 7.3 & 38.6 & 5.2 & 4.7 & 27.7 & 0.8 \\
         RoITr~\cite{yu2023rotation} (3DM) & 3.0 & 12.6 & 0.0 & 1.6 & 7.0 & 0.8 \\
         \ours~(3DM) & 10.0 & 32.8 & 13.9 & 6.2 &  22.8 & 3.5 \\
         \midrule
         \ours~(Mega) & \textbf{51.0} & \textbf{96.5} & \textbf{70.2} & \textbf{44.6} & \underline{92.8} & \underline{42.7} \\ 
         \midrule
         OverlapPredator~\cite{huang2021predator} (3DM + Mega) & 21.3 & 74.9 & 35.4 & 11.4 & 56.2 & 10.7 \\
         RoITr~\cite{yu2023rotation} (3DM + Mega) & 44.6 & 90.9 & 60.0 & 38.4 & 86.1 & 38.7 \\
         \ours~(3DM + Mega) & \underline{48.7} & \underline{95.1} & \underline{67.7} & \textbf{44.6} & \textbf{93.9} & \textbf{44.3} \\
         \bottomrule
    \end{tabular}
    }
    \label{tab:reg-sfm-bench}
\end{table}

\begin{table*}
\centering
    \caption{\textbf{Results of SfM registration on Cambridge Landmarks~\cite{kendall2015posenet}}. We evaluate unknown relative orientation SO(3) + unknown relative position = SE(3).
    The top portion contains methods only trained on the 3DMatch (3DM) dataset, the middle portion methods trained only on our proposed dataset (Mega), while the lower portion contains methods trained on the former and fine-tuned on the latter.}
    \resizebox{.7\linewidth}{!}{\begin{tabular}{l cc cc cc cc cc}
        \toprule
        & \multicolumn{2}{c}{Great Court} & \multicolumn{2}{c}{Kings College}& \multicolumn{2}{c}{Old Hospital} & \multicolumn{2}{c}{Shop Facade} &  \multicolumn{2}{c}{St Mary's Church}\\
        \cmidrule(lr){2-3} \cmidrule(lr){4-5} \cmidrule(lr){6-7} \cmidrule(lr){8-9} \cmidrule(lr){10-11}
        Method & IR & Matches & IR & Matches & IR & Matches & IR & Matches & IR & Matches \\
        \midrule
         OverlapPredator~\cite{huang2021predator} (3DM) & 1.4 & 356 & 0.5 & 372 & 0.0 & 363 & 2.5 & 354 & 0.2 & 382\\
         GeoTransformer~\cite{qin2022geometric} (3DM) & 0.0 & 382 & 0.1 & 785 & 0.0 & 438 & 0.1 & 1368 & 0.0 & 431\\
         PareNet~\cite{yao2025pare} (3DM) & 4.3 & 2000 & 1.6 & 2000 & 0 & 2000 & 1.2 & 2000 & 6.7 & 2000\\
         RoITr~\cite{yu2023rotation} (3DM) & 0.0 & 285 & 0.0 & 447 & 0.0 & 321 & 0.3 & 959 & 1.5 & 407\\
         \ours~(3DM) & 0.2 & 412 & 2.2 & 408 & 0.0 & 544 & 0.9 & 737 & 2.8 & 567\\
         \midrule
         \ours~(Mega)& \textbf{70.9} & \underline{4538} & \textbf{57.6} & \textbf{3977} & \textbf{31.5} & \textbf{2214} & \textbf{41.6} & \underline{1325} & \textbf{81.8} & \underline{5167}\\
         \midrule
         RoITr~\cite{yu2023rotation}  (3DM + Mega) & 52.1 & 1426 & 39.6 & 1377 & 21.9 & 603 & 28.0 & 700 & 64.5 & 2481\\
         \ours~(3DM + Mega)& \underline{69.4} & \textbf{4551} & \underline{54.0} & \underline{3246} & \underline{28.0} & \underline{1591} & \underline{39.1} & \textbf{1342} & \underline{77.8} & \textbf{5168}\\
         \bottomrule
    \end{tabular}}
    \label{tab:cambridge}
\end{table*}

\vspace{-1em}
\section{Experiments}
\label{sec:exp}

In this section, we evaluate the \ours~model presented in \cref{sec:model}.
Three different model instances were trained: 1) pre-trained \ours~on 3DMatch~\cite{zeng20173dmatch}; 2) the model in 1) fine-tuned on the dataset in \cref{sec:dataset} and; 3) \ours~trained from scratch on our dataset.
For comparison, we used RoITr~\cite{yu2023rotation}%
 and OverlapPredator~\cite{huang2021predator}
, both with pre-trained weights from 3DMatch, and fine-tuned weights trained on our dataset.
Additionally, we tested GeoTransformer~\cite{qin2022geometric}%
 also pre-trained on 3DMatch~\cite{zeng20173dmatch}. 
As we found that the performance, of fine-tuning 3DMatch trained models, on our dataset was similar for our proposed model, we did not train other models in our dataset from scratch.

\parsection{Metrics:}
We use  the standard registration metric \emph{Inlier Ratio} (IR). This measures the percentage of estimated correspondences lying within a threshold distance $\tau_{\text{IR}} = 0.1$ from each other. 
We additionally measure the \emph{Feature Matching Recall} (FMR), which is defined as the percentage of point cloud pairs whose IR is above a threshold $\tau_{\text{FMR}} = 5\%$.
Finally, we use \emph{Registration Recall} (RR), which consists of the percentage of pairs for which the rotation error and translation error are below $5^\circ$ and $0.05$, respectively.
We solve for the pose using the RANSAC~\cite{fischler1981} implementation from Open3D~\cite{Zhou2018open3d} on the matches retrieved by the neural networks. 
Details on error computation and RANSAC can be found in the Supplemental material.

\parsection{Computational cost:}
On a desktop machine equipped with an i9-13900K @ 3.00GHz CPU and an NVIDIA 4090 GPU, the median runtime for RoITr is 99.94 ms and for \ours is 102.84, which shows that the computational burden added by the modifications described in \cref{sec:improvements} is around 3\%.

\subsection{MegaDepth SfM Registration Benchmark}
\label{sec:sfm_bench}

We select 10 scenes from MegaDepth~\cite{li2018megadepth} dataset, which consists of 196 large scale 3D scenes reconstructed from internet images for our benchmark. The test scenes are, in order, [0008, 0015, 0019, 0021, 0022, 0024, 0025, 0032, 0063, 1589].
The average size of the pointclouds is $15.3$K points, using either SIFT or SOSNet features for the retriangulation.
To train and evaluate the network presented in \cref{sec:model} on our dataset, we define two different scenarios SE(3) and Sim(3).
We define a random rotation by sampling three Euler angles $\alpha, \beta, \gamma \in [0,2\pi]$.
\begin{table*}
\centering
    \caption{\textbf{Results of SfM registration on 7-Scenes}. We evaluate unknown relative orientation SO(3) + unknown relative position = SE(3). The top portion contains methods only trained on the 3DMatch (3DM) dataset, the middle portion methods trained only on our proposed dataset (Mega), while the lower portion contains methods trained on the former and fine-tuned on the latter.}
    \resizebox{.85\linewidth}{!}{\begin{tabular}{l cc cc cc cc cc cc cc}
        \toprule
        & \multicolumn{2}{c}{Chess} & \multicolumn{2}{c}{Fire}& \multicolumn{2}{c}{Heads} & \multicolumn{2}{c}{Office} &  \multicolumn{2}{c}{Pumpkin} & \multicolumn{2}{c}{Red Kitchen} & \multicolumn{2}{c}{Stairs}\\
        \cmidrule(lr){2-3} \cmidrule(lr){4-5} \cmidrule(lr){6-7} \cmidrule(lr){8-9} \cmidrule(lr){10-11} \cmidrule(lr){12-13} \cmidrule(lr){14-15}
        Method & IR & Matches & IR & Matches & IR & Matches & IR & Matches & IR & Matches & IR & Matches & IR & Matches\\
        \midrule
         RoITr~\cite{yu2023rotation} (3DM) & 0.0 & 307 & 0.0 & 1004 & 0.0 & 1699 & 0.0 & 1322 & 0.0 & 327 & 0.0 & 1163 & 0.0 & 730\\
         \ours~(3DM) &0.0&842&0.0&2204&0.0&1104&0.0&1073&0.0&1016&0.0&1721&0.0&1043\\

        \midrule
        
        \ours~(Mega) & \underline{23.4} & \textbf{6717} & \textbf{38.8} & \textbf{9043} & \textbf{50.6} & \underline{5419} & \textbf{68.0} & \textbf{9461} & \textbf{14.9} & \textbf{2893} & \textbf{27.9} & \textbf{8050} & 6.2 & \textbf{3230} \\
        \midrule
        RoITr~\citep{yu2023rotation} (3DM+Mega) & 20.4 & 2098 & 30.5 & 4165 & 36.4 & 3812 & 50.0 & 4752 & 7.9 & 1271 & 18.5 & 2438 & \textbf{9.1} & 1883 \\
        \ours~(3DM+Mega) & \textbf{23.6} & \underline{5632} & \underline{33.3} & \underline{8339} & \underline{50.5} & \textbf{5996} & \underline{62.7} & \underline{8887} & \underline{8.9} & \underline{2268} & \underline{26.8} & \underline{6616} & \underline{6.4} & \underline{2863}\\
         \bottomrule
    \end{tabular}}
    \label{tab:7scenes}
\end{table*}

The results of the \ours~method and the baselines are presented in \cref{tab:reg-sfm-bench}.
As expected, the methods which are not fine-tuned in our data perform the worst in both SE(3) and Sim(3) data. 
This shows the lack of generalization of current registration methods on SfM point clouds.
When considering FMR, the methods trained from scratch or fine-tuned on our data obtain high scores ($>$90$\%$) on SE(3) data with slightly lower scores on Sim(3). 
When looking at the three variations of \ours, we see that while the version trained on 3DMatch outperforms vanilla RoITr, the major boost in performance is due to fine-tuning the model on our dataset.
The best overall version is the one trained from scratch on our dataset, even though the improvement is not as significant.
Qualitative results are presented in \cref{fig:registration_results} and the Supplementary material.

\begin{table}
\centering
    \caption{\textbf{Results of SfM registration on Quad6k~\cite{crandall2011discrete}}. We evaluate unknown relative orientation SO(3) + unknown relative position = SE(3).
    The top portion contains methods only trained on the 3DMatch (3DM) dataset, the middle portion methods trained only on our proposed dataset (Mega), while the lower portion contains methods trained on the former and fine-tuned on the latter.}
    \resizebox{0.75\columnwidth}{!}{%
        \begin{tabular}{l ccccc }
            \toprule
            Method & IR & FMR & Matches & RR  \\
            \midrule
             OverlapPredator~\cite{huang2021predator} (3DM) & 1.6 & 10.2 & 121 & 0.0\\
             GeoTransformer~\cite{qin2022geometric} (3DM) & 0.6 & 0.0 & \textbf{1936} & 0.0 \\
             PareNet~\cite{yao2025pare} (3DM) & 1.6 & 10.4 & 2000 & 0.0 \\
             RoITr~\cite{yu2023rotation} (3DM) & 0.4 & 0.0 & 574 & 0.0 \\
             \ours~(3DM) & 1.7 & 6.1 & 617 & 2.0 \\
             \midrule
             \ours~(Mega) & \underline{17.2} & \textbf{67.4} & \underline{878} & \underline{22.4} \\
             \midrule
             RoITr~\cite{yu2023rotation} (3DM + Mega) & 14.2 & 63.3 & 492 & \textbf{24.5} \\
             \ours~(3DM + Mega)& \textbf{18.1} & \underline{65.3} & 715 & 16.3   \\
             \bottomrule
        \end{tabular}
        }
    \label{tab:quad6k}
\end{table}

\subsection{Standard Benchmark: Cambridge Landmarks}
The Cambridge Landmarks~\cite{kendall2015posenet} dataset consists of 5 large-scale 3D scenes recorded with a smartphone by a pedestrian in Cambridge, each of which consisting of a train and a test trajectories with camera poses. We retriangulate the train/test sequences independently, which results in two reconstructions for each scene. 
In this case, we evaluate the IR and number of matches. Here we only test the SE(3) case.
Rotation and translation errors and qualitative results for this dataset are reported in the supplementary material.
The results are presented in \cref{tab:cambridge}.
Similar to the evaluation results in the previous section, the models trained only on 3DMatch~\cite{zeng20173dmatch} perform poorly, while the ones fine-tuned and or trained from scratch on our dataset present the highest IR and number of matches, with small differences in favor of the version trained from scratch.

\subsection{Noisy Indoor Reconstructions: 7-Scenes}
The 7-Scenes~\citep{shotton2013scene} dataset is an indoor RGB-D dataset. This dataset is challenging for SfM, since untextured regions lead to significantly more noisy point clouds, and less geometric features. We present results in~\cref{tab:7scenes}. Encouragingly, we find that models trained using our pipeline perform well also on these indoor scenes, showing the generalizability of our approach.

\subsection{Low Overlap: Quad6k}
The Quad6k dataset~\cite{crandall2011discrete} consists of about 6500 images from the Arts Quad at Cornell University. To create partial reconstructions for testing, we first perform a full scene re-triangulation using the refined poses provided by~\cite{Turki_2022_CVPR} and then follow the procedure in \cref{sec:create_partial_recon}, generating a total of 24 trajectories, and 49 pairs.
The results are presented in \cref{tab:quad6k}. 
Similar to the previous datasets, the performance of the methods  has a substantial boost after training on the dataset presented in \cref{sec:dataset}.
GeoTransformer~\cite{qin2022geometric} trained only on 3DMatch~\cite{zeng20173dmatch} presents the most matches on average, but only a small fraction ($0.6\%$) are inliers, and its FMR and RR are both zero. Our Quad6k benchmark is more challenging and thus the performance is lower than in \cref{tab:reg-sfm-bench,tab:cambridge}, due to a low overlap of 16.5\%, in comparison with the 61.2\% of MegaDepth.
Qualitative results for this dataset are shown in the supplementary material.

\section{Conclusion}
\label{sec:conclusion}
We have proposed Collaborative Structure-from-Motion by Point Cloud Registration, ColabSfM, a challenging new task to enable scalable collaborative mapping and localization. By extensive experiments we showed that with our proposed dataset generation pipeline and SE(3) invariant model, this task is solvable, and showed that our model, while trained only on MegaDepth, is able to generalize to several challenging scenarios, such as noisy indoor reconstructions, and low-overlap registration. We believe our initial results will encourage further research on collaborative 3D reconstruction. \textbf{Limitations:}
a) Our proposed model \ours~is E(3) invariant like RoITr, and therefore can struggle with estimating correct matches in symmetric scenes.
b) Our models are trained and evaluated on retriangulated partial reconstructions. In practice, issues such as drift occur, making global alignment more difficult, however we find that our model still performs well in this setting in~\Cref{tab:cambridge-scratch}.

\section*{Acknowlegements}
This work was supported by the strategic research environment ELLIIT, funded by the Swedish government.

\newpage
{
    \small
    \bibliographystyle{ieeenat_fullname}
    \bibliography{main}

\begin{thebibliography}{68}
\providecommand{\natexlab}[1]{#1}
\providecommand{\url}[1]{\texttt{#1}}
\expandafter\ifx\csname urlstyle\endcsname\relax
  \providecommand{\doi}[1]{doi: #1}\else
  \providecommand{\doi}{doi: \begingroup \urlstyle{rm}\Url}\fi

\bibitem[Aoki et~al.(2019)Aoki, Goforth, Srivatsan, and Lucey]{aoki2019pointnetlk}
Yasuhiro Aoki, Hunter Goforth, Rangaprasad~Arun Srivatsan, and Simon Lucey.
\newblock Pointnetlk: Robust \& efficient point cloud registration using pointnet.
\newblock In \emph{IEEE Conf. Comput. Vis. Pattern Recog.}, pages 7163--7172, 2019.

\bibitem[Ba et~al.(2016)Ba, Kiros, and Hinton]{ba2016layer}
Jimmy~Lei Ba, Jamie~Ryan Kiros, and Geoffrey~E Hinton.
\newblock Layer normalization.
\newblock \emph{arXiv preprint arXiv:1607.06450}, 2016.

\bibitem[{Besl} and {McKay}(1992)]{besl1992}
Paul~J. {Besl} and Neil~D. {McKay}.
\newblock A method for registration of 3-d shapes.
\newblock \emph{IEEE Trans. Pattern Anal. Mach. Intell.}, 14\penalty0 (2):\penalty0 239--256, 1992.

\bibitem[Bökman and Kahl(2022)]{bokman2022case}
Georg Bökman and Fredrik Kahl.
\newblock A case for using rotation invariant features in state of the art feature matchers.
\newblock In \emph{IEEE Conf. Comput. Vis. Pattern Recog.}, pages 5110--5119, 2022.

\bibitem[Campbell et~al.(2020)Campbell, Liu, and Gould]{campbell2020solving}
Dylan Campbell, Liu Liu, and Stephen Gould.
\newblock Solving the blind perspective-n-point problem end-to-end with robust differentiable geometric optimization.
\newblock In \emph{Eur. Conf. Comput. Vis.}, pages 244--261, 2020.

\bibitem[Chen and Medioni(1991)]{Chen1991}
Yang Chen and Gerard Medioni.
\newblock Object modeling by registration of multiple range images.
\newblock In \emph{IEEE Int'l Conf. Robotics and Auto.}, pages 2724--2729, 1991.

\bibitem[Choy et~al.(2019)Choy, Park, and Koltun]{choy2019fully}
Christopher Choy, Jaesik Park, and Vladlen Koltun.
\newblock Fully convolutional geometric features.
\newblock In \emph{Int. Conf. Comput. Vis.}, pages 8958--8966, 2019.

\bibitem[Cohen et~al.(2015)Cohen, Sattler, and Pollefeys]{cohen2015merging}
Andrea Cohen, Torsten Sattler, and Marc Pollefeys.
\newblock Merging the unmatchable: Stitching visually disconnected sfm models.
\newblock In \emph{Int. Conf. Comput. Vis.}, pages 2129--2137, 2015.

\bibitem[Cohen et~al.(2016)Cohen, Sch{\"o}nberger, Speciale, Sattler, Frahm, and Pollefeys]{cohen2016indoor}
Andrea Cohen, Johannes~L Sch{\"o}nberger, Pablo Speciale, Torsten Sattler, Jan-Michael Frahm, and Marc Pollefeys.
\newblock Indoor-outdoor 3d reconstruction alignment.
\newblock In \emph{Eur. Conf. Comput. Vis.}, pages 285--300, 2016.

\bibitem[Coughlan and Yuille(2000)]{coughlan2000manhattan}
James Coughlan and Alan~L Yuille.
\newblock The manhattan world assumption: Regularities in scene statistics which enable bayesian inference.
\newblock \emph{Adv. Neural Inform. Process. Syst.}, 13, 2000.

\bibitem[Crandall et~al.(2011)Crandall, Owens, Snavely, and Huttenlocher]{crandall2011discrete}
David Crandall, Andrew Owens, Noah Snavely, and Dan Huttenlocher.
\newblock Discrete-continuous optimization for large-scale structure from motion.
\newblock In \emph{IEEE Conf. Comput. Vis. Pattern Recog.}, pages 3001--3008, 2011.

\bibitem[Dangwal et~al.(2021)Dangwal, Lee, Kim, Shen, Cowan, Shah, Trippel, Reagen, Sherwood, Balntas, et~al.]{dangwal2021mitigating}
Deeksha Dangwal, Vincent~T Lee, Hyo~Jin Kim, Tianwei Shen, Meghan Cowan, Rajvi Shah, Caroline Trippel, Brandon Reagen, Timothy Sherwood, Vasileios Balntas, et~al.
\newblock Mitigating reverse engineering attacks on local feature descriptors.
\newblock In \emph{Brit. Mach. Vis. Conf.}, 2021.

\bibitem[Deng et~al.(2018{\natexlab{a}})Deng, Birdal, and Ilic]{deng2018ppffold}
Haowen Deng, Tolga Birdal, and Slobodan Ilic.
\newblock Ppf-foldnet: Unsupervised learning of rotation invariant 3d local descriptors.
\newblock In \emph{Eur. Conf. Comput. Vis.}, pages 602--618, 2018{\natexlab{a}}.

\bibitem[Deng et~al.(2018{\natexlab{b}})Deng, Birdal, and Ilic]{deng2018ppfnet}
Haowen Deng, Tolga Birdal, and Slobodan Ilic.
\newblock Ppfnet: Global context aware local features for robust 3d point matching.
\newblock In \emph{IEEE Conf. Comput. Vis. Pattern Recog.}, pages 195--205, 2018{\natexlab{b}}.

\bibitem[Dosovitskiy and Brox(2016)]{dosovitskiy2016inverting}
Alexey Dosovitskiy and Thomas Brox.
\newblock Inverting visual representations with convolutional networks.
\newblock In \emph{IEEE Conf. Comput. Vis. Pattern Recog.}, pages 4829--4837, 2016.

\bibitem[Drost et~al.(2010)Drost, Ulrich, Navab, and Ilic]{drost2010model}
Bertram Drost, Markus Ulrich, Nassir Navab, and Slobodan Ilic.
\newblock Model globally, match locally: Efficient and robust 3d object recognition.
\newblock In \emph{IEEE Conf. Comput. Vis. Pattern Recog.}, pages 998--1005, 2010.

\bibitem[Dusmanu et~al.(2021)Dusmanu, Miksik, Sch{\"o}nberger, and Pollefeys]{dusmanu2021cross}
Mihai Dusmanu, Ondrej Miksik, Johannes~L Sch{\"o}nberger, and Marc Pollefeys.
\newblock Cross-descriptor visual localization and mapping.
\newblock In \emph{Int. Conf. Comput. Vis.}, pages 6058--6067, 2021.

\bibitem[Fischler and Bolles(1981)]{fischler1981}
Martin~A Fischler and Robert~C Bolles.
\newblock Random sample consensus: a paradigm for model fitting with applications to image analysis and automated cartography.
\newblock \emph{Communications of the ACM}, 24\penalty0 (6):\penalty0 381--395, 1981.

\bibitem[Fu et~al.(2021)Fu, Liu, Luo, and Wang]{fu2021robust}
Kexue Fu, Shaolei Liu, Xiaoyuan Luo, and Manning Wang.
\newblock Robust point cloud registration framework based on deep graph matching.
\newblock In \emph{IEEE Conf. Comput. Vis. Pattern Recog.}, pages 8893--8902, 2021.

\bibitem[Geiger et~al.(2012)Geiger, Lenz, and Urtasun]{geiger2012}
Andreas Geiger, Philip Lenz, and Raquel Urtasun.
\newblock Are we ready for autonomous driving? the kitti vision benchmark suite.
\newblock In \emph{IEEE Conf. Comput. Vis. Pattern Recog.}, pages 3354--3361, 2012.

\bibitem[Gojcic et~al.(2019)Gojcic, Zhou, Wegner, and Wieser]{gojcic2019perfect}
Zan Gojcic, Caifa Zhou, Jan~D Wegner, and Andreas Wieser.
\newblock The perfect match: 3d point cloud matching with smoothed densities.
\newblock In \emph{IEEE Conf. Comput. Vis. Pattern Recog.}, pages 5545--5554, 2019.

\bibitem[Havlena et~al.(2010)Havlena, Torii, and Pajdla]{havlena2010efficient}
Michal Havlena, Akihiko Torii, and Tom{\'a}{\v{s}} Pajdla.
\newblock Efficient structure from motion by graph optimization.
\newblock In \emph{Eur. Conf. Comput. Vis.}, pages 100--113, 2010.

\bibitem[Hendrycks and Gimpel(2016)]{hendrycks2016gaussian}
Dan Hendrycks and Kevin Gimpel.
\newblock Gaussian error linear units (gelus).
\newblock \emph{arXiv preprint arXiv:1606.08415}, 2016.

\bibitem[Huang et~al.(2021)Huang, Gojcic, Usvyatsov, Wieser, and Schindler]{huang2021predator}
Shengyu Huang, Zan Gojcic, Mikhail Usvyatsov, Andreas Wieser, and Konrad Schindler.
\newblock Predator: Registration of 3d point clouds with low overlap.
\newblock In \emph{IEEE Conf. Comput. Vis. Pattern Recog.}, pages 4267--4276, 2021.

\bibitem[Huang et~al.(2020)Huang, Mei, and Zhang]{huang2020feature}
Xiaoshui Huang, Guofeng Mei, and Jian Zhang.
\newblock Feature-metric registration: A fast semi-supervised approach for robust point cloud registration without correspondences.
\newblock In \emph{IEEE Conf. Comput. Vis. Pattern Recog.}, pages 11366--11374, 2020.

\bibitem[Jin et~al.(2024)Jin, Armeni, Pollefeys, and Barath]{jin2024}
Shengze Jin, Iro Armeni, Marc Pollefeys, and Daniel Barath.
\newblock Multiway point cloud mosaicking with diffusion and global optimization.
\newblock In \emph{IEEE Conf. Comput. Vis. Pattern Recog.}, pages 20838--20849, 2024.

\bibitem[Kamath(2019)]{azuresp}
Neena Kamath.
\newblock Announcing azure spatial anchors for collaborative, cross-platform mixed reality apps, 2019.
\newblock URL \url{https://azure.microsoft.com/en-us/blog/announcing-azure-spatial-anchors-for-collaborative-cross-platform-mixed-reality-apps/}.

\bibitem[Kendall et~al.(2015)Kendall, Grimes, and Cipolla]{kendall2015posenet}
Alex Kendall, Matthew Grimes, and Roberto Cipolla.
\newblock Posenet: A convolutional network for real-time 6-dof camera relocalization.
\newblock In \emph{Int. Conf. Comput. Vis.}, pages 2938--2946, 2015.

\bibitem[Li and Hartley(2007)]{Li2007}
Hongdong Li and Richard Hartley.
\newblock The 3d-3d registration problem revisited.
\newblock In \emph{Int. Conf. Comput. Vis.}, pages 1--8, 2007.

\bibitem[Li and Snavely(2018)]{li2018megadepth}
Zhengqi Li and Noah Snavely.
\newblock Megadepth: Learning single-view depth prediction from internet photos.
\newblock In \emph{IEEE Conf. Comput. Vis. Pattern Recog.}, pages 2041--2050, 2018.

\bibitem[Lindenberger et~al.(2023)Lindenberger, Sarlin, and Pollefeys]{lindenberger2023lightglue}
Philipp Lindenberger, Paul-Edouard Sarlin, and Marc Pollefeys.
\newblock {LightGlue: Local Feature Matching at Light Speed}.
\newblock In \emph{Int. Conf. Comput. Vis.}, 2023.

\bibitem[Liu et~al.(2021)Liu, Li, Yao, and Zha]{liu2021pluckernet}
Liu Liu, Hongdong Li, Haodong Yao, and Ruyi Zha.
\newblock Pluckernet: Learn to register 3d line reconstructions.
\newblock In \emph{Proceedings of the IEEE/CVF Conference on Computer Vision and Pattern Recognition}, pages 1842--1852, 2021.

\bibitem[Locher et~al.(2018)Locher, Havlena, and Van~Gool]{locher2018progressive}
Alex Locher, Michal Havlena, and Luc Van~Gool.
\newblock Progressive structure from motion.
\newblock In \emph{Eur. Conf. Comput. Vis.}, pages 20--35, 2018.

\bibitem[Lopes et~al.(2022)Lopes, Souza, and Pedrini]{lopes2022survey}
Alexandre Lopes, Roberto Souza, and Helio Pedrini.
\newblock A survey on rgb-d datasets.
\newblock \emph{Comput. Vis. Img. Underst.}, 222:\penalty0 103489, 2022.

\bibitem[Lowe(2004)]{lowe2004distinctive}
David~G Lowe.
\newblock Distinctive image features from scale-invariant keypoints.
\newblock \emph{Int. J. Comput. Vis.}, 60:\penalty0 91--110, 2004.

\bibitem[Mateus et~al.(2023)Mateus, Ranade, Ramalingam, and Miraldo]{mateus2023fast}
Andr{\'e} Mateus, Siddhant Ranade, Srikumar Ramalingam, and Pedro Miraldo.
\newblock Fast and accurate 3d registration from line intersection constraints.
\newblock \emph{Int. J. Comput. Vis.}, pages 1--26, 2023.

\bibitem[Miraldo et~al.(2019)Miraldo, Saha, and Ramalingam]{miraldo19}
Pedro Miraldo, Surojit Saha, and Srikumar Ramalingam.
\newblock Minimal solvers for mini-loop closures in 3d multi-scan alignment.
\newblock In \emph{IEEE Conf. Comput. Vis. Pattern Recog.}, 2019.

\bibitem[Pittaluga et~al.(2019)Pittaluga, Koppal, Kang, and Sinha]{pittaluga2019revealing}
Francesco Pittaluga, Sanjeev~J Koppal, Sing~Bing Kang, and Sudipta~N Sinha.
\newblock Revealing scenes by inverting structure from motion reconstructions.
\newblock In \emph{IEEE Conf. Comput. Vis. Pattern Recog.}, pages 145--154, 2019.

\bibitem[Qin et~al.(2022)Qin, Yu, Wang, Guo, Peng, and Xu]{qin2022geometric}
Zheng Qin, Hao Yu, Changjian Wang, Yulan Guo, Yuxing Peng, and Kai Xu.
\newblock Geometric transformer for fast and robust point cloud registration.
\newblock In \emph{IEEE Conf. Comput. Vis. Pattern Recog.}, pages 11143--11152, 2022.

\bibitem[Reinhardt(2019)]{googlevps}
Tilman Reinhardt.
\newblock Using global localization to improve navigation, 2019.
\newblock URL \url{https://blog.research.google/2019/02/using-global-localization-to-improve.html}.

\bibitem[Rusu et~al.(2009)Rusu, Blodow, and Beetz]{Rusu2009}
Radu~Bogdan Rusu, Nico Blodow, and Michael Beetz.
\newblock Fast point feature histograms (fpfh) for 3d registration.
\newblock In \emph{IEEE Int'l Conf. Robotics and Auto.}, pages 3212--3217, 2009.

\bibitem[Sarlin et~al.(2019)Sarlin, Cadena, Siegwart, and Dymczyk]{sarlin2019coarse}
Paul-Edouard Sarlin, Cesar Cadena, Roland Siegwart, and Marcin Dymczyk.
\newblock From coarse to fine: Robust hierarchical localization at large scale.
\newblock In \emph{IEEE Conf. Comput. Vis. Pattern Recog.}, pages 12716--12725, 2019.

\bibitem[Sarlin et~al.(2020)Sarlin, DeTone, Malisiewicz, and Rabinovich]{sarlin20superglue}
Paul-Edouard Sarlin, Daniel DeTone, Tomasz Malisiewicz, and Andrew Rabinovich.
\newblock {SuperGlue}: Learning feature matching with graph neural networks.
\newblock In \emph{IEEE Conf. Comput. Vis. Pattern Recog.}, 2020.

\bibitem[Sch\"{o}nberger and Frahm(2016)]{schoenberger2016sfm}
Johannes~Lutz Sch\"{o}nberger and Jan-Michael Frahm.
\newblock Structure-from-motion revisited.
\newblock In \emph{IEEE Conf. Comput. Vis. Pattern Recog.}, 2016.

\bibitem[Schonemann(1966)]{Schonemann1966}
Peter~H. Schonemann.
\newblock A generalized solution of the orthogonal procrustes problem.
\newblock \emph{Psychometrika}, 31\penalty0 (1):\penalty0 1--10, 1966.

\bibitem[Shotton et~al.(2013)Shotton, Glocker, Zach, Izadi, Criminisi, and Fitzgibbon]{shotton2013scene}
Jamie Shotton, Ben Glocker, Christopher Zach, Shahram Izadi, Antonio Criminisi, and Andrew Fitzgibbon.
\newblock Scene coordinate regression forests for camera relocalization in rgb-d images.
\newblock In \emph{Proceedings of the IEEE conference on computer vision and pattern recognition}, pages 2930--2937, 2013.

\bibitem[Smith(2022)]{nianticvps}
Tory Smith.
\newblock Introducing lightship vps, 2022.
\newblock URL \url{https://lightship.dev/blog/introducing-lightship-vps/}.

\bibitem[Strecha et~al.(2010)Strecha, Pylv{\"a}n{\"a}inen, and Fua]{strecha2010dynamic}
Christoph Strecha, Timo Pylv{\"a}n{\"a}inen, and Pascal Fua.
\newblock Dynamic and scalable large scale image reconstruction.
\newblock In \emph{IEEE Conf. Comput. Vis. Pattern Recog.}, pages 406--413, 2010.

\bibitem[Sun et~al.(2020)Sun, Cheng, Zhang, Zhang, Zheng, Wang, and Wei]{sun2020circle}
Yifan Sun, Changmao Cheng, Yuhan Zhang, Chi Zhang, Liang Zheng, Zhongdao Wang, and Yichen Wei.
\newblock Circle loss: A unified perspective of pair similarity optimization.
\newblock In \emph{IEEE Conf. Comput. Vis. Pattern Recog.}, pages 6398--6407, 2020.

\bibitem[Sweeney et~al.(2016)Sweeney, Fragoso, H{\"o}llerer, and Turk]{sweeney2016large}
Chris Sweeney, Victor Fragoso, Tobias H{\"o}llerer, and Matthew Turk.
\newblock Large scale sfm with the distributed camera model.
\newblock In \emph{Int'l Conf. 3D Vision}, pages 230--238, 2016.

\bibitem[Tian et~al.(2019)Tian, Yu, Fan, Wu, Heijnen, and Balntas]{tian2019sosnet}
Yurun Tian, Xin Yu, Bin Fan, Fuchao Wu, Huub Heijnen, and Vassileios Balntas.
\newblock Sosnet: Second order similarity regularization for local descriptor learning.
\newblock In \emph{IEEE Conf. Comput. Vis. Pattern Recog.}, pages 11016--11025, 2019.

\bibitem[Turki et~al.(2022)Turki, Ramanan, and Satyanarayanan]{Turki_2022_CVPR}
Haithem Turki, Deva Ramanan, and Mahadev Satyanarayanan.
\newblock Mega-nerf: Scalable construction of large-scale nerfs for virtual fly-throughs.
\newblock In \emph{IEEE Conf. Comput. Vis. Pattern Recog.}, pages 12922--12931, 2022.

\bibitem[Tyszkiewicz et~al.(2020)Tyszkiewicz, Fua, and Trulls]{tyszkiewicz2020disk}
Micha{\l} Tyszkiewicz, Pascal Fua, and Eduard Trulls.
\newblock Disk: Learning local features with policy gradient.
\newblock \emph{Adv. Neural Inform. Process. Syst.}, 33:\penalty0 14254--14265, 2020.

\bibitem[Untzelmann et~al.(2013)Untzelmann, Sattler, Middelberg, and Kobbelt]{untzelmann2013scalable}
Ole Untzelmann, Torsten Sattler, Sven Middelberg, and Leif Kobbelt.
\newblock A scalable collaborative online system for city reconstruction.
\newblock In \emph{Proceedings of the IEEE International Conference on Computer Vision Workshops}, pages 644--651, 2013.

\bibitem[Wang et~al.(2024)Wang, Kannala, and Barath]{wang2023dgc}
Shuzhe Wang, Juho Kannala, and Daniel Barath.
\newblock Dgc-gnn: Leveraging geometry and color cues for visual descriptor-free 2d-3d matching.
\newblock In \emph{Proceedings of the IEEE/CVF Conference on Computer Vision and Pattern Recognition}, pages 20881--20891, 2024.

\bibitem[Wang and Solomon(2019{\natexlab{a}})]{wang2019dcp}
Yue Wang and Justin~M Solomon.
\newblock Deep closest point: Learning representations for point cloud registration.
\newblock In \emph{Int. Conf. Comput. Vis.}, pages 3523--3532, 2019{\natexlab{a}}.

\bibitem[Wang and Solomon(2019{\natexlab{b}})]{wang2019prnet}
Yue Wang and Justin~M Solomon.
\newblock Prnet: Self-supervised learning for partial-to-partial registration.
\newblock \emph{Adv. Neural Inform. Process. Syst.}, 32, 2019{\natexlab{b}}.

\bibitem[Weinzaepfel et~al.(2011)Weinzaepfel, J{\'e}gou, and P{\'e}rez]{weinzaepfel2011reconstructing}
Philippe Weinzaepfel, Herv{\'e} J{\'e}gou, and Patrick P{\'e}rez.
\newblock Reconstructing an image from its local descriptors.
\newblock In \emph{IEEE Conf. Comput. Vis. Pattern Recog.}, pages 337--344, 2011.

\bibitem[Xu et~al.(2021)Xu, Liu, Wang, Liu, and Zeng]{xu2021omnet}
Hao Xu, Shuaicheng Liu, Guangfu Wang, Guanghui Liu, and Bing Zeng.
\newblock Omnet: Learning overlapping mask for partial-to-partial point cloud registration.
\newblock In \emph{Int. Conf. Comput. Vis.}, pages 3132--3141, 2021.

\bibitem[Yang et~al.(2013)Yang, Li, and Jia]{Yang2013}
Jiaolong Yang, Hongdong Li, and Yunde Jia.
\newblock Go-icp: Solving 3d registration efficiently and globally optimally.
\newblock In \emph{Int. Conf. Comput. Vis.}, pages 1457--1464, 2013.

\bibitem[Yao et~al.(2024)Yao, Du, Cui, Tang, and Yang]{yao2025pare}
Runzhao Yao, Shaoyi Du, Wenting Cui, Canhui Tang, and Chengwu Yang.
\newblock Pare-net: Position-aware rotation-equivariant networks for robust point cloud registration.
\newblock In \emph{ECCV}, 2024.

\bibitem[Yew and Lee(2020)]{yew2020rpm}
Zi~Jian Yew and Gim~Hee Lee.
\newblock Rpm-net: Robust point matching using learned features.
\newblock In \emph{IEEE Conf. Comput. Vis. Pattern Recog.}, pages 11824--11833, 2020.

\bibitem[Yew and Lee(2022)]{yew2022regtr}
Zi~Jian Yew and Gim~Hee Lee.
\newblock Regtr: End-to-end point cloud correspondences with transformers.
\newblock In \emph{IEEE Conf. Comput. Vis. Pattern Recog.}, pages 6677--6686, 2022.

\bibitem[Yu et~al.(2023)Yu, Qin, Hou, Saleh, Li, Busam, and Ilic]{yu2023rotation}
Hao Yu, Zheng Qin, Ji Hou, Mahdi Saleh, Dongsheng Li, Benjamin Busam, and Slobodan Ilic.
\newblock Rotation-invariant transformer for point cloud matching.
\newblock In \emph{IEEE Conf. Comput. Vis. Pattern Recog.}, pages 5384--5393, 2023.

\bibitem[Zeng et~al.(2017)Zeng, Song, Nie{\ss}ner, Fisher, Xiao, and Funkhouser]{zeng20173dmatch}
Andy Zeng, Shuran Song, Matthias Nie{\ss}ner, Matthew Fisher, Jianxiong Xiao, and Thomas Funkhouser.
\newblock 3dmatch: Learning local geometric descriptors from rgb-d reconstructions.
\newblock In \emph{IEEE Conf. Comput. Vis. Pattern Recog.}, pages 1802--1811, 2017.

\bibitem[Zhang et~al.(2025)Zhang, Wang, and Kannala]{zhang2025a2}
Yejun Zhang, Shuzhe Wang, and Juho Kannala.
\newblock A2-gnn: Angle-annular gnn for visual descriptor-free camera relocalization.
\newblock In \emph{International Conference on 3D Vision (3DV)}, 2025.
\newblock arXiv preprint arXiv:2502.20036.

\bibitem[Zhou et~al.(2022)Zhou, Agostinho, O{\v{s}}ep, and Leal-Taix{\'e}]{zhou2022geometry}
Qunjie Zhou, S{\'e}rgio Agostinho, Aljo{\v{s}}a O{\v{s}}ep, and Laura Leal-Taix{\'e}.
\newblock Is geometry enough for matching in visual localization?
\newblock In \emph{Eur. Conf. Comput. Vis.}, pages 407--425, 2022.

\bibitem[Zhou et~al.(2018)Zhou, Park, and Koltun]{Zhou2018open3d}
Qian-Yi Zhou, Jaesik Park, and Vladlen Koltun.
\newblock {Open3D}: {A} modern library for {3D} data processing.
\newblock \emph{arXiv:1801.09847}, 2018.

\end{thebibliography}
}
\clearpage
\setcounter{page}{1}
\maketitlesupplementary
In this supplementary material, we provide details on the training, RANSAC implementation, computation of point cloud overlap, additional registration results, and additional qualitative results that could not fit in the main paper.

\section{Training Details}
\label{sec:train}
The models were trained on a NVIDIA 4090 GPU, both in 3DMatch~\cite{zeng20173dmatch} and our dataset. We used the Adam Optimizer, trained for $150$ epochs, with a batch size of 1, learning rate of $0.0001$, and exponential decay $0.05$.

\section{RANSAC}
\label{sec:pose_est}
We solve for the pose using the RANSAC~\cite{fischler1981} implementation from Open3D~\cite{Zhou2018open3d} on the matches retrieved by the neural networks. 
The samples consist of three points, and inlier counting is performed by checking if the correspondence distance is below $\tau_{\text{IR}} = 0.05$. We use at most 1000 correspondences, if the models return more than 1000 matches, we random sample 1000 matches. If correspondence scores are given by the methods, we use them to weight the correspondence sampling.
Rotation and translation error are computed as
\begin{align}
    \epsilon_{R}\left(R\right) & = \text{acos} \left(\tfrac{\text{trace}\left(R^{-1} R_{\text{GT}}\right)- 1}{2}\right)\ \label{eq:rot_error} \\ 
    \epsilon_{\mathbf{t}}\left(\mathbf{t} \right) & =  \| \mathbf{t} - \mathbf{t}_{\text{GT}}\|, \label{eq:trans_error}
\end{align}
respectively. Where $\|\cdot\|$ represents the Euclidean norm.

\section{Computing Overlaps} 
We compute the overlap between two point clouds as the approximate intersection over union of the points. We define the directional intersection of $\mathbf{P}\to\mathbf{Q}$ as
\begin{equation}
    \text{I}_{\mathbf{P}\to\mathbf{Q}} = \sum_{p\in\mathbf{P}} (\min_{q\in \mathbf{Q}}  \text{dist}(Rp+t, q) \leq \tau),
\end{equation}
with $\tau = 0.1$, and the union as
\begin{equation}
    \text{U}_{\mathbf{P}\to\mathbf{Q}} = |\mathbf{P}|.
\end{equation}
The directional IoU from $\mathbf{P}\to\mathbf{Q}$ is then
\begin{equation}
    \text{IoU}_{{\mathbf{P}\to\mathbf{Q}}} = \frac{\text{I}_{\mathbf{P}\to\mathbf{Q}}}{\text{U}_{\mathbf{P}\to\mathbf{Q}}}.
\end{equation}
We define the overlap between the two point clouds as the geometric mean of their directional IoUs, \ie, 
\begin{equation}
    \text{IoU}_{\mathbf{P},\mathbf{Q}} = \sqrt{\text{IoU}_{{\mathbf{P}\to\mathbf{Q}}}\cdot \text{IoU}_{{\mathbf{Q}\to\mathbf{P}}}}.
\end{equation}

\section{Cambridge Landmarks Additional Results}

The rotation and translation errors computed using \eqref{eq:rot_error} and \eqref{eq:trans_error} are presented in \cref{tab:cambridge_reg}.
The methods trained only on 3DMatch~\cite{zeng20173dmatch} present poor results, which is expected given the low number of inliers (\cf \cref{tab:cambridge} of the main paper).
The methods trained and/or fine-tuned on our dataset have small errors, with the performance being similar for all four.
To assess the performance of the registration methods against visual cues, we reconstructed all scenes using SIFT descriptors, subsampled the 3D reconstructions to $30$k points (the same number used for the registration models) and computed the average descriptor for each 2D point in a 3D point's track. Then, we used Nearest Neighboor (NN) matching and RANSAC to find the pose, this method is called SIFT+NN, see \cref{tab:cambridge_reg}.
Notice that for most scenes, our descriptor-free approach has comparable results to the descriptor-based approach, validating our results.
The most challenging scene, for the registration methods, was \emph{Old Hospital}, which consists of a facade with repetitive structure.
The point cloud consists mostly of the facade and further way points in much sparser areas leading to a set of bad matches found in those regions, see \cref{fig:registration_results_cambridge}.
The same happens in \emph{Shop Facade} which consists of a corner shop, which also contains a high amount of points in further and sparser areas, resulting in a high number of bad matches.
Nevertheless, the correct matches found by the fine-tuned versions \ours~and RoITr~\cite{yu2023rotation} are enough to find accurate poses for the registration.

We additionally further investigate out-of-distribution results of our model trained on SOSNet and SIFT reconstructions using DISK~\citep{tyszkiewicz2020disk} and LightGlue~\citep{lindenberger2023lightglue}.
Results are presented in \cref{tab:cambridge_disk}. 
Our trained models demonstrate the ability to generalize to
reconstructions of a different nature, however showing a slight decrease in performance. It is likely that including a more diverse set of detectors in the training set would decrease this gap, and further improve the generalizability of our approach.

\begin{table*}
\centering
    \caption{\textbf{Results of SfM registration on Cambridge Landmarks~\cite{kendall2015posenet}}. We evaluate unknown relative orientation SO(3) + unknown relative position = SE(3).
    The top portion contains methods only trained on the 3DMatch (3DM) dataset, the middle portion methods trained only on our proposed dataset (Mega), while the lower portion contains methods trained on the former and fine-tuned on the latter. Rotation error is reported in degrees, translation error is unitless since the pointclouds are scaled.}
    \resizebox{1\linewidth}{!}{\begin{tabular}{l cc cc cc cc cc}
        \toprule
        & \multicolumn{2}{c}{Great Court} & \multicolumn{2}{c}{Kings College}& \multicolumn{2}{c}{Old Hospital} & \multicolumn{2}{c}{Shop Facade} &  \multicolumn{2}{c}{St Mary's Church}\\
        \cmidrule(lr){2-3} \cmidrule(lr){4-5} \cmidrule(lr){6-7} \cmidrule(lr){8-9} \cmidrule(lr){10-11}
        Method & $\epsilon_R$ & $\epsilon_{\mathbf{t}}$ & $\epsilon_R$ & $\epsilon_{\mathbf{t}}$ & $\epsilon_R$ & $\epsilon_{\mathbf{t}}$ & $\epsilon_R$ & $\epsilon_{\mathbf{t}}$ & $\epsilon_R$ & $\epsilon_{\mathbf{t}}$ \\
        \midrule
        SIFT + NN & \textbf{0.1}$^\circ$ & 0.09 & \textbf{0.2}$^\circ$ & \underline{0.02} & \textbf{0.2}$^\circ$ & \underline{0.05} & \textbf{0.1}$^\circ$ & \textbf{0.01} & \textbf{0.04}$^\circ$ & \textbf{0.01} \\
        \midrule
         OverlapPredator~\cite{huang2021predator} (3DM) & 6.9$^\circ$ & 0.43 & 27.2$^\circ$ & 0.24 & 135.5$^\circ$ & 2.14 & 179.4$^\circ$ & 0.82 & 170.7$^\circ$ & 4.27 \\
         GeoTransformer~\cite{qin2022geometric} (3DM) & 177.2$^\circ$ & 7.26 & 81.2$^\circ$ & 15.13 & 179.0$^\circ$ & 14.65 & 123.3$^\circ$ & 19.27 & 146.8$^\circ$ & 12.13\\
         PareNet~\cite{yao2025pare} (3DM) & 173.7$^\circ$ & 4.24 & 149.4$^\circ$& 4.82 &  171.1$^\circ$ & 4.69 & 32.74$^\circ$ & 1.10 & 176.07$^\circ$ & 3.48 \\
         RoITr~\cite{yu2023rotation} (3DM) & 179.4$^\circ$ & 6.93 & 157.3$^\circ$ & 22.42 & 76.7$^\circ$ & 0.0 & 176.9$^\circ$ & 1.09 & 99.3$^\circ$ & 2.14 \\
         \ours~(3DM) & 73.8$^\circ$ & 1.76 & 90.3$^\circ$ & 6.13 & 81.6$^\circ$ & 3.97 & 84.8$^\circ$ & 1.04 & 138.6$^\circ$ & 15.83 \\
         \midrule
         \ours~(Mega) & 0.5$^\circ$ & \textbf{0.02} & \underline{0.5}$^\circ$ & \underline{0.02} & \underline{1.16}$^\circ$ & \underline{0.05} & 0.9$^\circ$ & \underline{0.02} & 0.3$^\circ$ & \textbf{0.01} \\
         \ours~(Mega w\textbackslash~color)  & \underline{0.3}$^\circ$ & \textbf{0.02} & 0.6$^\circ$ & 0.03 & 3.0$^\circ$ & 0.15 & \underline{0.5}$^\circ$ & 0.03 & 0.3$^\circ$ & \textbf{0.01} \\
         \midrule
         RoITr~\cite{yu2023rotation}  (3DM + Mega) & 0.4$^\circ$ & \textbf{0.02} & 0.6$^\circ$ & \textbf{0.01} & 2.6$^\circ$ & 0.10 & 2.6$^\circ$ & \underline{0.02} & \underline{0.2}$^\circ$ & \underline{0.02}\\
         \ours~(3DM + Mega) & 0.5$^\circ$ & \underline{0.03} & \textbf{0.2}$^\circ$ & \underline{0.02} & 2.1$^\circ$ & \textbf{0.04} & 0.9$^\circ$ & 0.04 & 0.3$^\circ$ & \textbf{0.01} \\
         \bottomrule
    \end{tabular}}
    \label{tab:cambridge_reg}
\end{table*}

\begin{table*}
\centering
    \caption{\textbf{Results of Out-of-Distribution SfM registration on Cambridge Landmarks~\cite{kendall2015posenet}}. Here we take a network trained on SIFT and SOSNet reconstructions, and evaluate it on DISK reconstructions. We evaluate unknown relative orientation SO(3) + unknown relative position = SE(3).
    The top portion contains methods only trained on the 3DMatch (3DM) dataset, the middle portion methods trained only on our proposed dataset (Mega), while the lower portion contains methods trained on the former and fine-tuned on the latter. }
    \resizebox{1\linewidth}{!}{\begin{tabular}{l cc cc cc cc cc}
        \toprule
        & \multicolumn{2}{c}{Great Court} & \multicolumn{2}{c}{Kings College}& \multicolumn{2}{c}{Old Hospital} & \multicolumn{2}{c}{Shop Facade} &  \multicolumn{2}{c}{St Mary's Church}\\
        \cmidrule(lr){2-3} \cmidrule(lr){4-5} \cmidrule(lr){6-7} \cmidrule(lr){8-9} \cmidrule(lr){10-11}
        Method & IR & Matches & IR & Matches & IR & Matches & IR & Matches & IR & Matches \\
        \midrule
         OverlapPredator~\cite{huang2021predator} (3DM) & 2.6 & 352 & 1.1 & 361 & 0 & 286 & 1.7 & 359 & 0 & 373\\
         GeoTransformer~\cite{qin2022geometric} (3DM) & 0 & 256 & 0 & 283 & 0 & 238 & 0 & 278 & 0 & 309\\
         RoITr~\cite{yu2023rotation} (3DM) & 0 & 400 & 0 & 733 & 0 & 256 & 0 & \textbf{930} & 0.2 & 867\\
         \ours~(3DM) & 0 & 306 & 0 & 581 & 0 & 256 & 0 & \underline{742} & 2.5 & 603\\
         \midrule
         \ours~(Mega)& \underline{36.9} & \textbf{3942} & \textbf{54.1} & \textbf{2466} & \underline{9.5} & \textbf{1437} & \textbf{13.5} & 370 & \underline{60.0} & \textbf{4594}\\
         \midrule
         RoITr~\cite{yu2023rotation}  (3DM + Mega) & 29.4 & 361 & 30.5 & 511 & 5.9 & 152 & 2.7 & 449 & 57.9 & 1528\\
         \ours~(3DM + Mega)& \textbf{39.8} & \underline{2848} & \underline{49.5} & \underline{1744} & \textbf{9.7} & \underline{1152} & \underline{10.1} & 485 & \textbf{61.0} & \underline{4281}\\
         \bottomrule
    \end{tabular}}
    \label{tab:cambridge_disk}
\end{table*}

\section{Additional qualitative results}

\begin{figure*}
    \centering
    \resizebox{\linewidth}{!}{\begin{tabular}{lcccc}
    {\LARGE \textbf{Matches}} & & & & \\ %
    & {\Large Input} & {\Large OverlapPredator~\cite{huang2021predator} (3DM)} & {\Large RoITr~\cite{yu2023rotation} (3DM)} & {\Large \ours~(Mega)} \\ 
    &
    \includegraphics[height=.21\textheight]{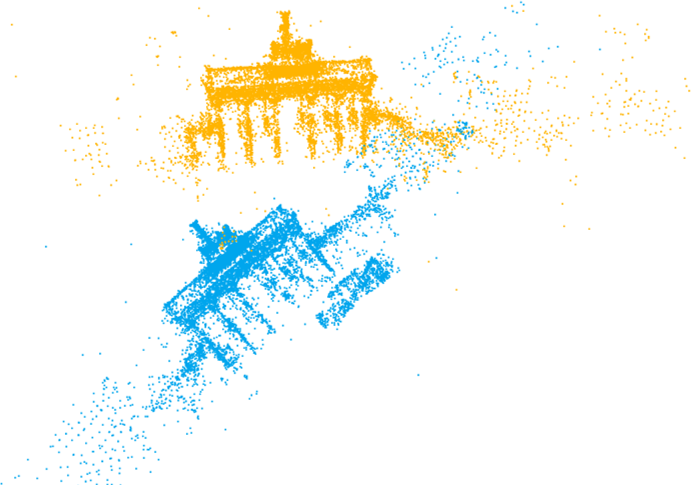} &
    \includegraphics[height=.21\textheight]{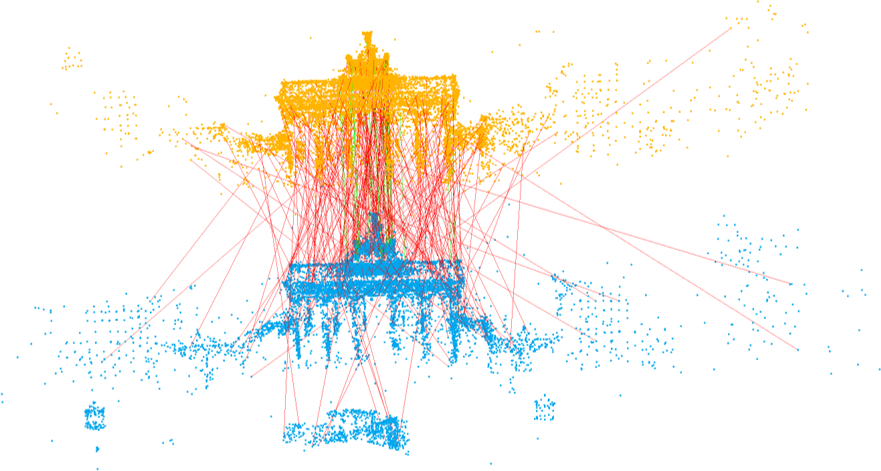} &
    \includegraphics[height=.21\textheight]{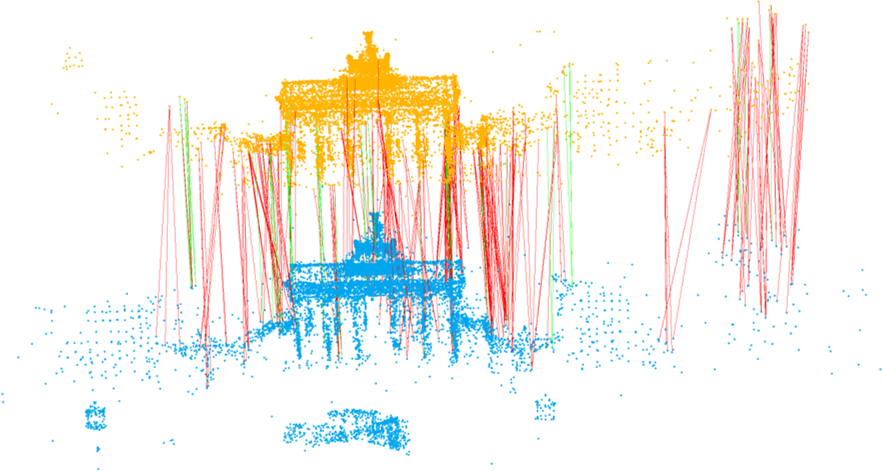} &
    \includegraphics[height=.21\textheight]{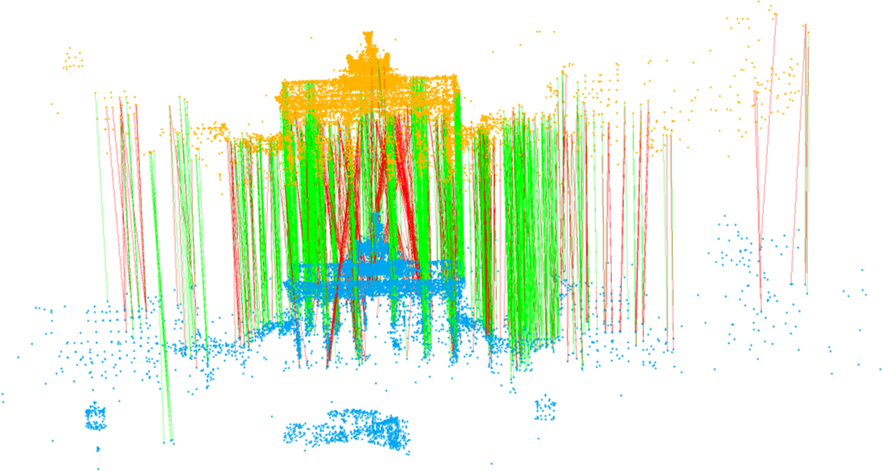} \\
    &
    \includegraphics[height=.21\textheight]{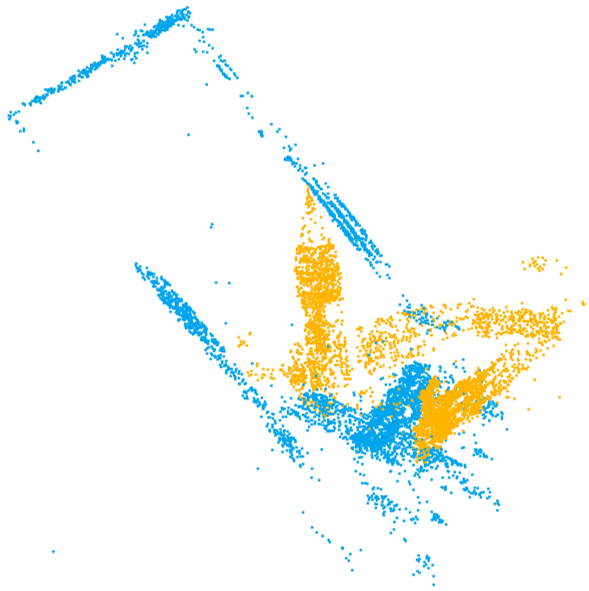} &
    \includegraphics[height=.21\textheight]{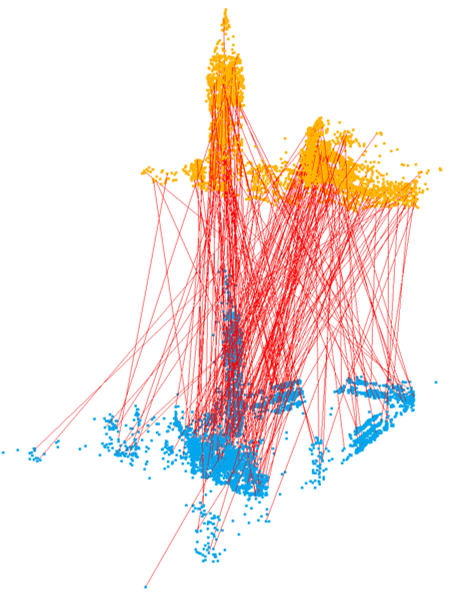} &
    \includegraphics[height=.21\textheight]{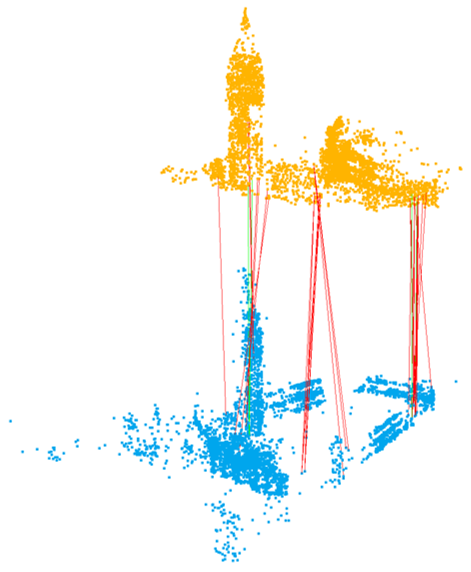} &
    \includegraphics[height=.21\textheight]{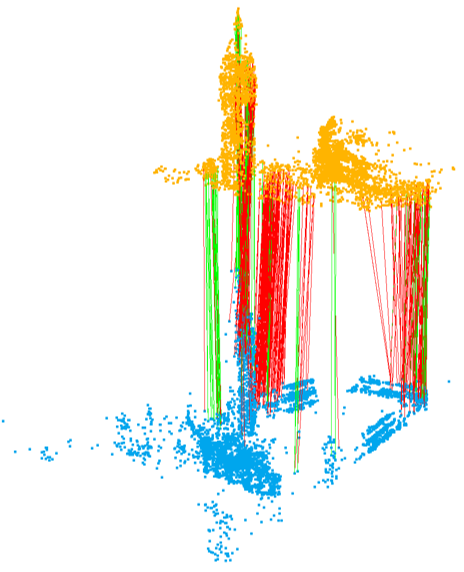} \\
    \midrule \\
    {\LARGE \textbf{Registration}} & & & & \\
    & {\large GT} & {\Large OverlapPredator~\cite{huang2021predator} (3DM)} & {\Large RoITr~\cite{yu2023rotation} (3DM)} & {\Large \ours~(Mega)} \\ 
    &
    \includegraphics[height=.21\textheight]{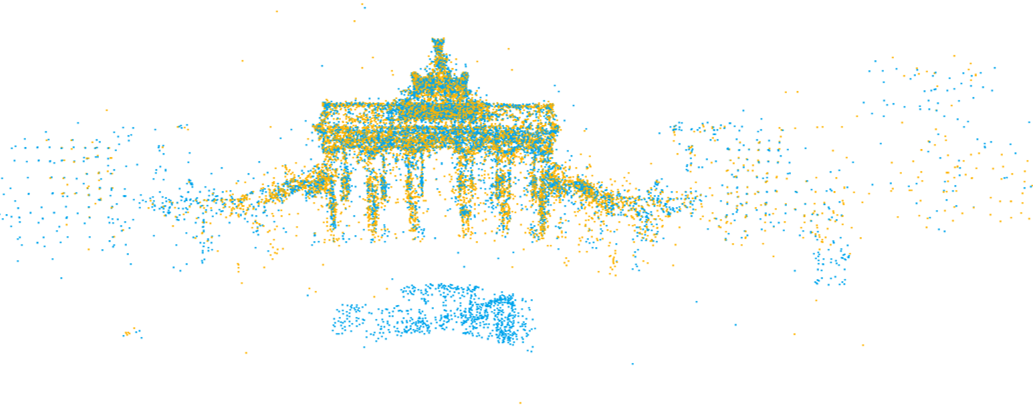} &
    \includegraphics[height=.21\textheight]{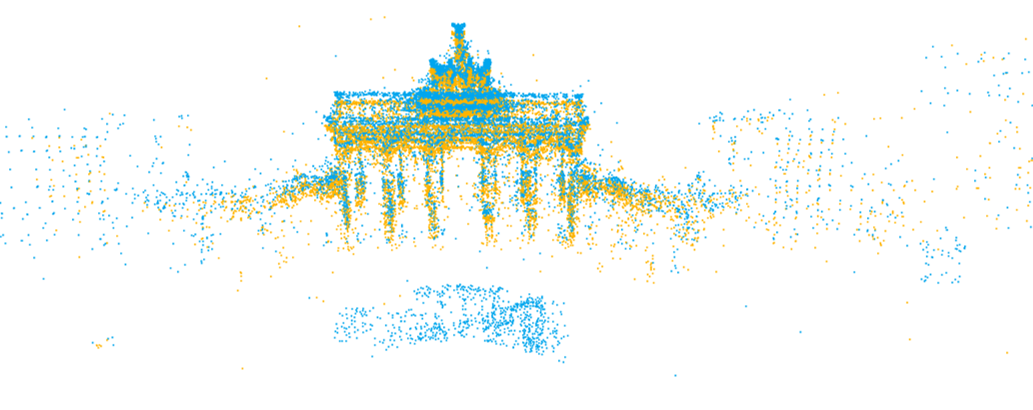} &
    \includegraphics[height=.21\textheight]{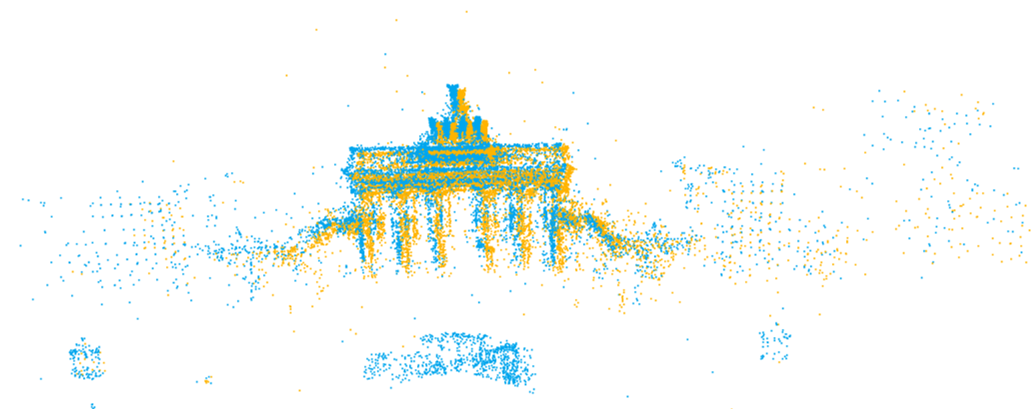} &
    \includegraphics[height=.21\textheight]{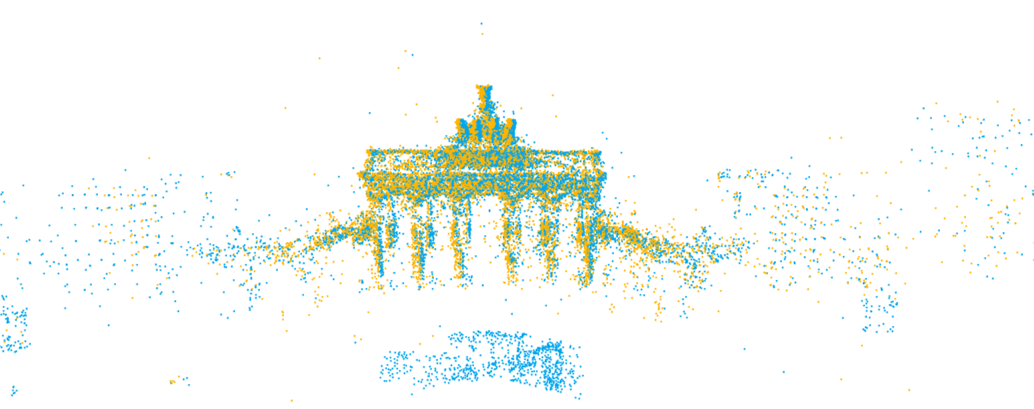} \\
    &
    \includegraphics[height=.21\textheight]{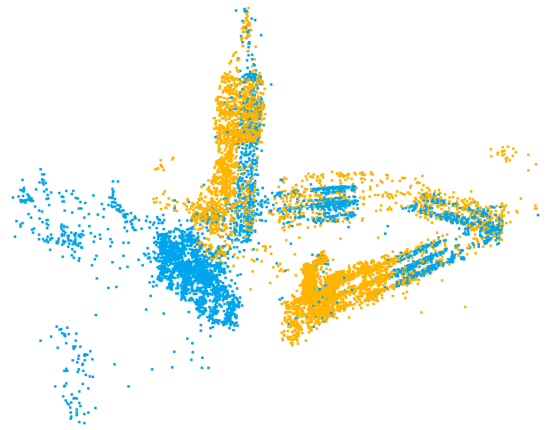} &
    \includegraphics[height=.21\textheight]{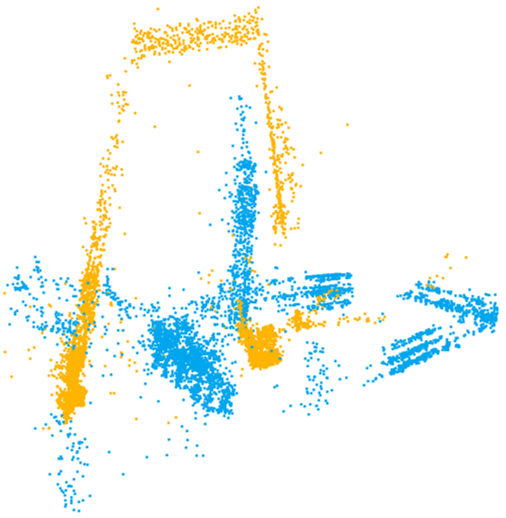} &
    \includegraphics[height=.21\textheight]{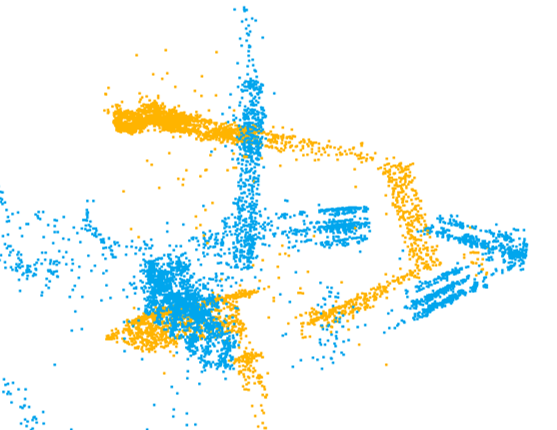} &
    \includegraphics[height=.21\textheight]{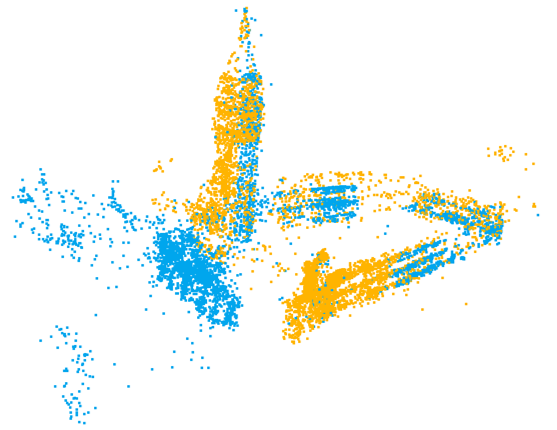} \\
    \\
    
    \end{tabular}
    }
    \caption{\textbf{Additional qualitative comparison on our dataset.} We compare our approach to previous point cloud registration methods on the Brandenburger Tor and Piazza San Marco test scenes (first and second row, respectively). Without training on our proposed SfM registration dataset (columns 2,3), previous methods are unable to produce sufficiently good matches (top two rows) to produce accurate relative pose estimation results (bottom two rows). In contrast, our proposed model \ours, trained on the proposed dataset, is able to register the scenes well.}
    \label{fig:registration_results_mega}
    
\end{figure*}

\parsection{SfM Registration Benchmark:} 
We evaluate \ours~trained from scratch on our proposed dataset against OverlapPredator~\cite{huang2021predator} and RoITr~\cite{yu2023rotation} trained on 3DMatch~\cite{zeng20173dmatch} qualitatively.
The results are presented in \cref{fig:registration_results_mega}.
We present two more pairs for two different test scenes, namely the Brandenburger Tor and Piazza San Marco test scenes.
From inspection, we can see that OverlapPredator tends to produce a high number of matches, but only a very small fraction are inliers (0.8$\%$ on average).
This results in poor pose estimations as can be seen in \cref{fig:registration_results_mega} second column bottom three rows.
On the other hand RoITr yields fewer matches, but the majority of those are still not good, see the inlier ratios presented in the figure.
Similar to OverlapPredator the registration results are not accurate enough to produce a good merging of the source and target point clouds, in yellow and blue respectively.
Finally, our \ours~is capable of finding a high number of matches, with the majority of them being inliers.

\begin{figure*}
    \centering
    \resizebox{1\linewidth}{!}{\begin{tabular}{lcccc}
    {\LARGE \textbf{Matches}} & & & & \\ %
    & {\Large Input} & {\Large OverlapPredator~\cite{huang2021predator} (3DM)} & {\Large RoITr~\cite{yu2023rotation} (3DM)} & {\Large \ours~(Mega)} \\ 
    &
    \includegraphics[height=.23\textheight]{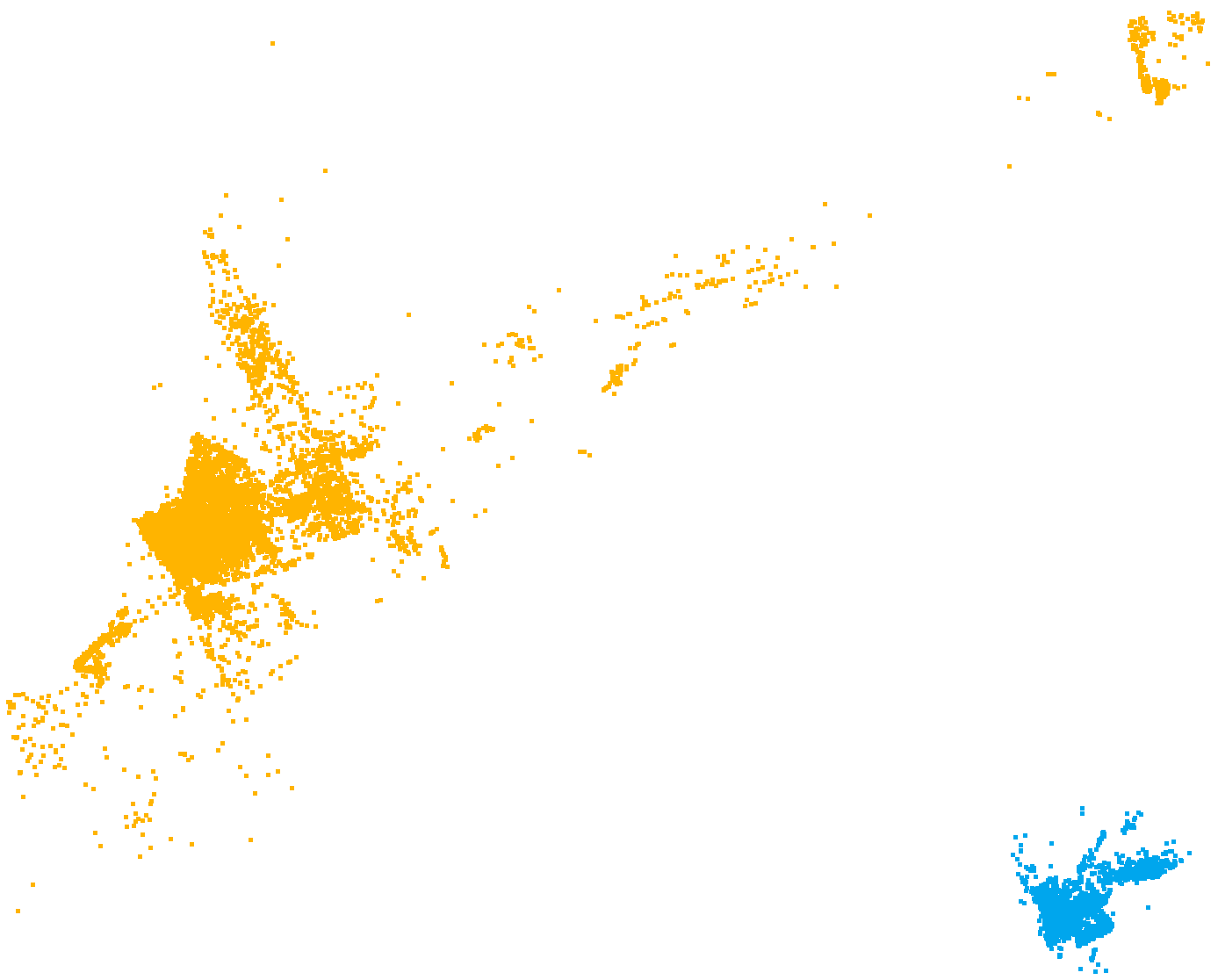} &
    \includegraphics[height=.23\textheight]{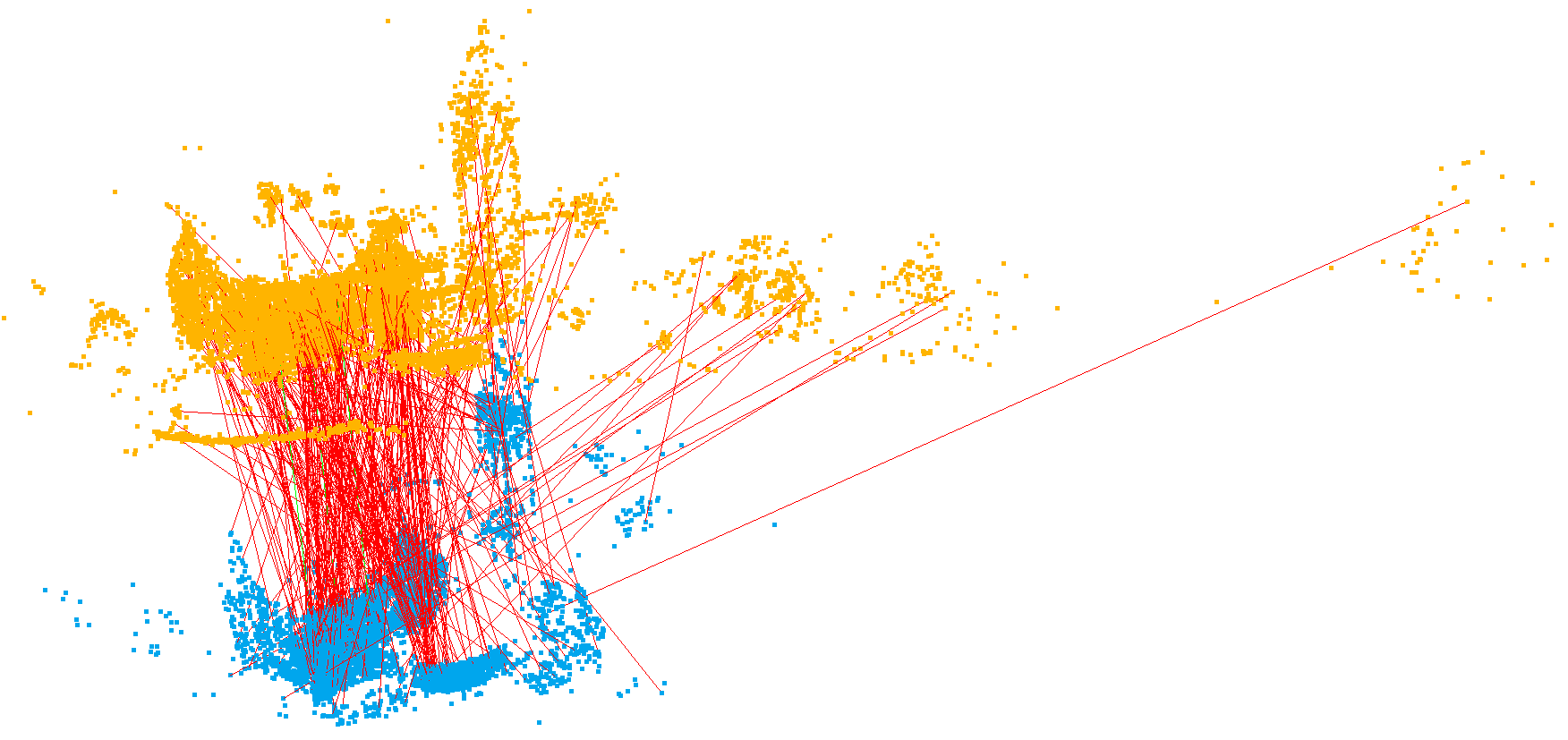} &
    \includegraphics[height=.23\textheight]{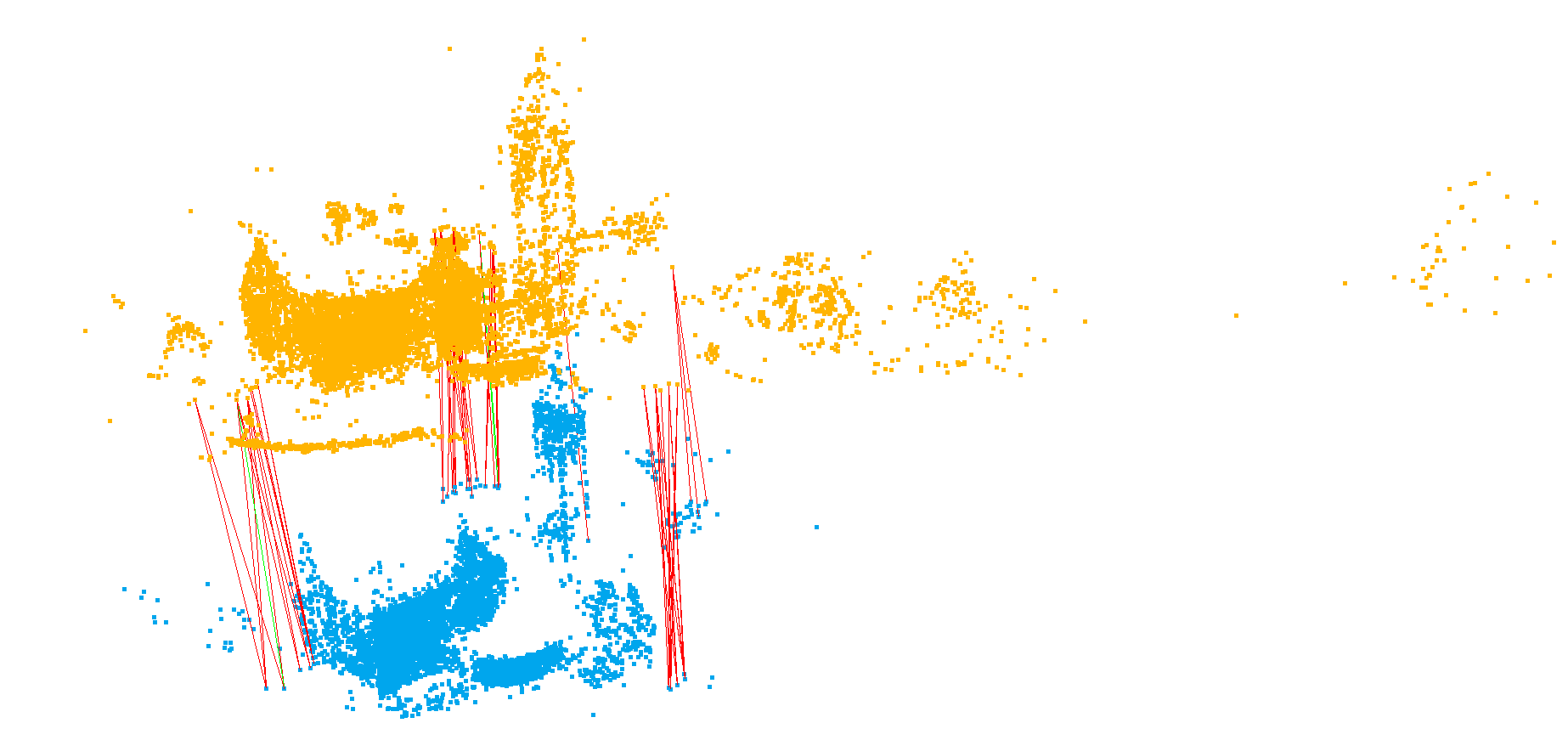} &
    \includegraphics[height=.23\textheight]{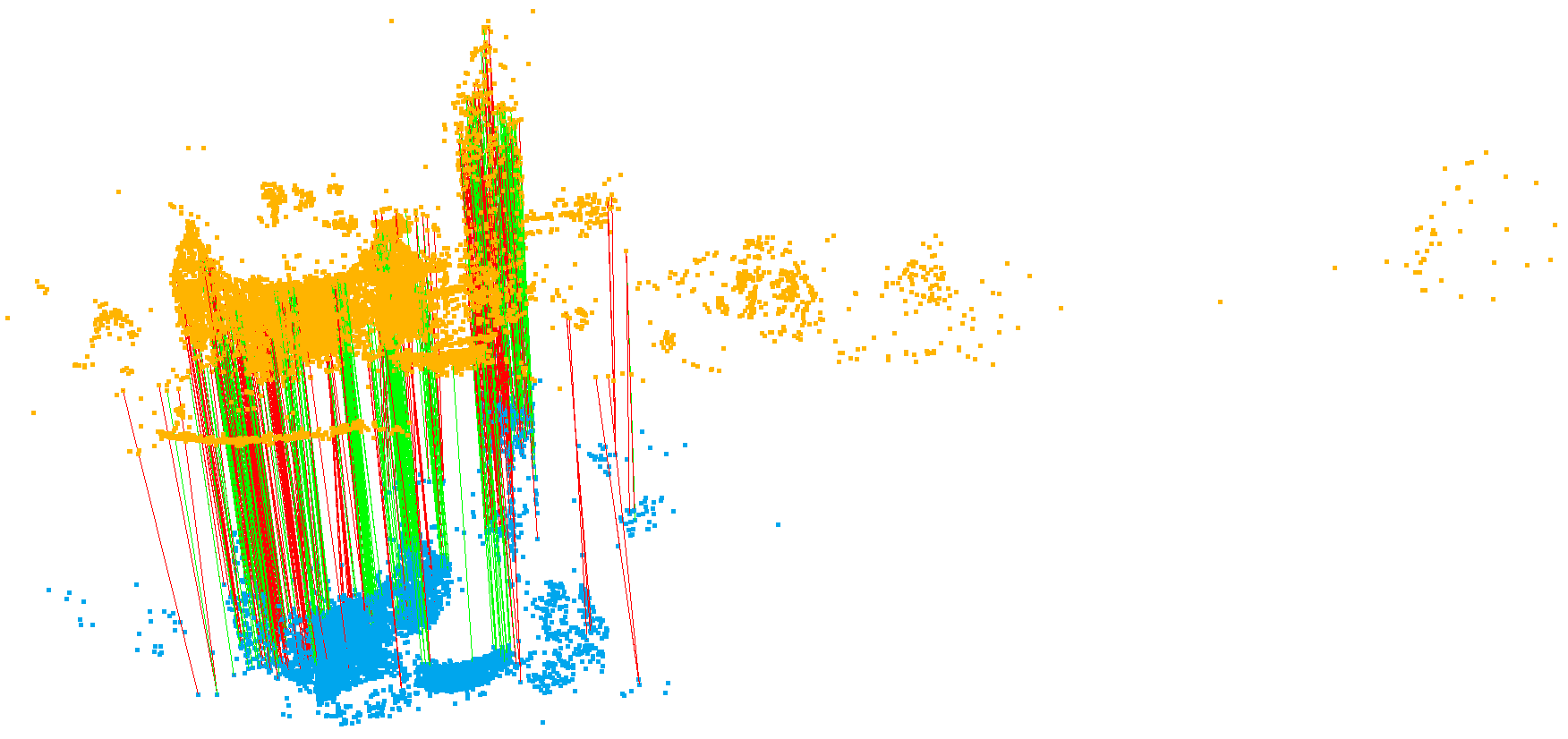} \\
    {\Large IR} &  & 1.1 & 4.4 & 57.2 \\ \\
    &
    \includegraphics[height=.23\textheight]{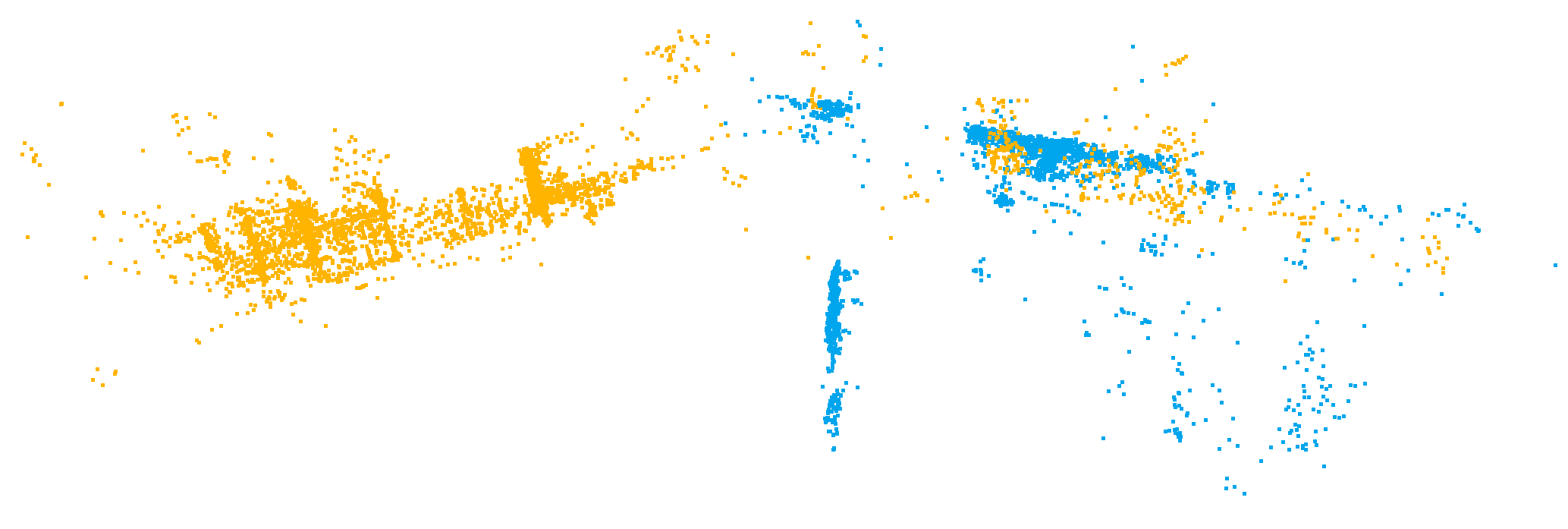} &
    \includegraphics[height=.23\textheight]{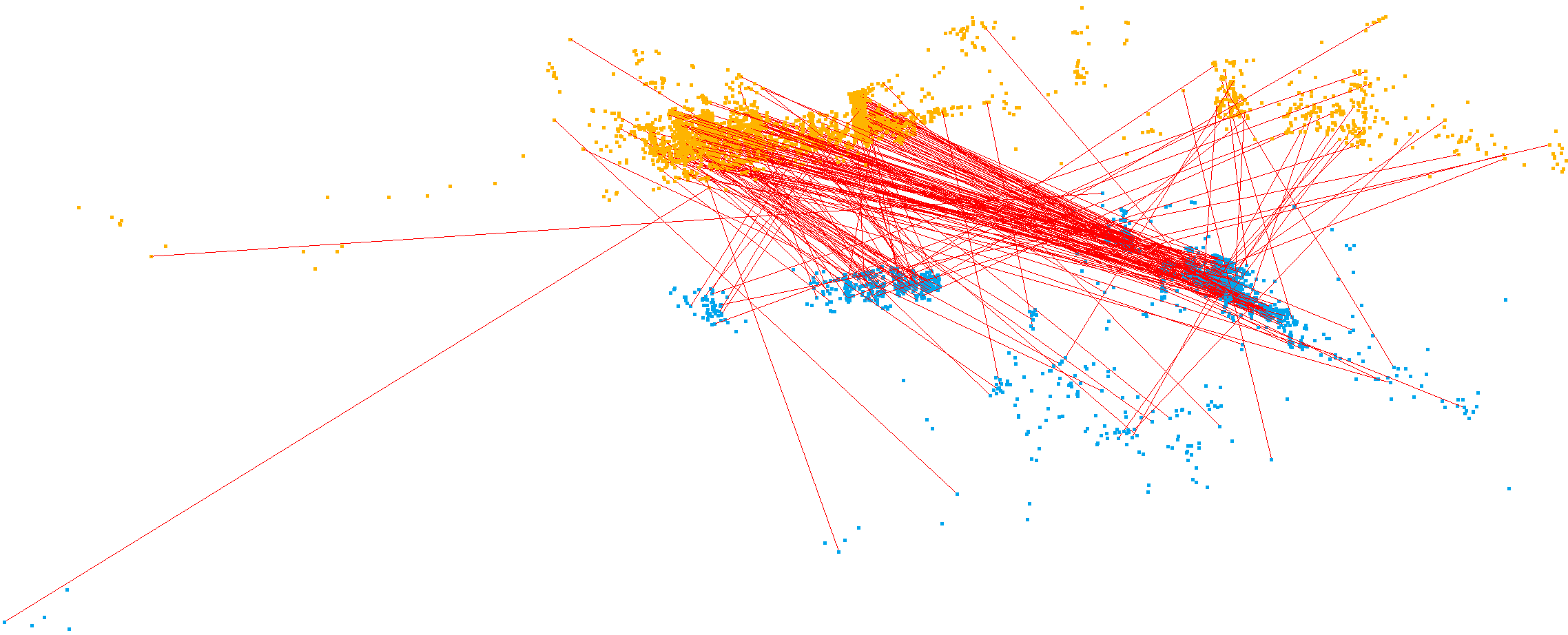} &
    \includegraphics[height=.23\textheight]{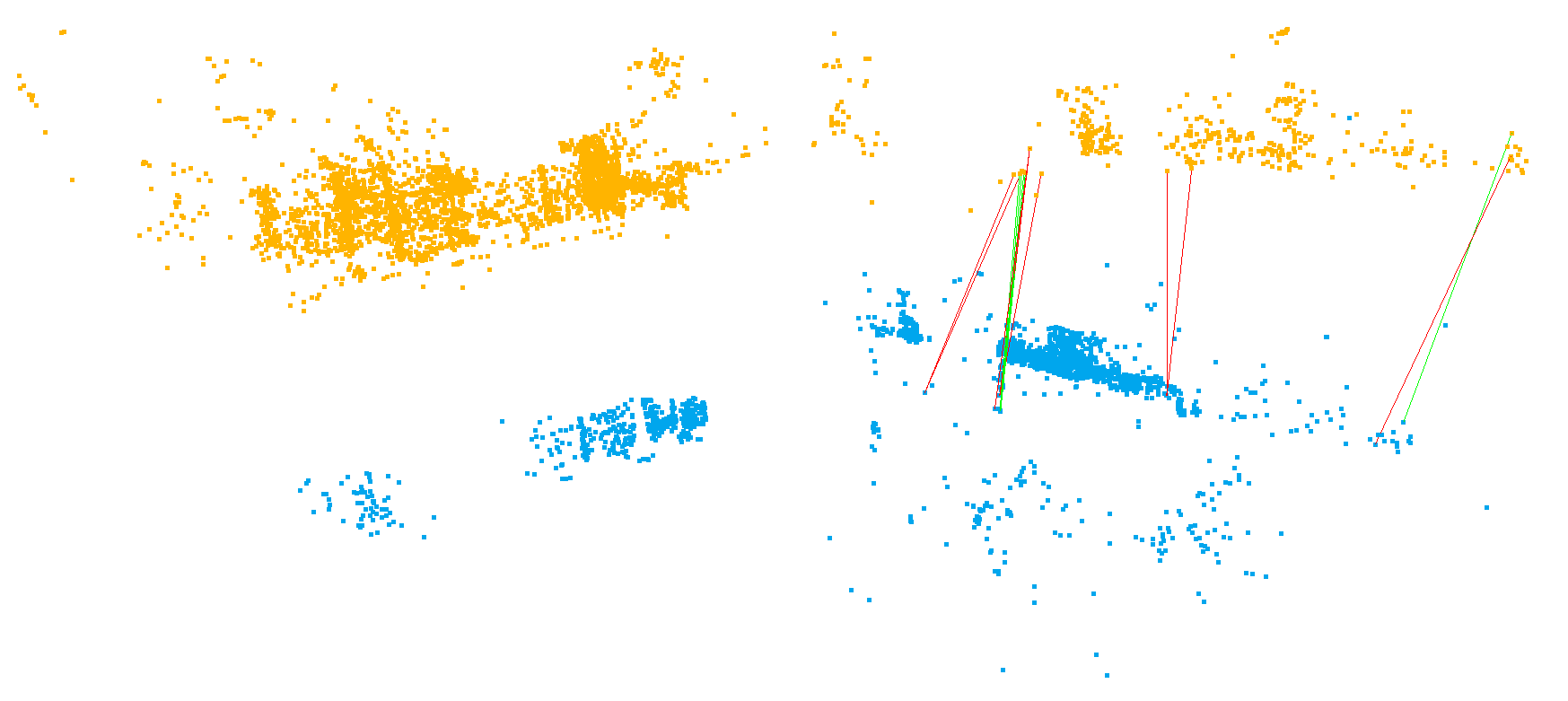} &
    \includegraphics[height=.23\textheight]{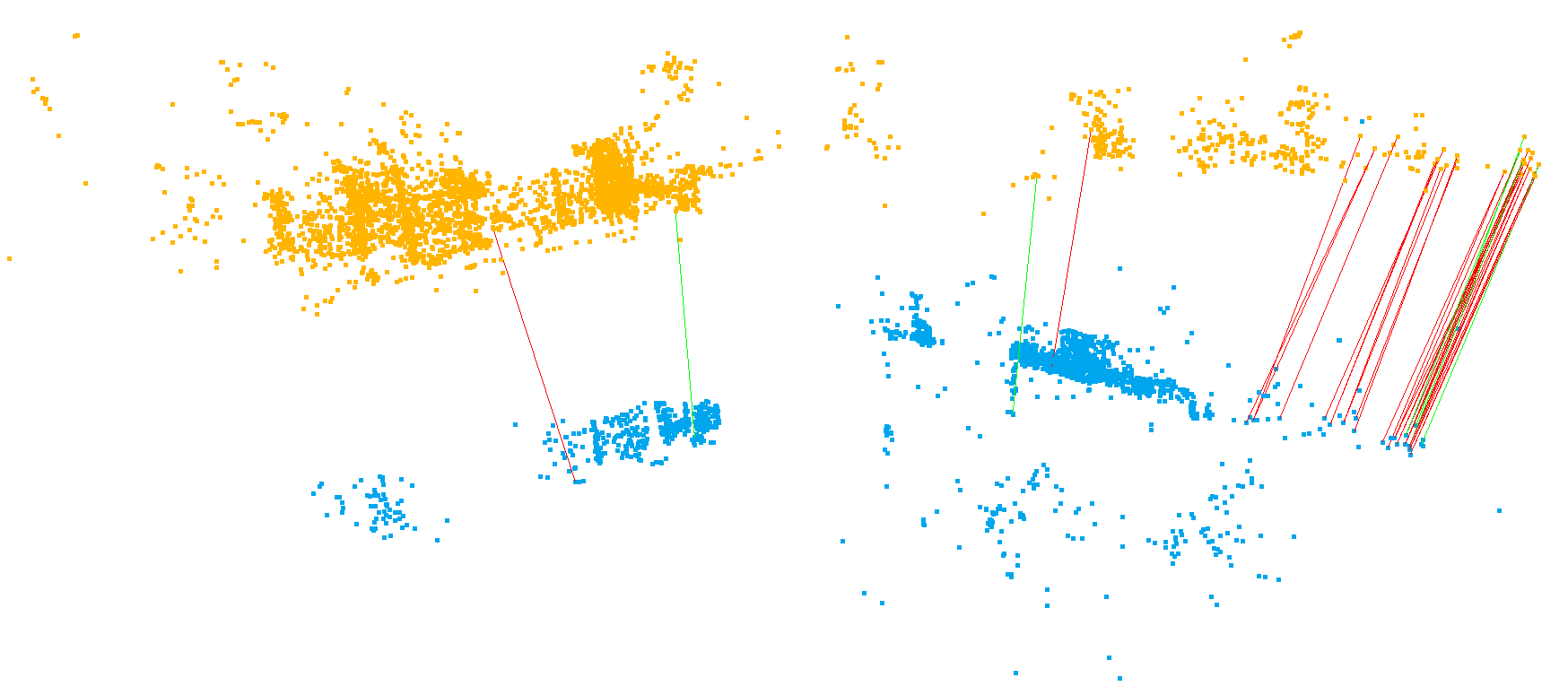} \\
    {\Large IR} & & 0.0 & 41.2 & 15.9\\ \\
    &
    \includegraphics[height=.23\textheight]{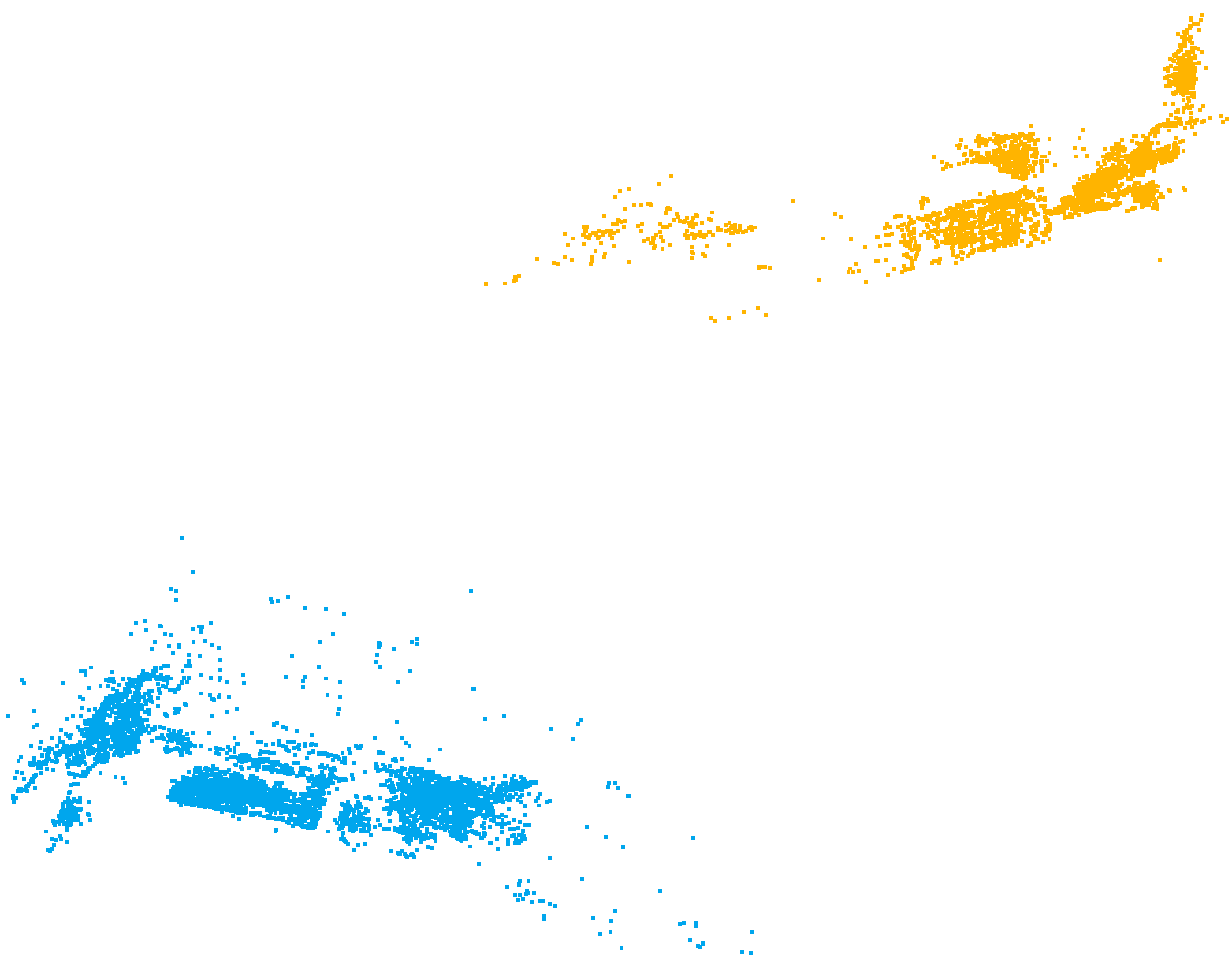} &
    \includegraphics[height=.23\textheight]{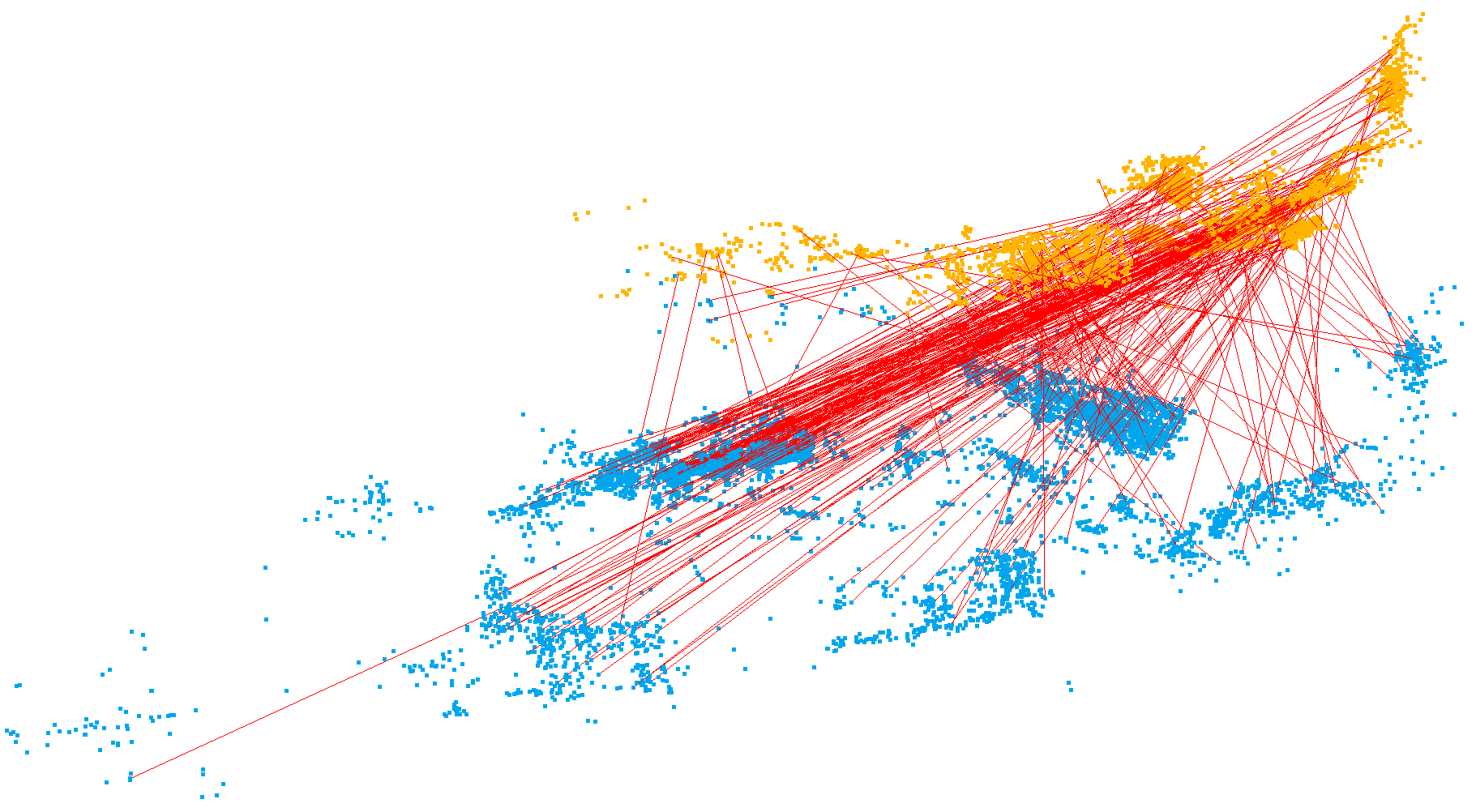} &
    \includegraphics[height=.23\textheight]{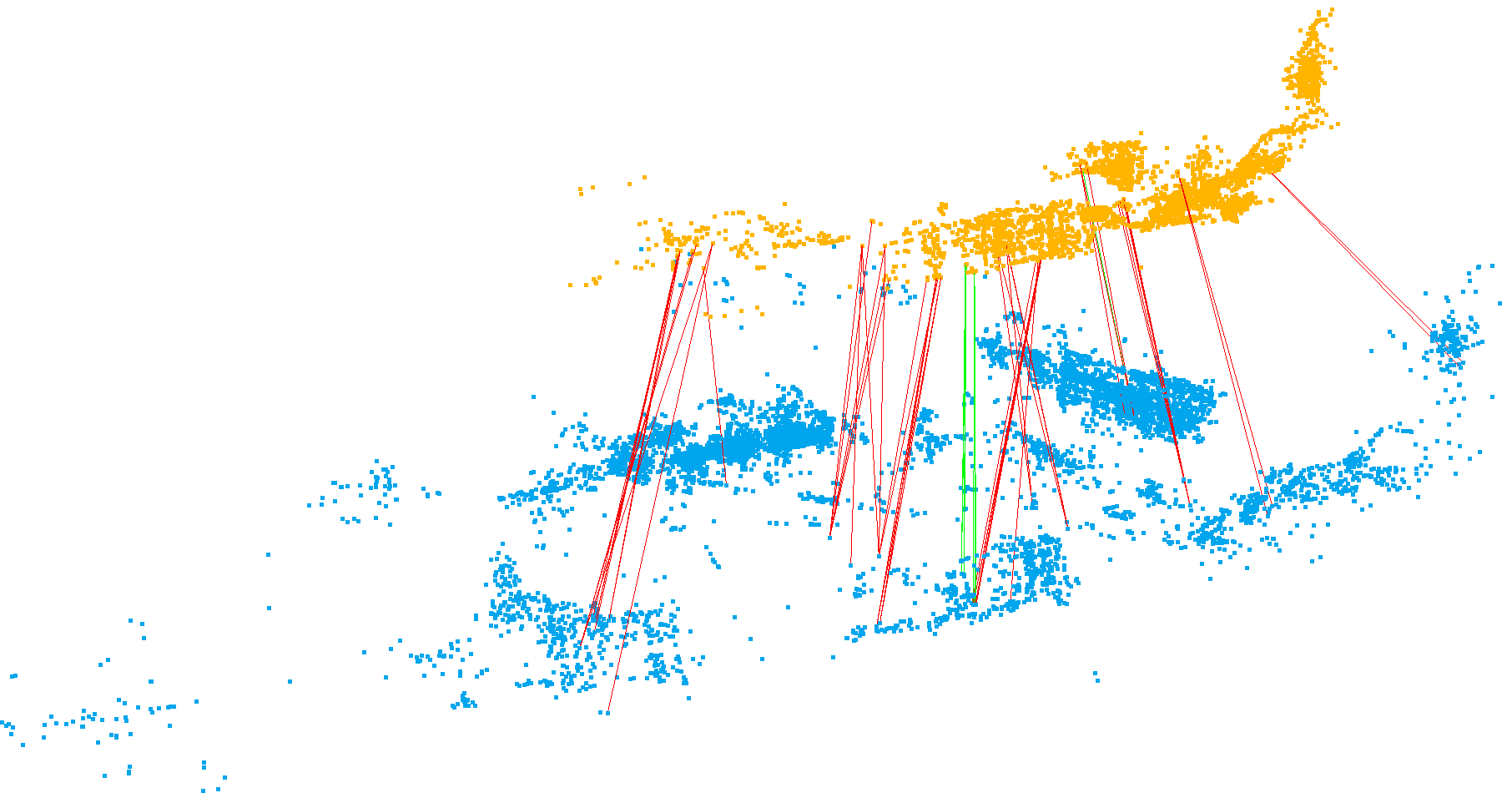} &
    \includegraphics[height=.23\textheight]{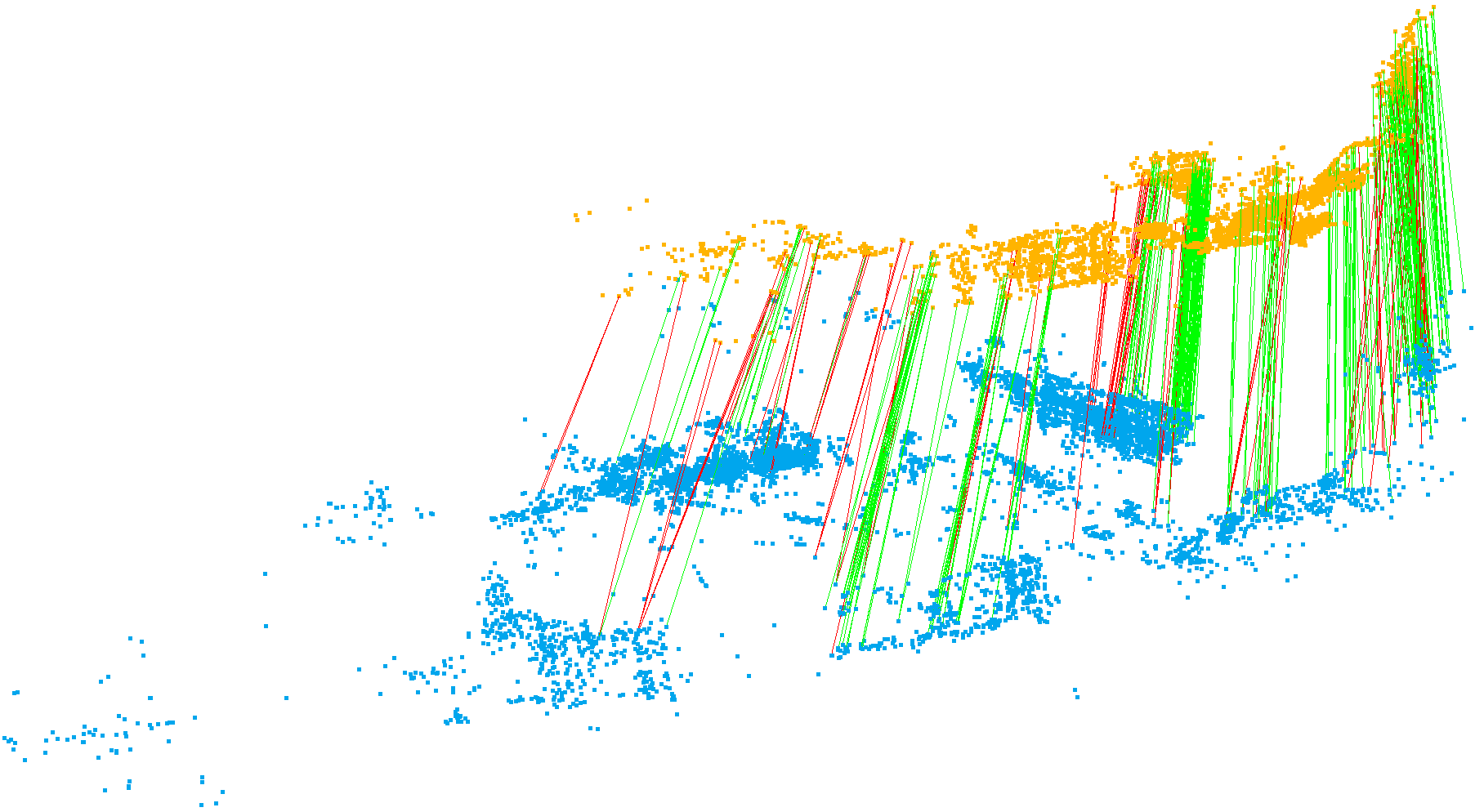} \\
    {\Large IR} & & 0.0 & 18.1 & 75.9\\ \\
    \midrule \\
    {\LARGE \textbf{Registration}} & & & & \\
    & {\large GT} & {\Large OverlapPredator~\cite{huang2021predator} (3DM)} & {\Large RoITr~\cite{yu2023rotation} (3DM)} & {\Large \ours~(Mega)} \\ 
    &
    \includegraphics[height=.23\textheight]{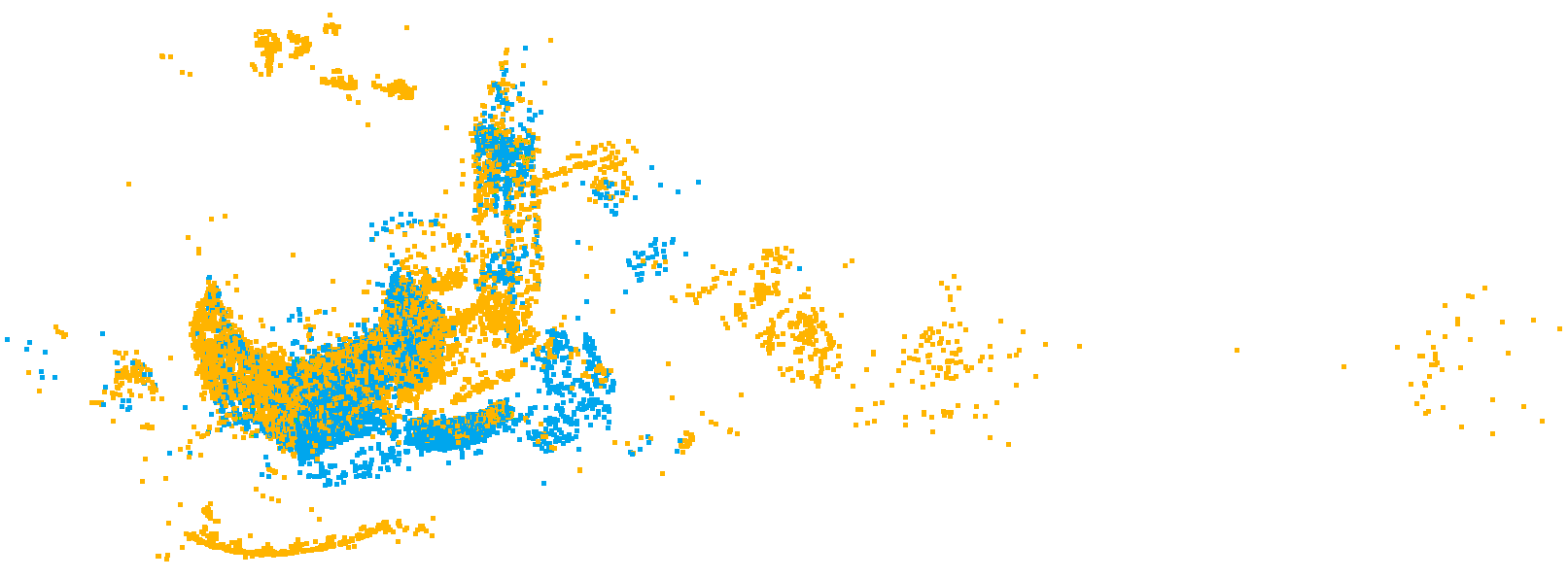} &
    \includegraphics[height=.23\textheight]{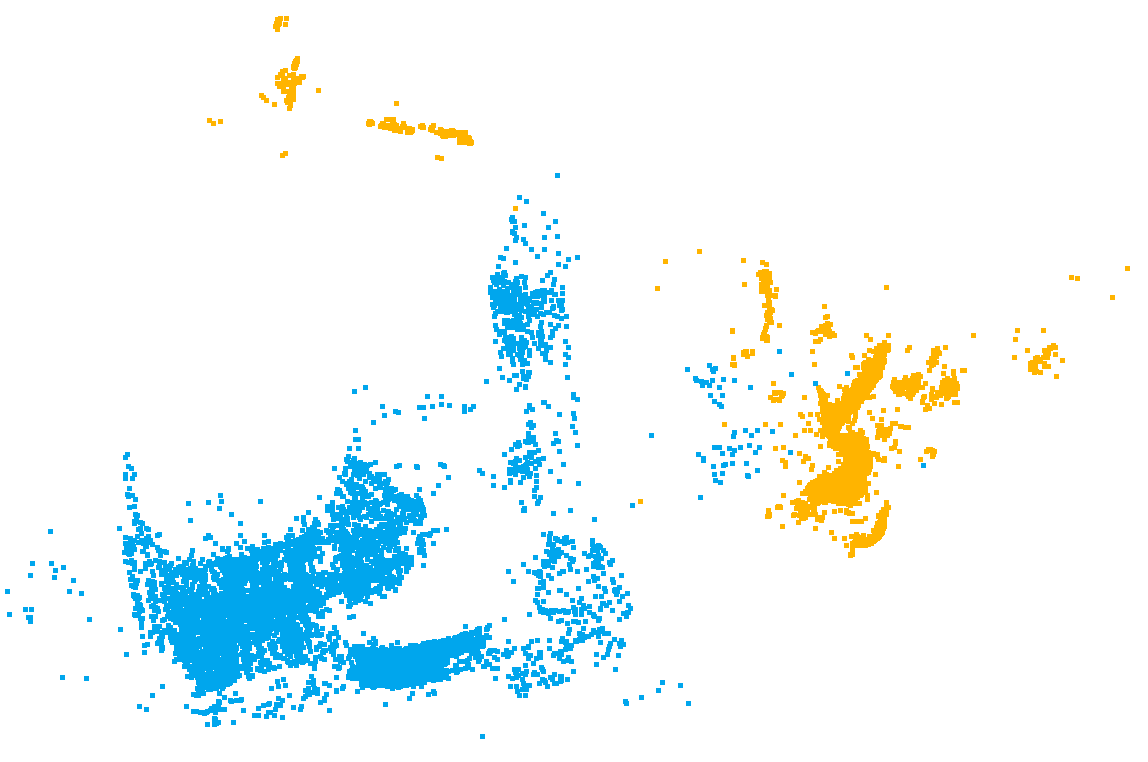} &
    \includegraphics[height=.23\textheight]{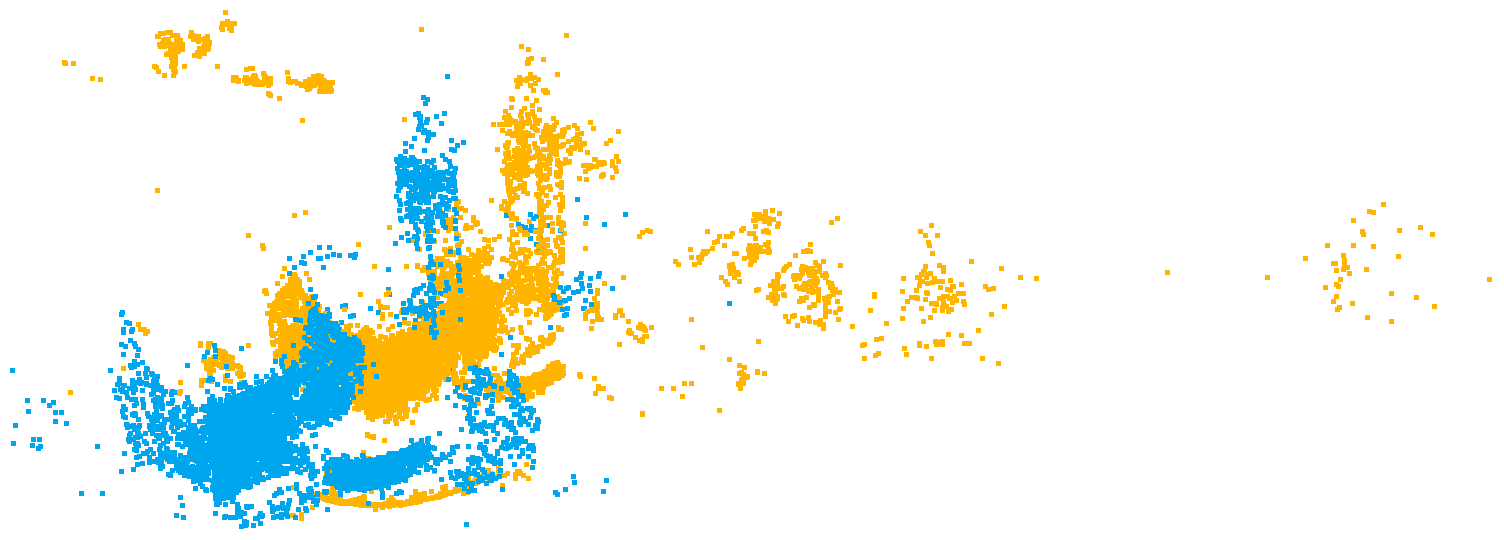} &
    \includegraphics[height=.23\textheight]{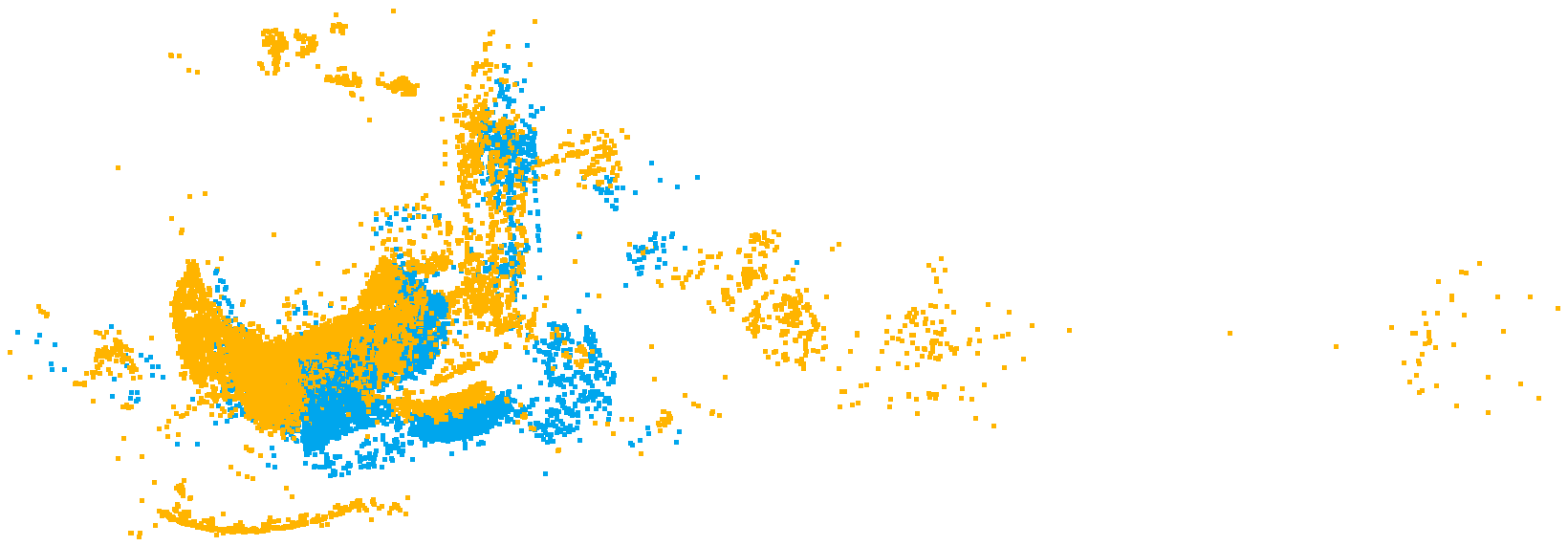} \\
    {\Large ($\epsilon_R,\epsilon_{t}$)} &  & (80.5$^\circ$,38.92) & (6.7$^\circ$,3.52) & (1.1$^\circ$,0.54)\\ \\
    &
    \includegraphics[height=.23\textheight]{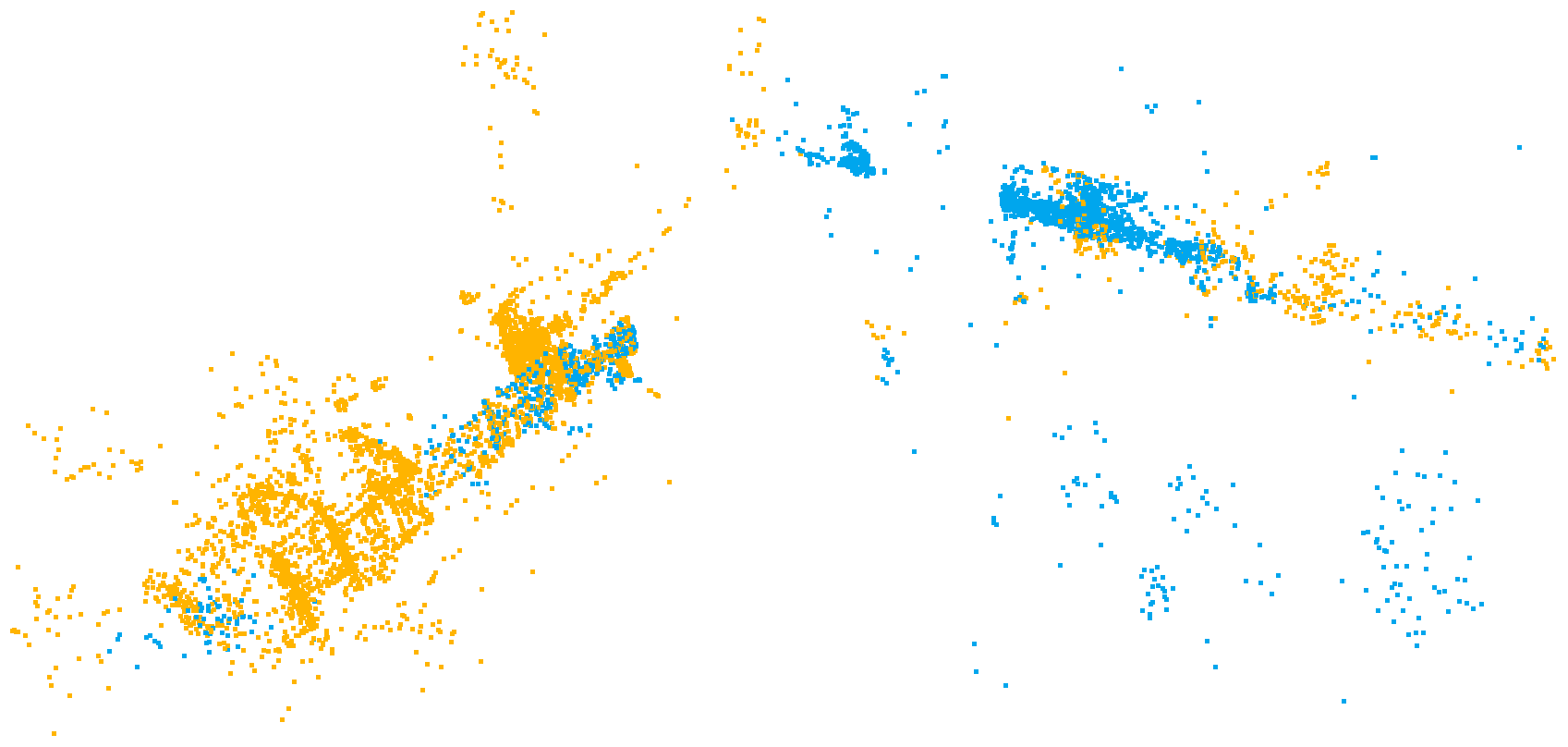} &
    \includegraphics[height=.23\textheight]{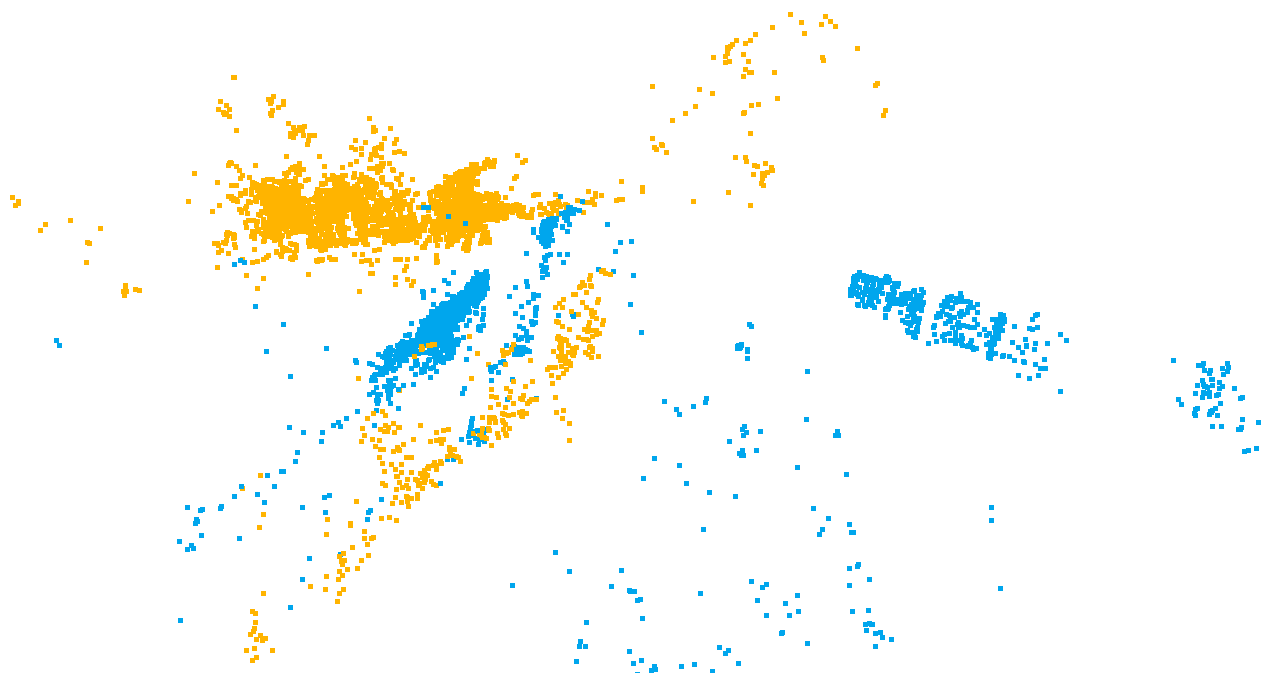} &
    \includegraphics[height=.23\textheight]{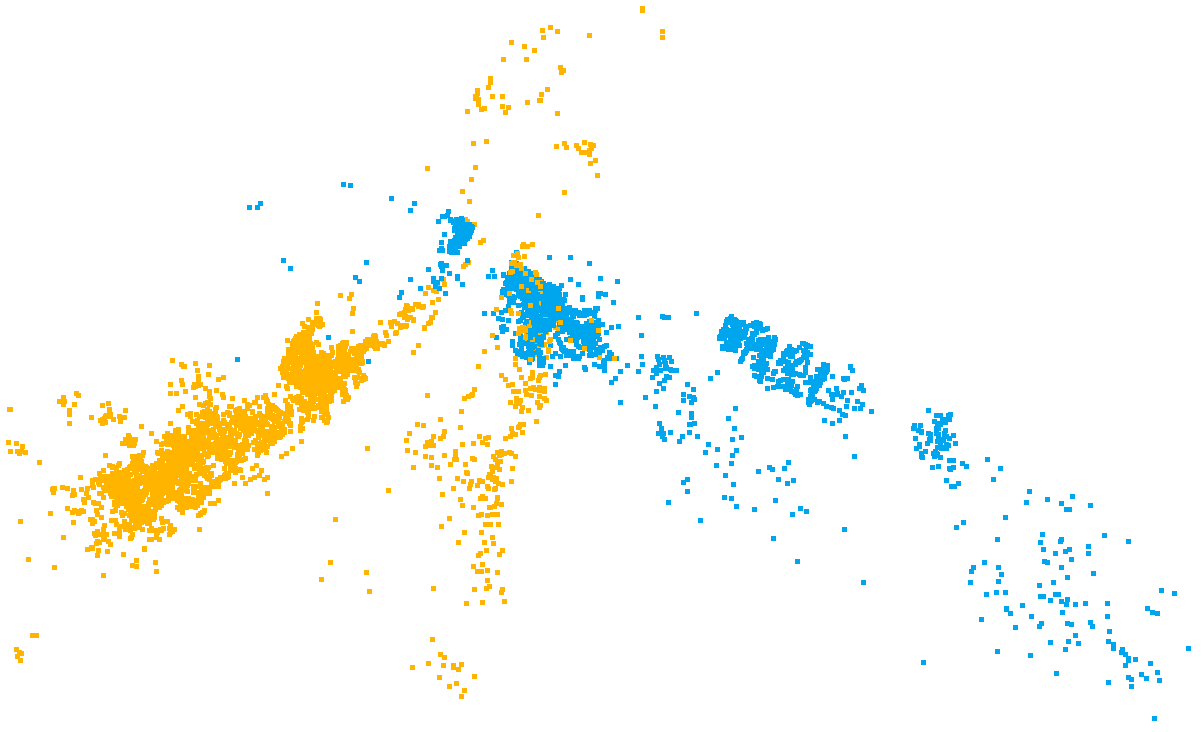} &
    \includegraphics[height=.23\textheight]{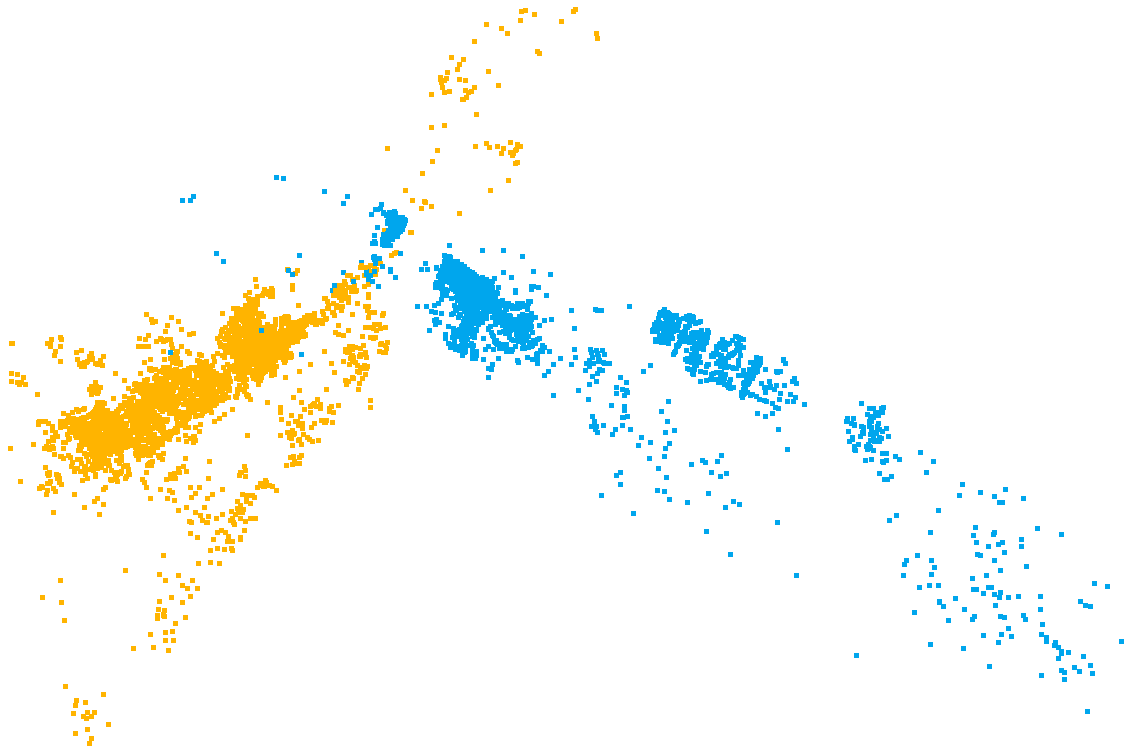} \\
    {\Large ($\epsilon_R,\epsilon_{t}$)} &  & (158.3$^\circ$,5.13) & (115.2$^\circ$,1.04) & (31.4$^\circ$,0.91)\\ \\
    &
    \includegraphics[height=.23\textheight]{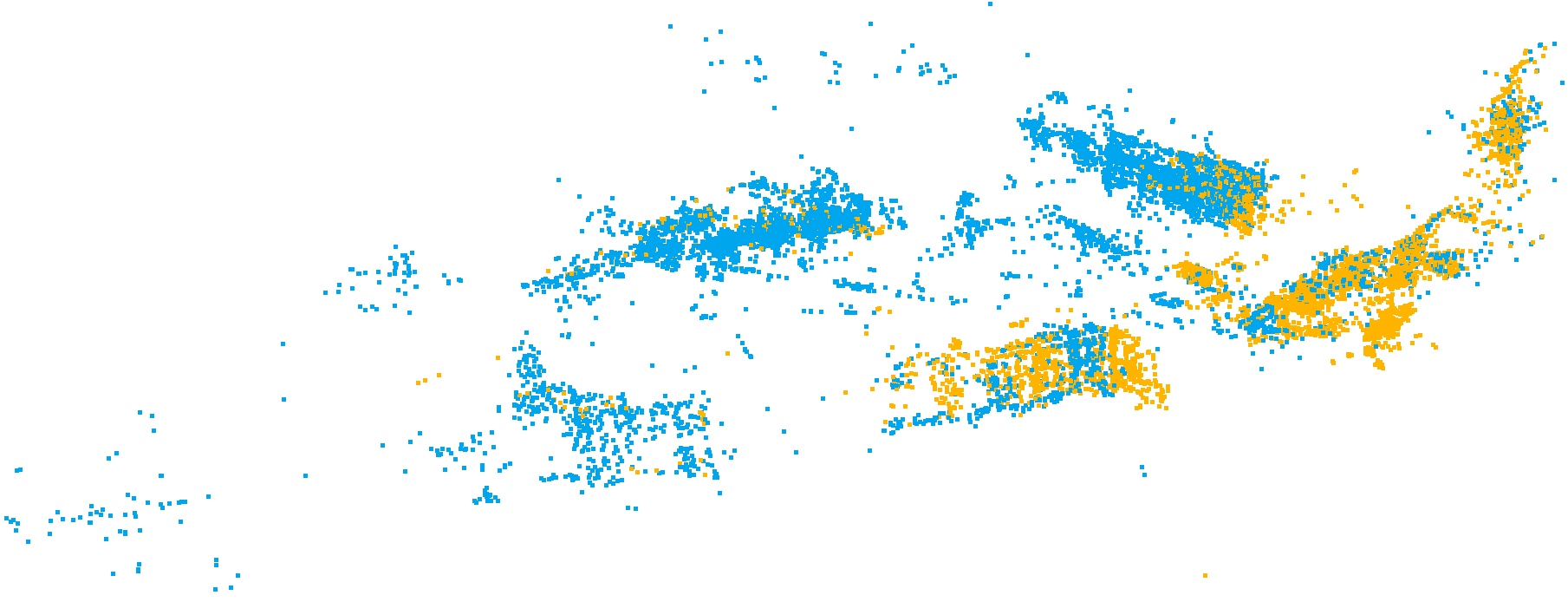} &
    \includegraphics[height=.23\textheight]{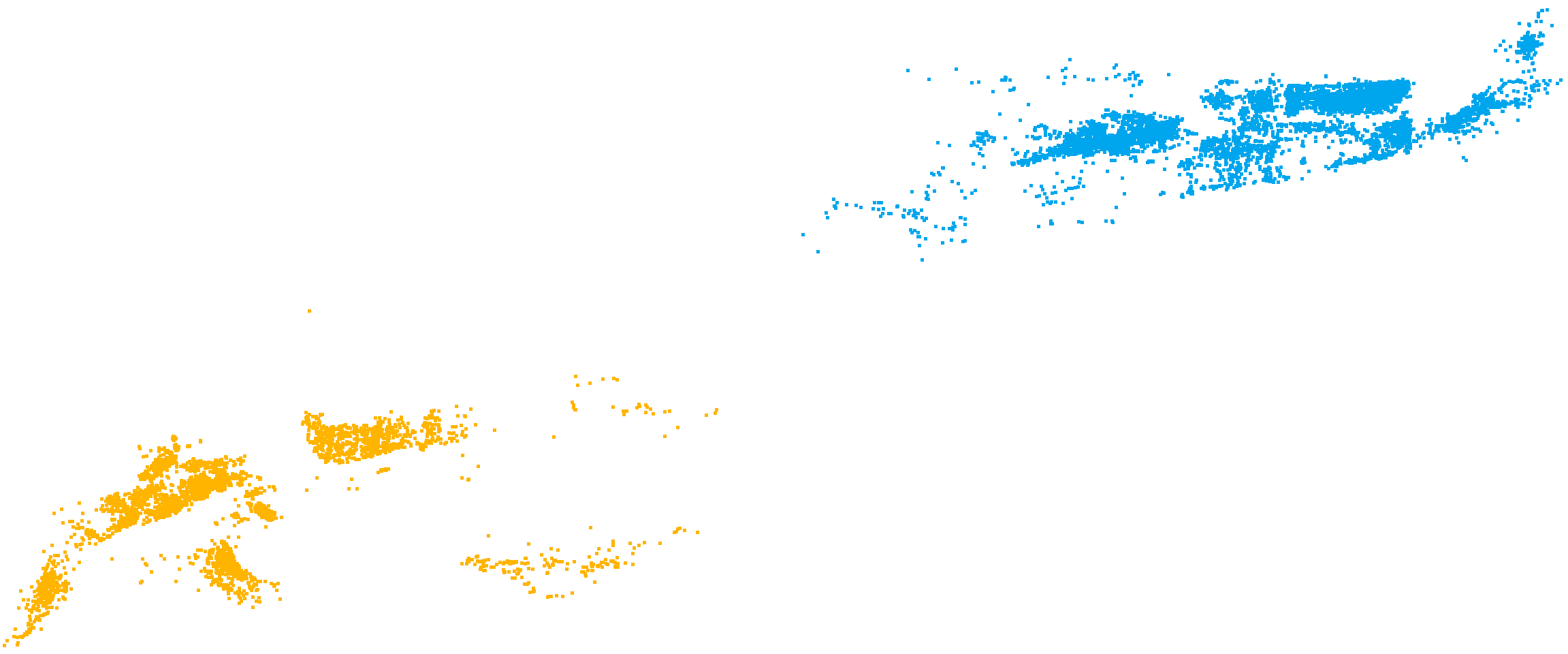} &
    \includegraphics[height=.23\textheight]{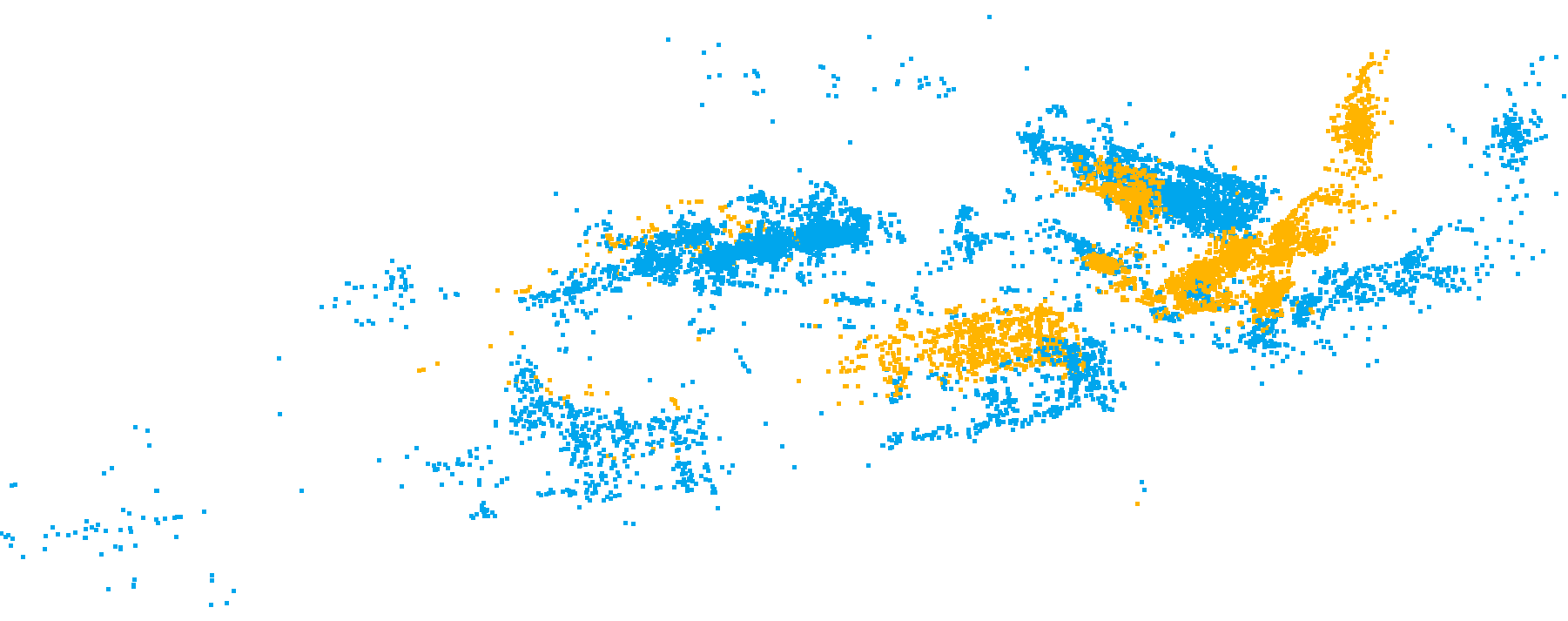} &
    \includegraphics[height=.23\textheight]{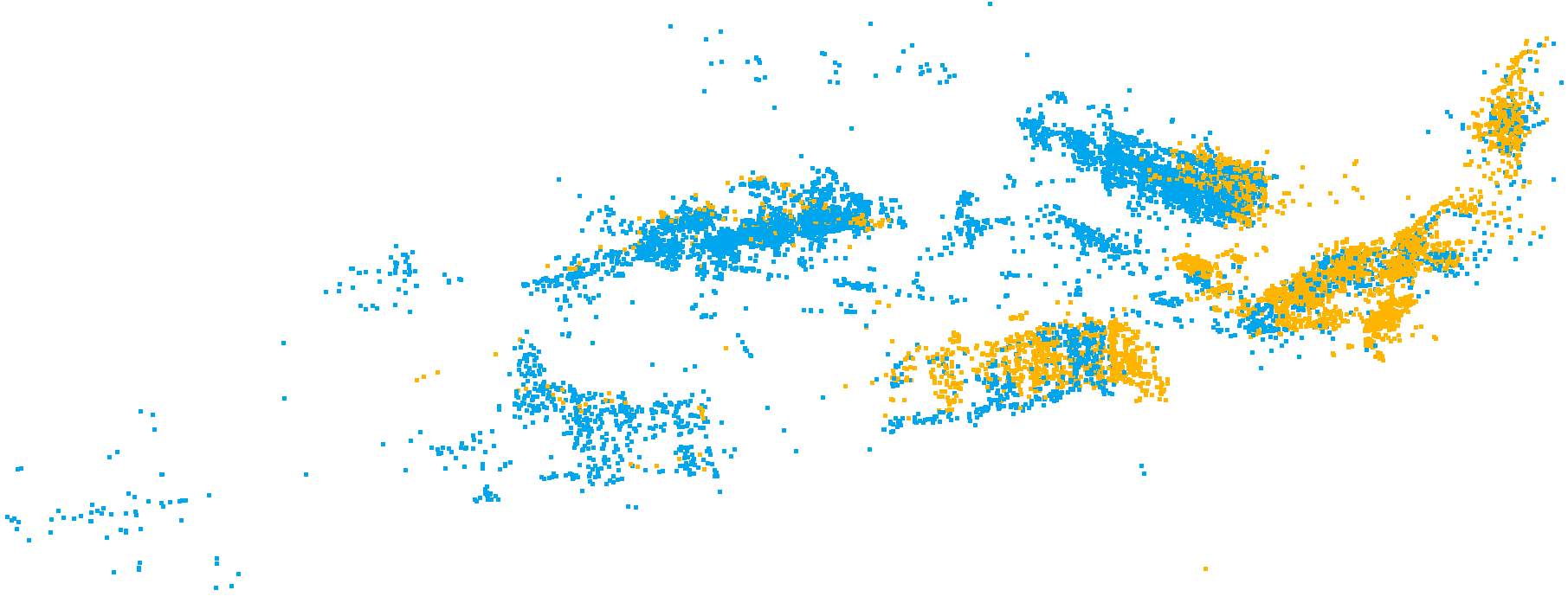} \\
    {\Large ($\epsilon_R,\epsilon_{t}$)} &  & (176.0$^\circ$,10.59) & (5.7$^\circ$,0.45) & (0.5$^\circ$,0.05)\\ \\
    
    \end{tabular}
    }
    \caption{\textbf{Qualitative comparison Quad6k~\cite{lowe2004distinctive}.} We compare our approach to previous point cloud registration methods on three test scenes (first to third row, respectively). Without training on our proposed SfM registration dataset (columns 2,3), previous methods are unable to produce sufficiently good matches (top three rows) and accurate relative pose estimation results (bottom three rows). In contrast, our proposed model \ours, trained on the proposed dataset, is able to find better matches and hence register the scenes well. The exception is the second scene, where our model struggles to find matches and fails to register the point clouds. The source and target point clouds are depicted in yellow and blue, respectively.}
    \label{fig:registration_results_qaud6k}
    
\end{figure*}

\parsection{Quad6k~\cite{lowe2004distinctive}:}
To evaluate the generalization of \ours, we evaluate it and two baselines OverlapPredator~\cite{huang2021predator} and RoITr~\cite{yu2023rotation} trained on 3DMatch~\cite{zeng20173dmatch}.
The results are presented in \cref{fig:registration_results_qaud6k}.
This dataset is more challenging and it consists mainly of a square surrounded by the facade of several buildings.
The performance of the baselines is similar to what was observed in our SfM Registration Benchmark test scenes.
When looking at the performance of \ours, we can see that when the pairs refer to distinctive structures, like the tower in both the first and third pairs of \cref{fig:registration_results_qaud6k}.
The model is capable of finding good matches and find an accurate pose to register the point clouds.
However, it struggles to find matches for the second pair, which contains the facade of two buildings with repetitive and symmetric structure, leading to a failure in the registration.

\begin{figure*}
    \centering
    \resizebox{1\linewidth}{!}{\begin{tabular}{lcccc}
    & {\Large Input} & {\Large Matches} & {\Large Registration} & {\Large GT} \\ 
    \raisebox{1\normalbaselineskip}[0pt][0pt]{\rotatebox[origin=l]{90}{\tt \large Great Court}} 
    &
    \includegraphics[height=.21\textheight]{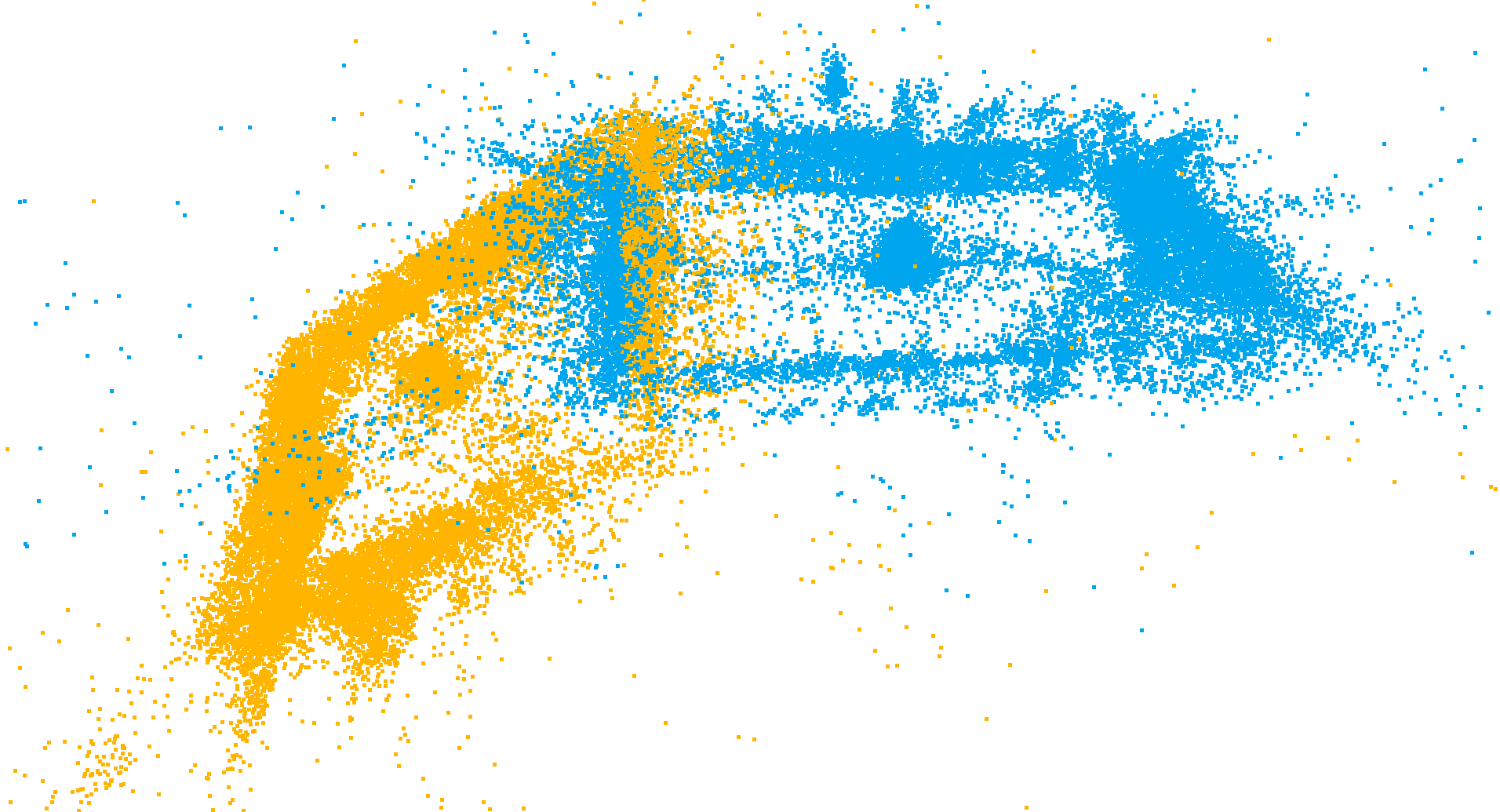} &
    \includegraphics[height=.21\textheight]{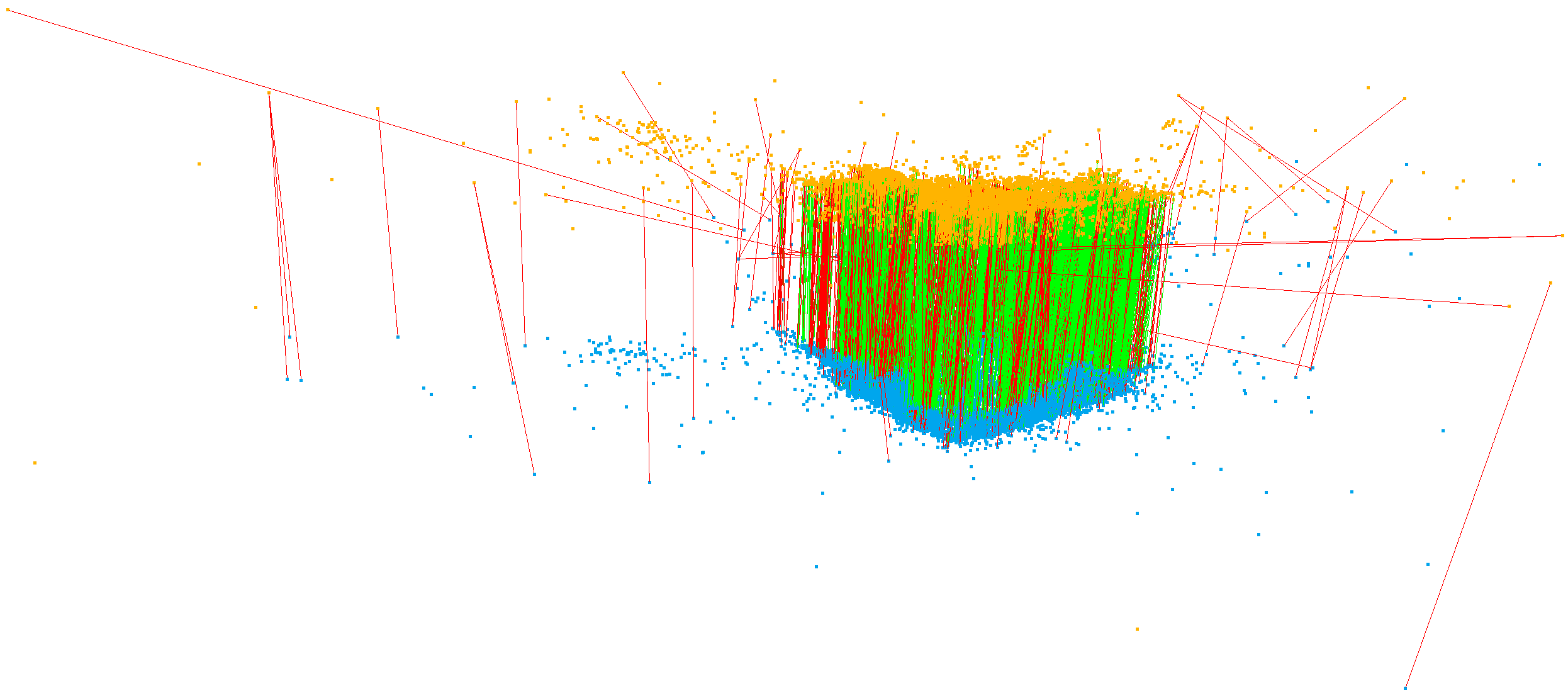} &
    \includegraphics[height=.21\textheight]{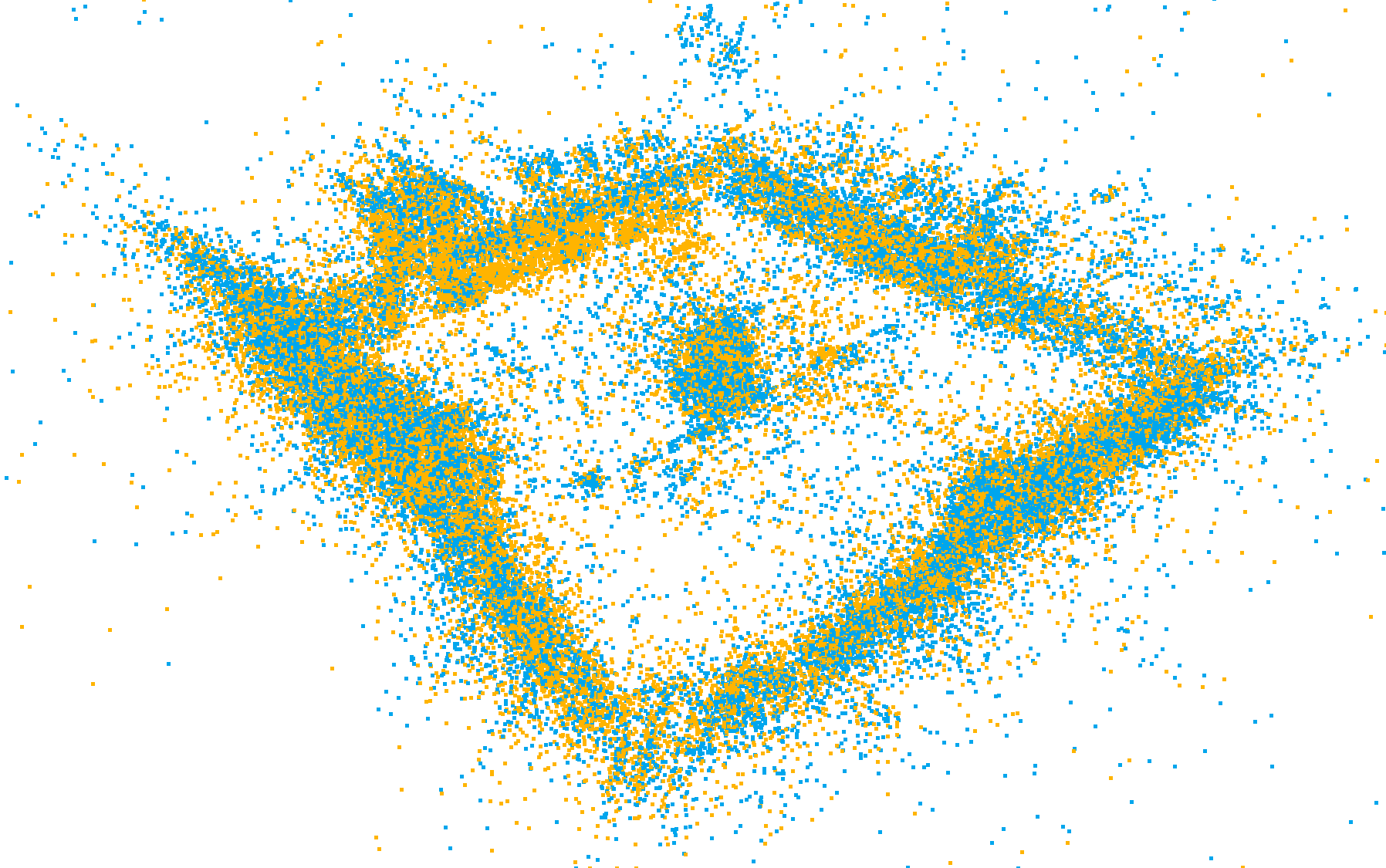} &
    \includegraphics[height=.21\textheight]{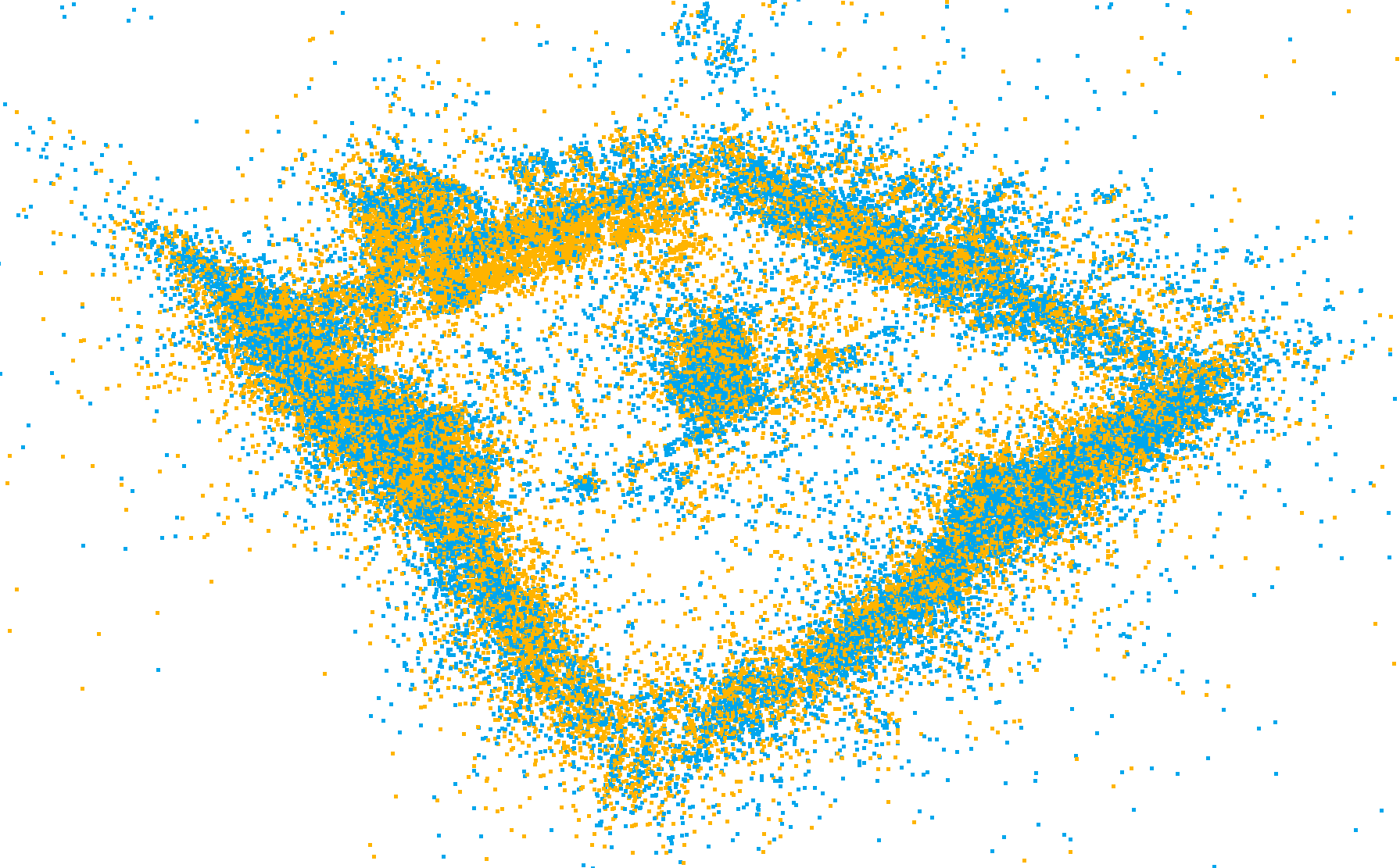} \\
    
    \raisebox{1\normalbaselineskip}[0pt][0pt]{\rotatebox[origin=l]{90}{\tt \large Kings College}} &
    \includegraphics[height=.21\textheight]{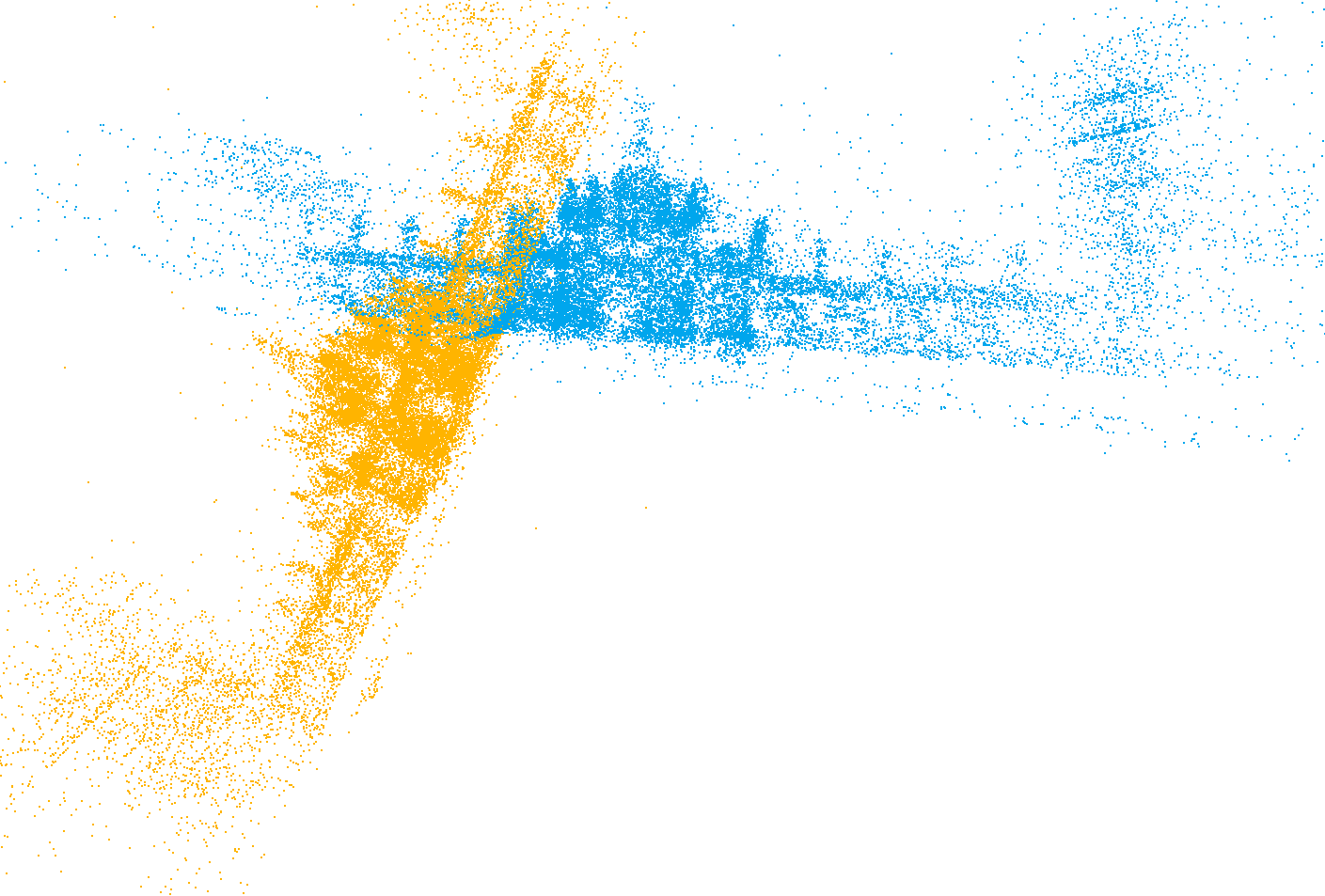} &
    \includegraphics[height=.21\textheight]{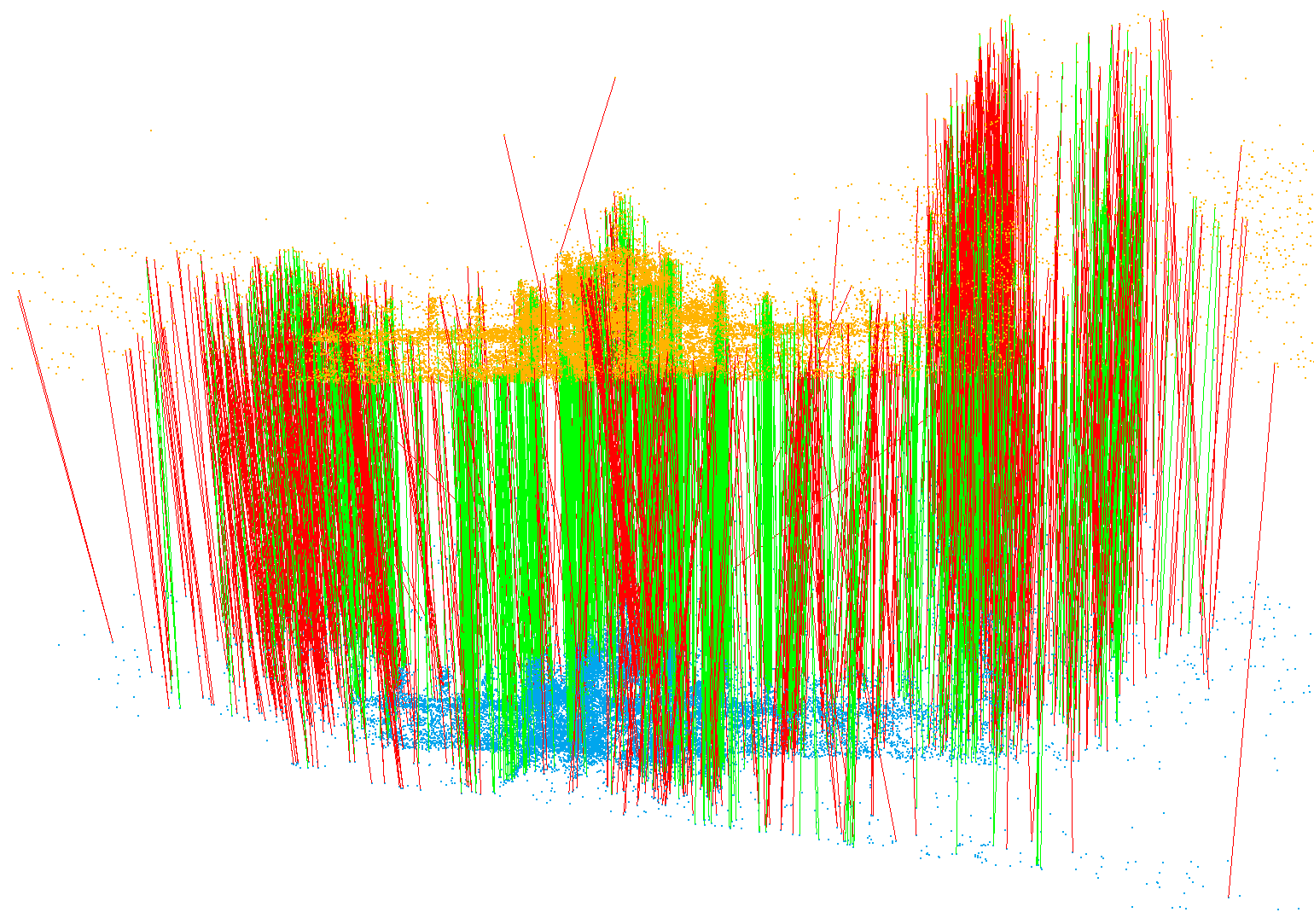} &
    \includegraphics[height=.21\textheight]{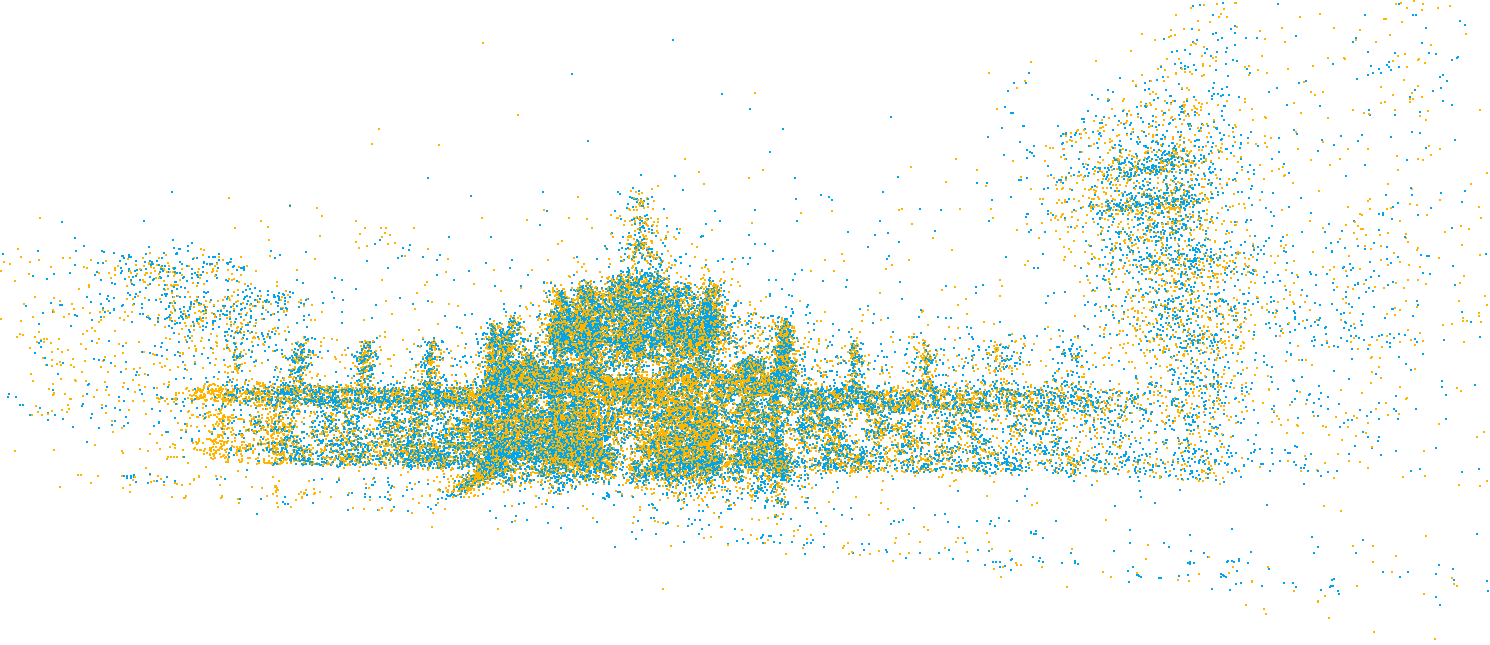} &
    \includegraphics[height=.21\textheight]{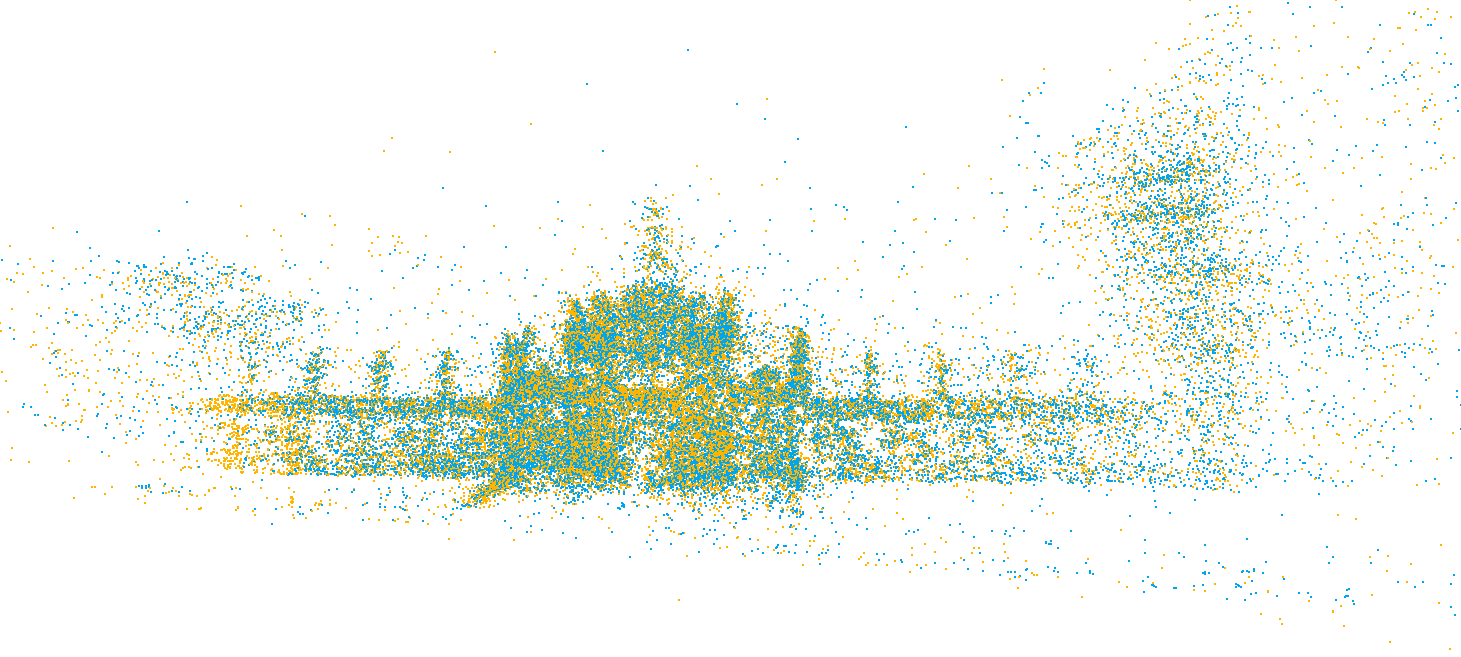} \\
   
    \raisebox{1\normalbaselineskip}[0pt][0pt]{\rotatebox[origin=l]{90}{\tt \large Old Hospital}}&
    \includegraphics[height=.21\textheight]{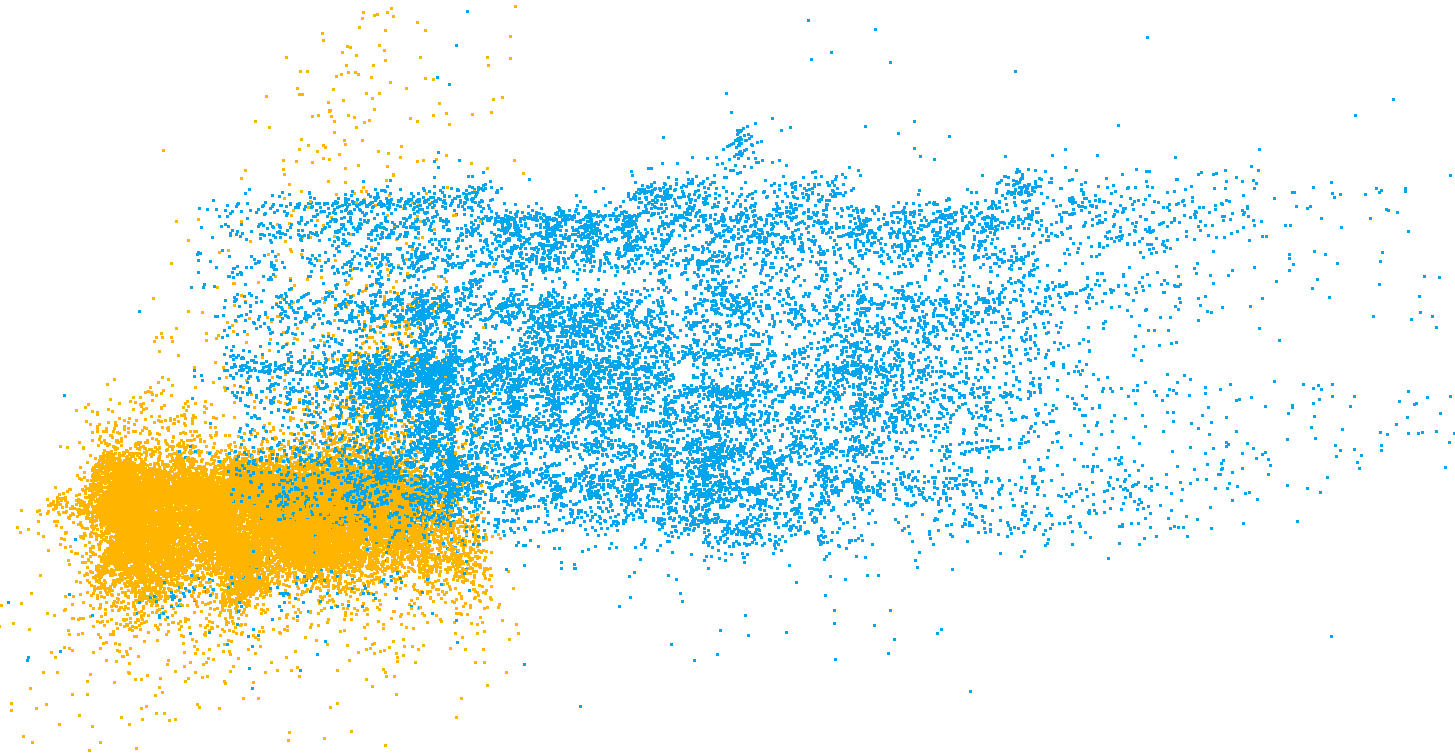} &
    \includegraphics[height=.21\textheight]{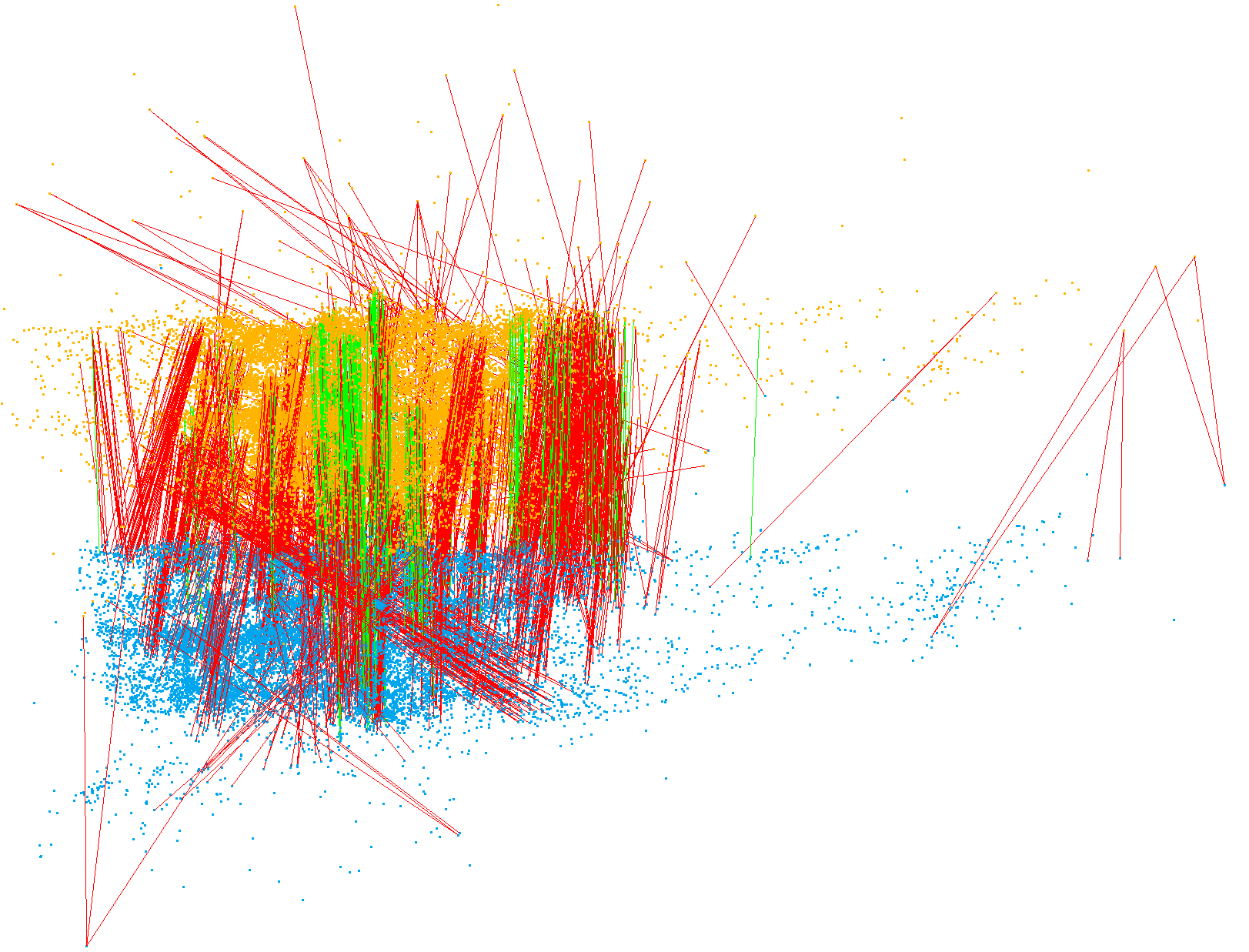} &
    \includegraphics[height=.21\textheight]{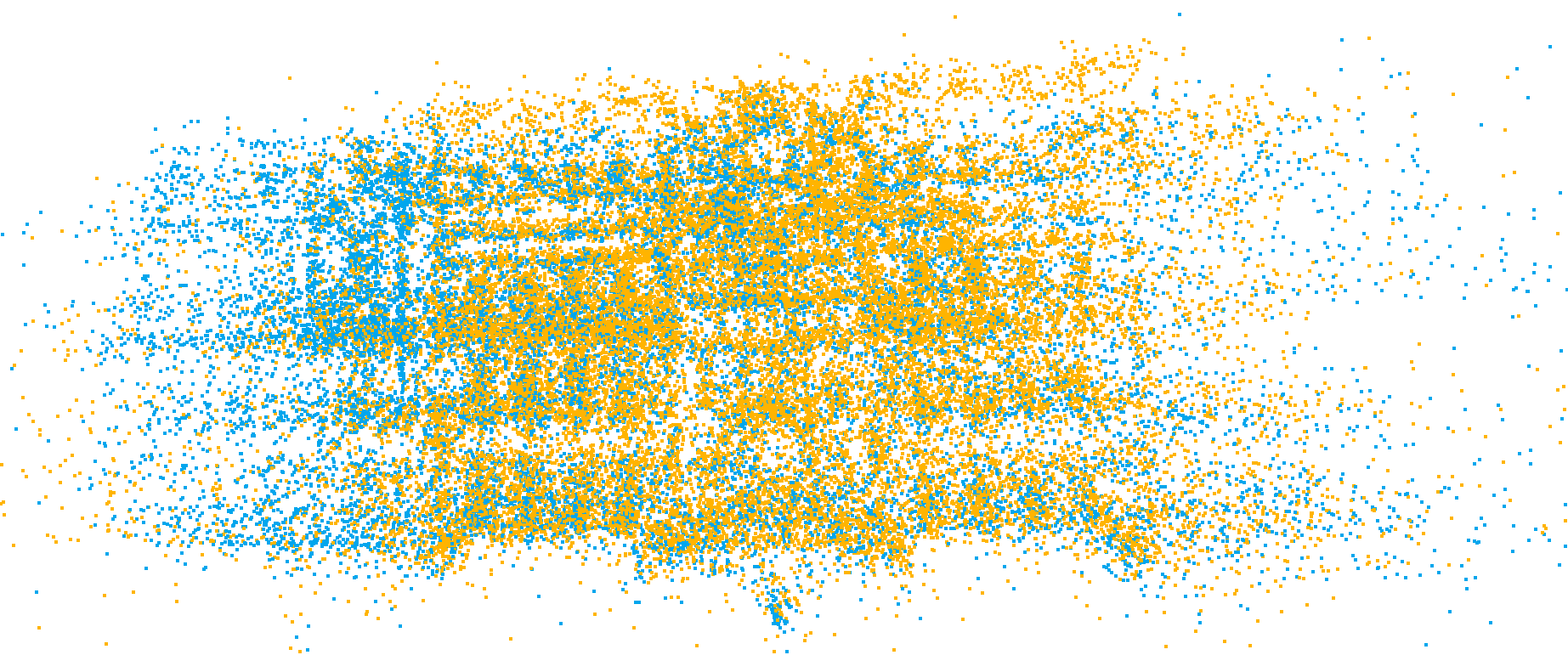} &
    \includegraphics[height=.21\textheight]{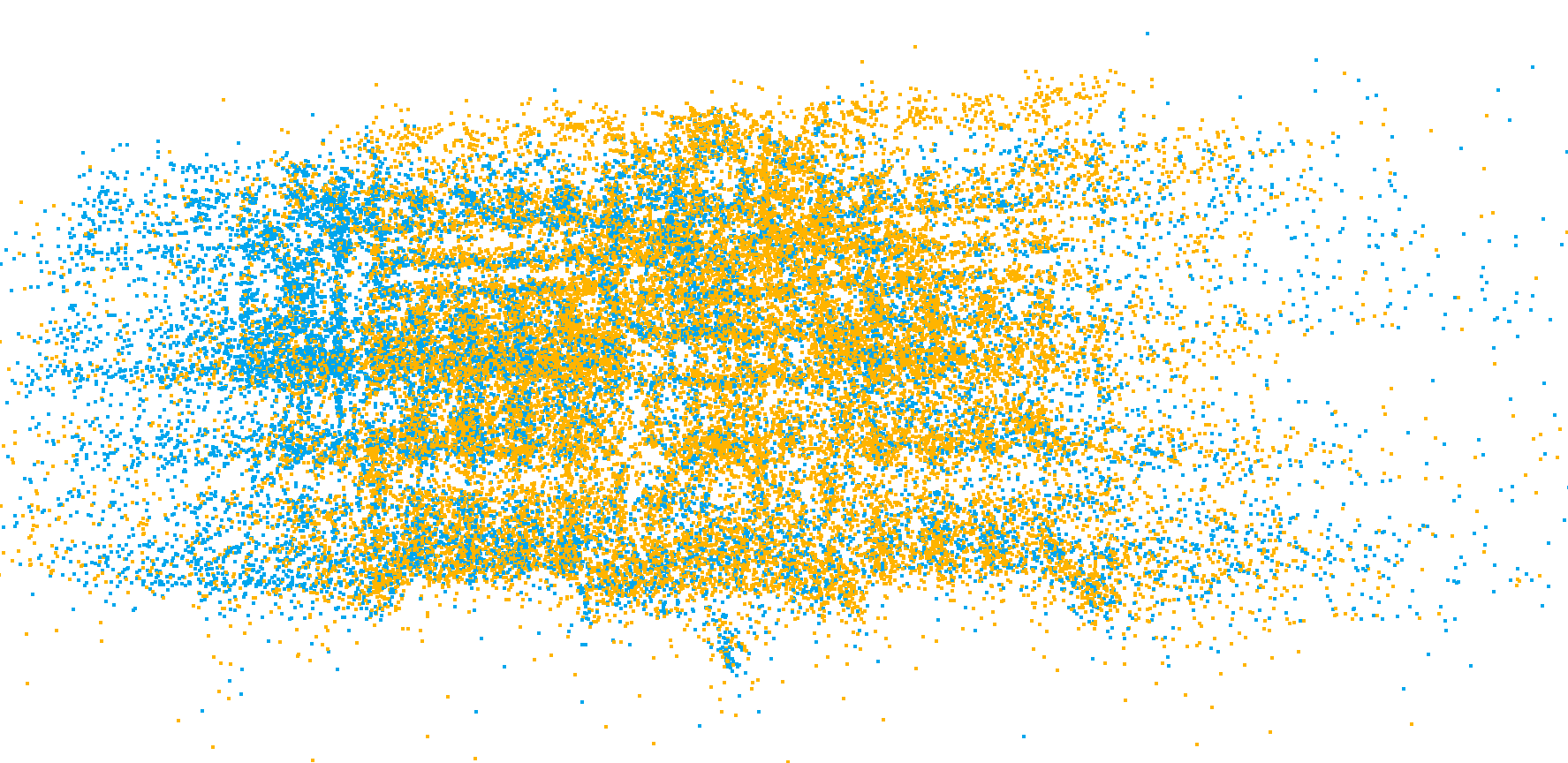} \\
    
    \raisebox{1\normalbaselineskip}[0pt][0pt]{\rotatebox[origin=l]{90}{\tt \large Shop Facade}}&
    \includegraphics[height=.21\textheight]{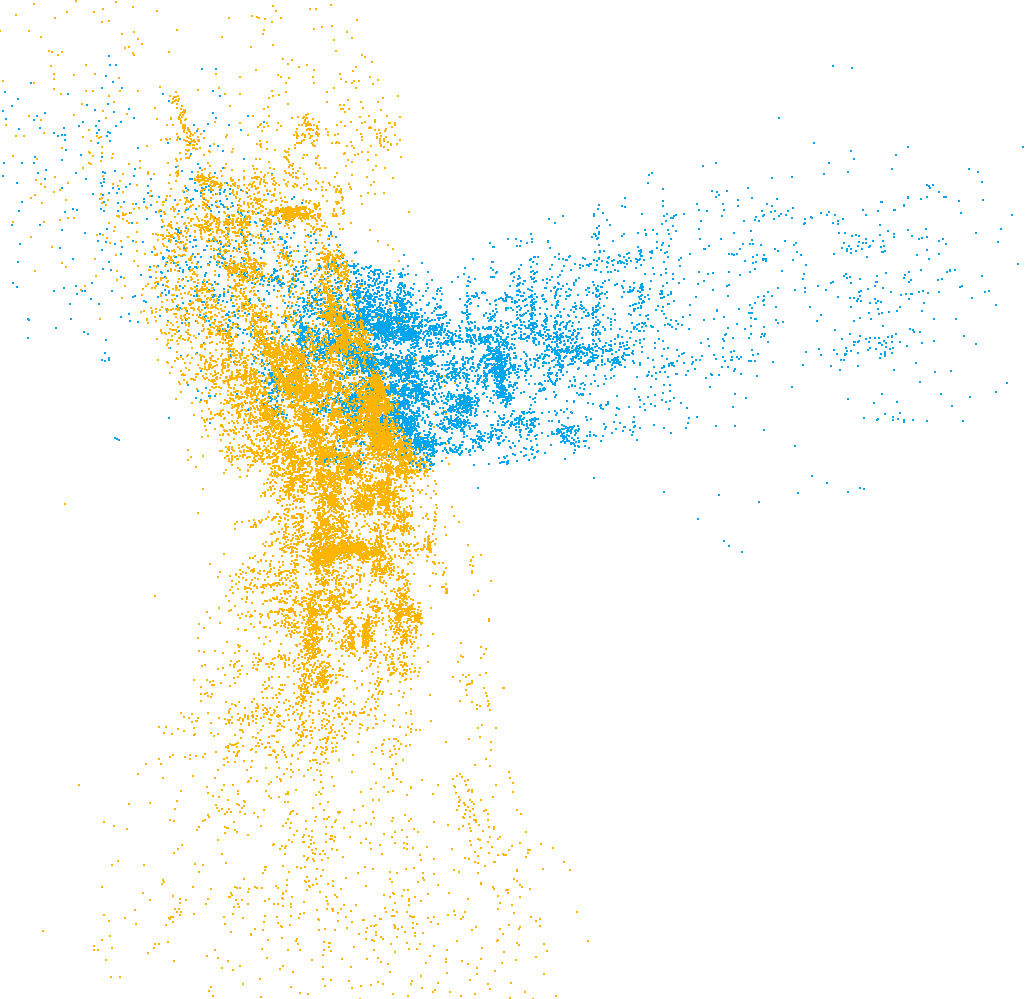} &
    \includegraphics[height=.21\textheight]{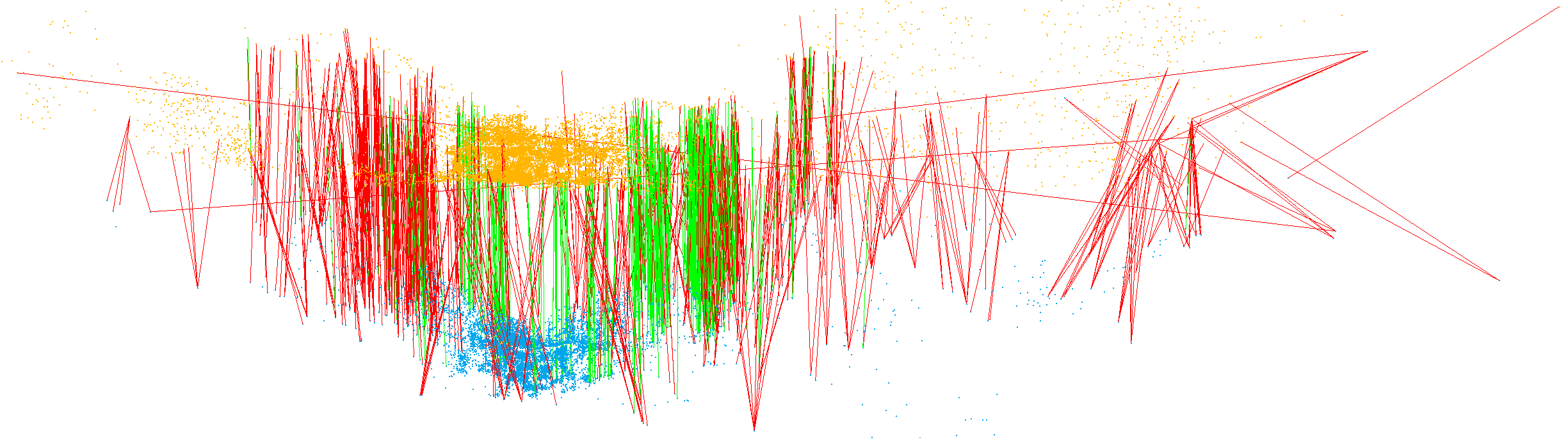} &
    \includegraphics[height=.21\textheight]{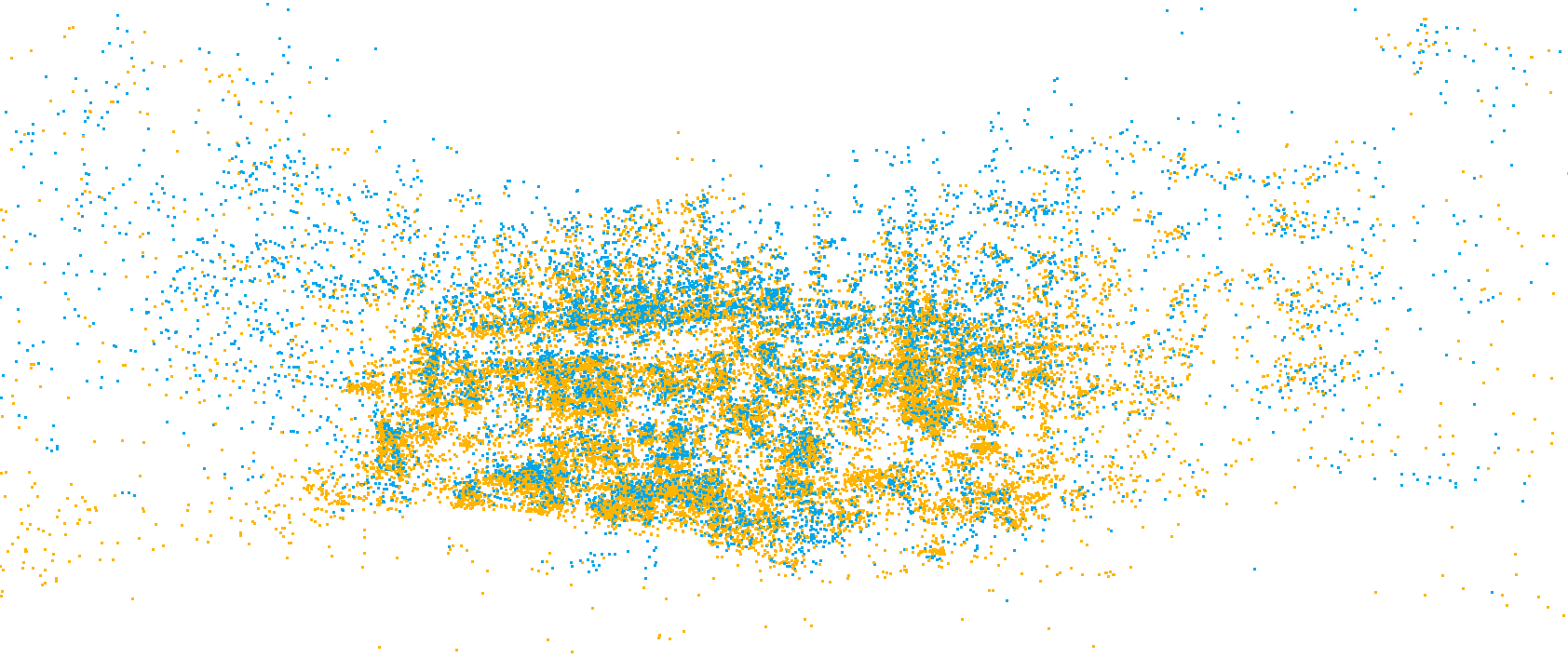} &
    \includegraphics[height=.21\textheight]{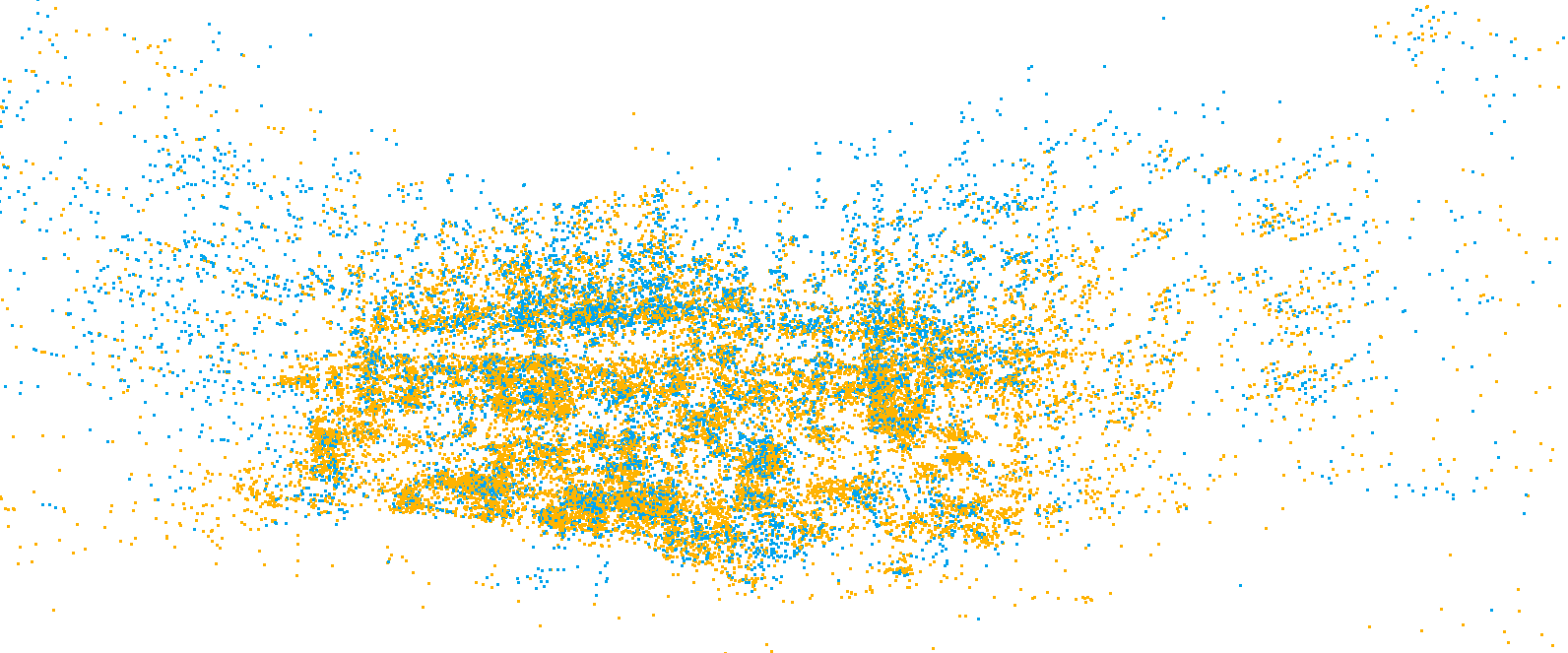} \\ \\
    
    \raisebox{1\normalbaselineskip}[0pt][0pt]{\rotatebox[origin=l]{90}{\tt \large St. Mary's Church}}&
    \includegraphics[height=.21\textheight]{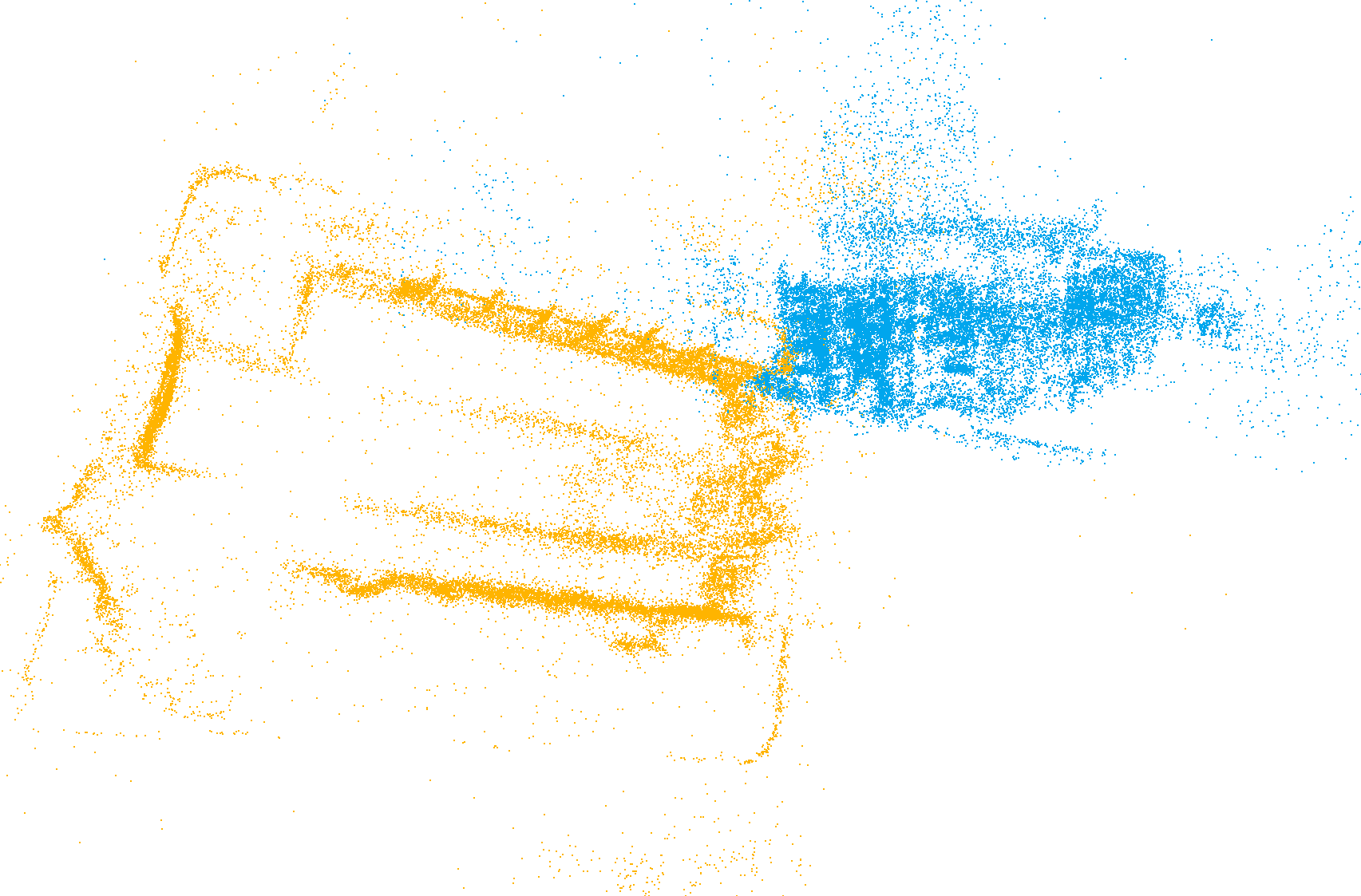} &
    \includegraphics[height=.21\textheight]{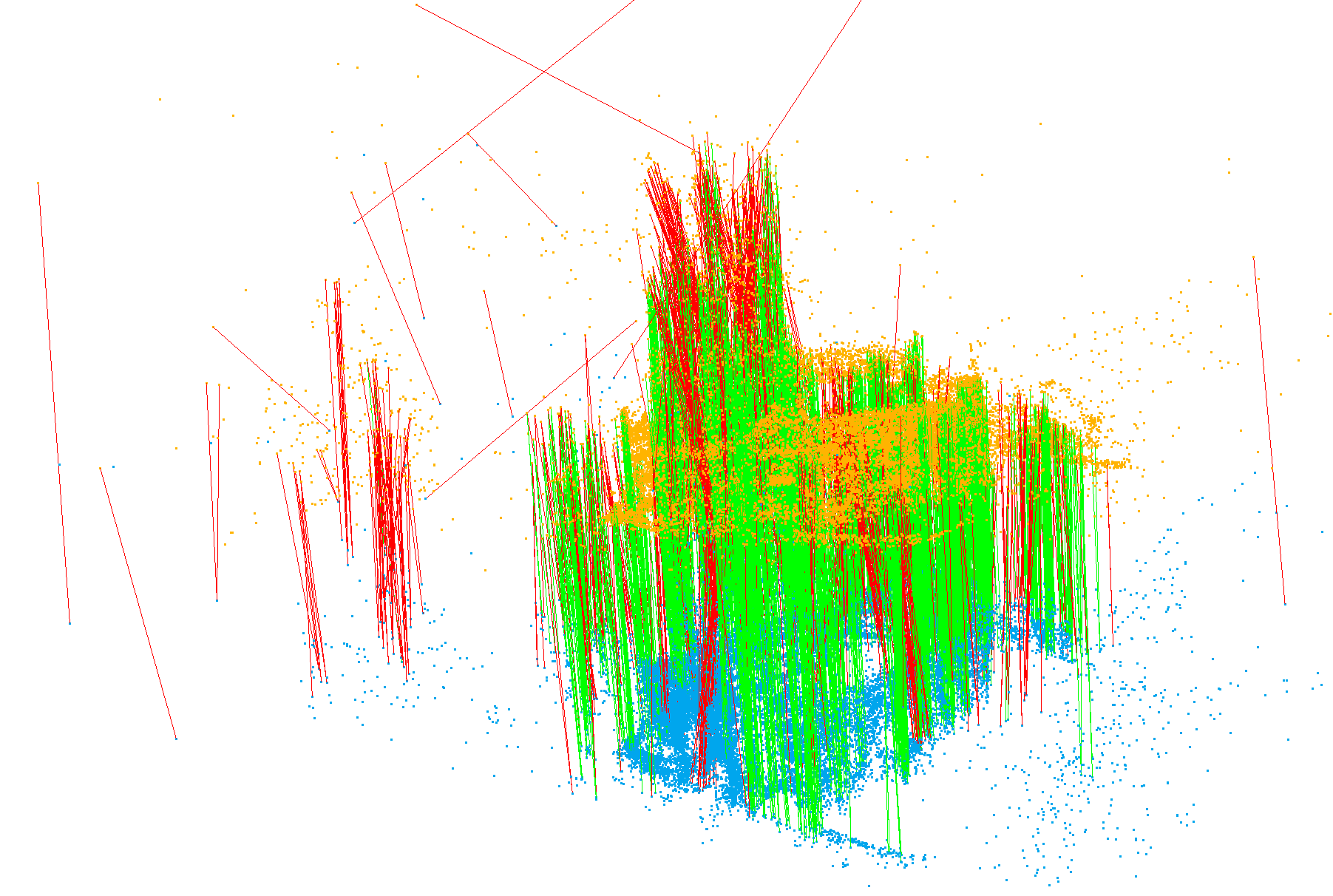} &
    \includegraphics[height=.21\textheight]{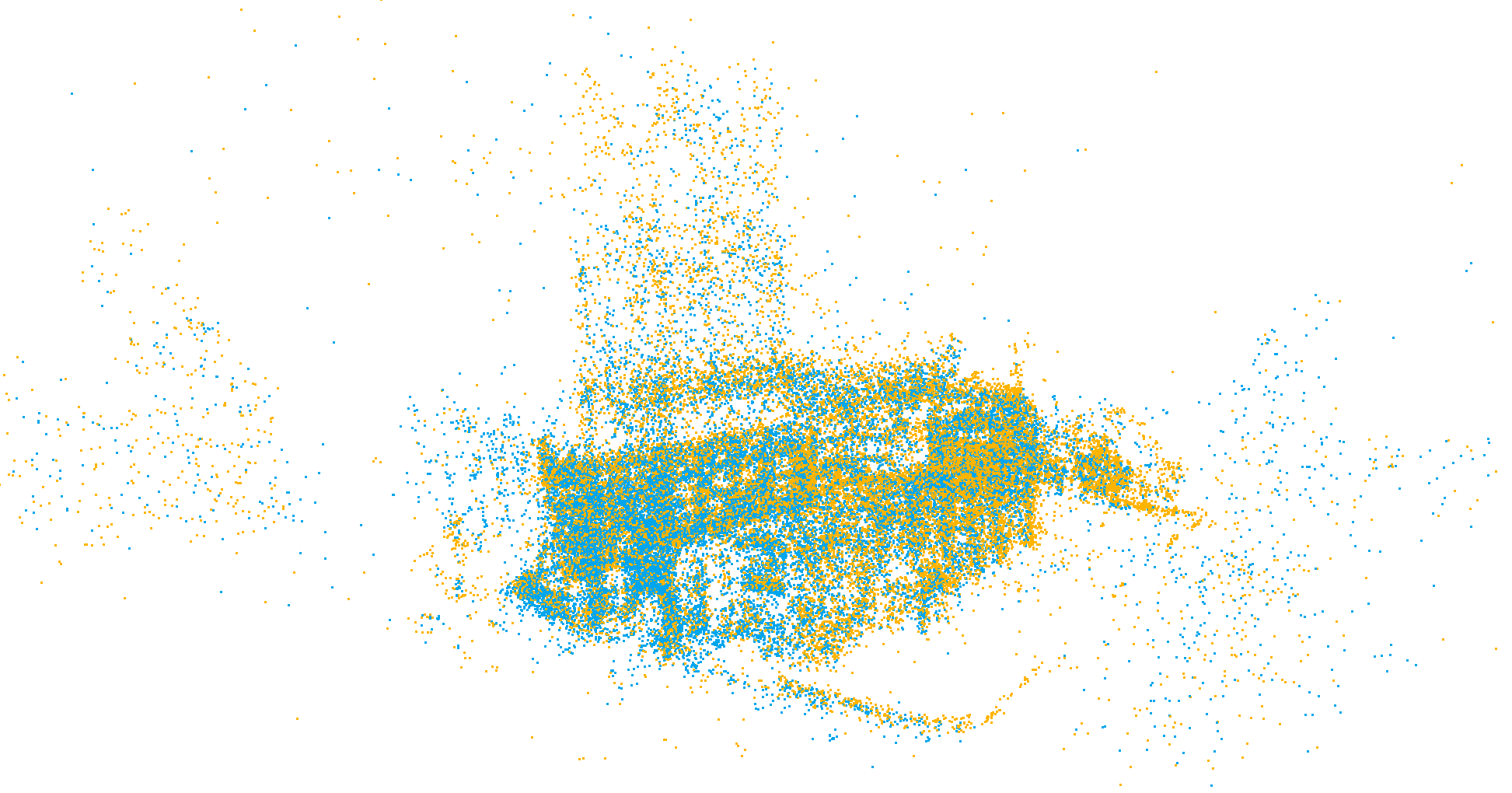} &
    \includegraphics[height=.21\textheight]{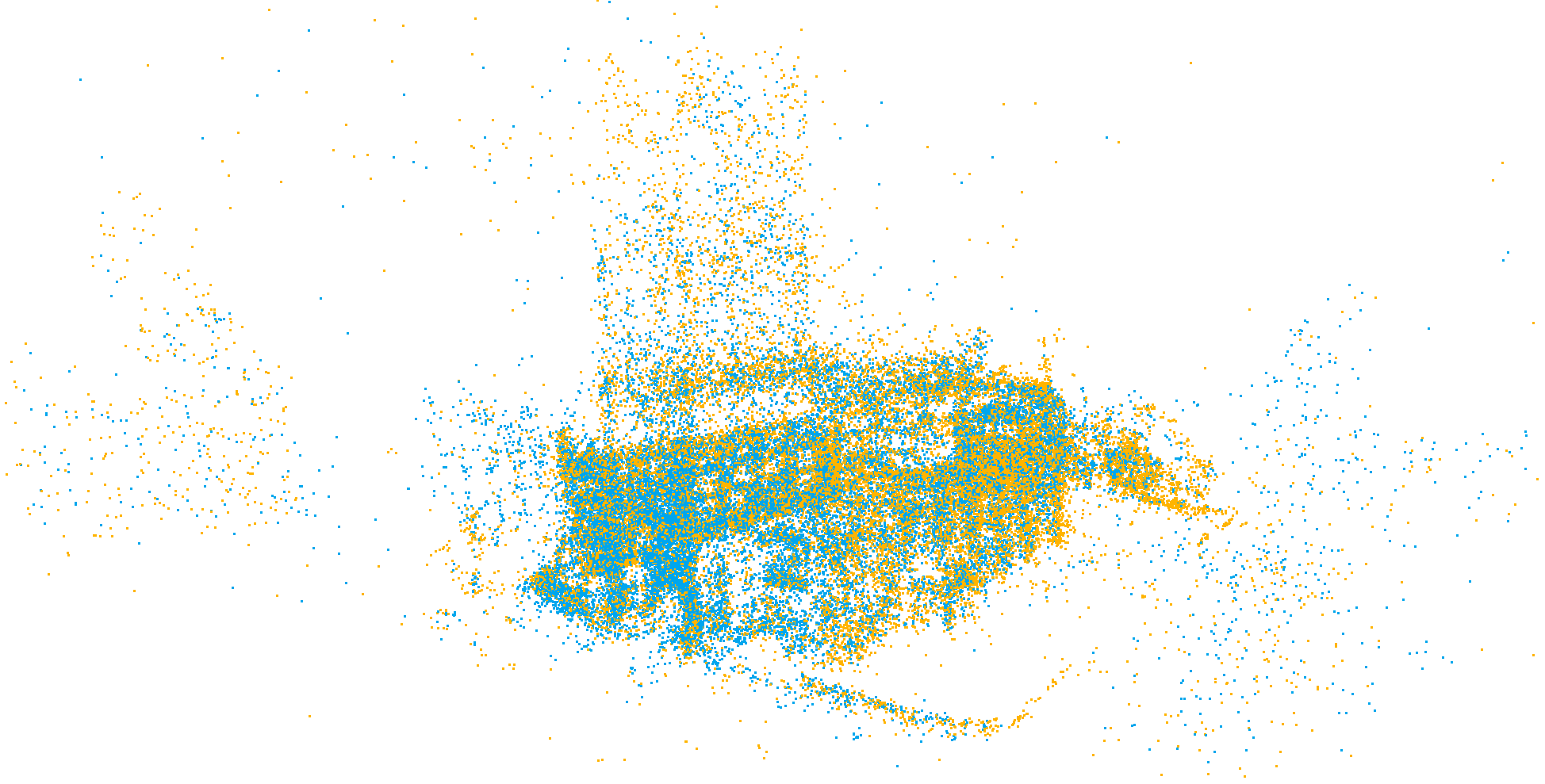} \\
    
    \end{tabular}
    }
    \caption{\textbf{Qualitative Results SfMRoITr on Cambridge Landmarks~\cite{kendall2015posenet}.} We evaluate SfMRoITr trained on our SfM Registration dataset on the five test scenes. We present the found matches and the results of the point cloud registration with the process presented in \cref{sec:pose_est}. The method is capable of obtaining accurate registration and a high number of good matches for all scenes. However, it tends to find outlier matches in sparser regions of the point clouds.}
    \label{fig:registration_results_cambridge}
    
\end{figure*}

\parsection{Cambridge Landmarks~\cite{kendall2015posenet}:}
Additionally, we also present qualitative results of \ours~trained only on our SfM Registration dataset in the five test scenes of the Cambridge Landmarks dataset.
The matches and registration results are presented in \cref{fig:registration_results_cambridge}.
Our method obtained a high number of good matches and hence the accuracy of the registration results, see \cref{tab:cambridge_reg}.
However, it produces a number of outliers in sparser areas of the point clouds, which was not the case in the other datasets see \cref{fig:registration_results_mega,fig:registration_results_qaud6k}.
Since the scenes in this dataset consist of videos instead of random sets of images, there is less viewpoint variability, which also leads to more far points being matched across multiple images and accepted by the SfM pipeline.
The scenes where this was more evident were \emph{Old Hospital} and \emph{Shop Facade}, which present almost flat facades with repetitive and symmetric structure. 
This is a limitation of the current method, which indicates an avenue for future research.

\section{Registration for Partial Scenes Reconstructed from Scratch}
In practice, registering two reconstructions will face issues of drift, as the partial scenes will be reconstructed from scratch, and not retriangulated (as in our training data).
To investigate the performance of our approach on this more realistic setting, we evaluated the SfM registration error on scenes from the Cambridge Landmarks dataset reconstructed from scratch (i.e., without using the ground truth poses for triangulation for one of the partial reconstructions), in~\Cref{tab:cambridge-scratch}.
Encouragingly, we find that the registration error is still low, indicating that there is no major distribution shift between the two tasks.

\begin{table*}
\centering
    \caption{\small{ \textbf{Results of SfM registration on Cambridge Landmarks}. Results for point clouds reconstructed from scratch (\ie, not retrianguled).}}
    \begin{tabular}{l cc cc cc cc cc}
        \toprule
        & \multicolumn{2}{c}{Great Court} & \multicolumn{2}{c}{Kings College}& \multicolumn{2}{c}{Old Hospital} & \multicolumn{2}{c}{Shop Facade} &  \multicolumn{2}{c}{St Mary's Church}\\
        \cmidrule(lr){2-3} \cmidrule(lr){4-5} \cmidrule(lr){6-7} \cmidrule(lr){8-9} \cmidrule(lr){10-11}
        Method & $\epsilon_R$ & $\epsilon_{\mathbf{t}}$ & $\epsilon_R$ & $\epsilon_{\mathbf{t}}$ & $\epsilon_R$ & $\epsilon_{\mathbf{t}}$ & $\epsilon_R$ & $\epsilon_{\mathbf{t}}$ & $\epsilon_R$ & $\epsilon_{\mathbf{t}}$ \\
        \midrule
        SIFT + NN & 0.32 & \textbf{0.09} & \textbf{0.3} & \textbf{0.02} & \textbf{1.37} & \textbf{0.02} & \textbf{0.92} & \textbf{0.06} & 0.49 & \textbf{0.01} \\
        \midrule
        \ours~(Mega)  & \textbf{0.30} & 0.10 & 0.55 & 0.05 & 1.41 & \textbf{0.02} & 1.03 & 0.15 & \textbf{0.21} & 0.02 \\
        RoiTr (Mega)  & 0.86 & 0.17 & 0.46 & 0.04 & 10.23 & 1.07 & 2.71 & 0.53 & 0.50 & 0.03 \\
        \bottomrule
    \end{tabular}
    \label{tab:cambridge-scratch}
\end{table*}

\section{Details on RoITr Architecture}
Here we expand on our baseline RoITr's~\citep{yu2023rotation} architecture in more detail.
\paragraph{Encoder $e_{\theta}$:}
The encoder consists of an Attentional Abstraction Layer (AAL) followed by $e$ PPF Attention Layers (PAL)~\citep[Figure 2, Section 3.2]{yu2023rotation}.
\paragraph{Global Transformer $g_{\theta}$:}
The global Transformer $g_{\theta}$ consists of $g$ blocks, each consisting of a Geometry-Aware Self-Attention Module (GSM), followed by a Position-Aware Cross-Attention Module (PCM)~\citep[Section 3.3, Figure 2, Figure 5]{yu2023rotation}.
\paragraph{Decoder $d_{\theta}$:}
The decoder $d_{\theta}$ consist of a Transition Up Layer (TUL) for upsampling and context aggregation, followed by $d$ PALs~\citep[Section 3.2, Figure 2]{yu2023rotation}.

\end{document}